\documentclass{article}

\usepackage{microtype}
\usepackage{graphicx}
\usepackage{subcaption}
\usepackage{booktabs} 

\usepackage{hyperref}

\usepackage[preprint]{icml2026}

\usepackage{amsmath}
\usepackage{amssymb}
\usepackage{mathtools}
\usepackage{amsthm}

\usepackage[capitalize,noabbrev]{cleveref}

\usepackage[textsize=tiny]{todonotes}

\usepackage{xurl}
\usepackage{listings}
\usepackage{float}
\usepackage{xcolor}
\usepackage{booktabs}
\usepackage{lipsum}
\usepackage{multicol}
\usepackage{multirow}
\usepackage{makecell}
\usepackage{adjustbox}
\usepackage{colortbl}
\usepackage{xspace}
\usepackage{microtype}
\usepackage{graphicx}
\usepackage{amssymb,amsmath}
\usepackage{mathtools}
\usepackage{amsthm}
\usepackage{pifont}
\usepackage{array}
\usepackage{longtable}

\usepackage{csquotes}
\usepackage{bm}
\usepackage{arydshln}
\usepackage{soul}
\usepackage[most]{tcolorbox}
\usepackage{subcaption} 
\usepackage{caption}

\definecolor{darkgreen}{RGB}{0,100,0} 
\definecolor{lightgreen}{RGB}{150, 255, 150}
\definecolor{lightblue}{RGB}{173, 216, 230}
\definecolor{lightred}{RGB}{255, 204, 204}
\definecolor{myblue}{RGB}{30, 144, 255}
\definecolor{mediumblue}{RGB}{0, 0, 205}
\definecolor{darkorange}{RGB}{255, 140, 0}

\newcommand{\hlc}[2][lightred]{\sethlcolor{#1}\hl{#2}}
\newcommand{\hlerror}[2][lightred]{\sethlcolor{#1}\hl{#2}}
\newcommand{\hlcorrect}[2][lightgreen]{\sethlcolor{#1}\hl{#2}}

\newtcolorbox{custombox}[1]{ 
  title=#1,
  colframe=black,
  colback=gray!10,
  coltitle=white,
  colbacktitle=black,
  fonttitle=\bfseries,
  sharp corners=south,
  boxrule=0.8pt
}

\newcommand{\ourmethod}{\textsc{\textcolor{black}{Self-Thought}}\xspace}
\newcommand{\ourmethodsmall}{\textsc{\textcolor{black}{Distil-Thought}}\xspace}

\newcommand{\initial}{\textsc{Initial \scriptsize{($t=0$)}}\xspace}
\newcommand{\vanilla}{\textsc{Reflex}\xspace}
\newcommand{\refine}{\textsc{Self-Refine}\xspace}
\newcommand{\tick}{\textsc{Self-TICK}\xspace}
\newcommand{\reflexion}{\textsc{Reflexion}\xspace}
\newcommand{\consistency}{\textsc{Self-Consistency}\xspace}
\newcommand{\supercorrect}{\textsc{SuperCorrect}\xspace}
\newcommand{\stwor}{\textsc{S$^{2}$R}\xspace}
\newcommand{\stasc}{\textsc{ST}a\textsc{SC}\xspace}

\newcommand{\thought}{\textsc{Thought-First}\xspace}
\newcommand{\metadata}{\textsc{Self-Metadata}\xspace}
\newcommand{\summary}{\textsc{Self-Summary}\xspace}

\newcommand\gptfouromini{\textsc{GPT-4o-}{\footnotesize \textsc{Mini}}\xspace}
\newcommand\gptfouro{\textsc{GPT-4o}\xspace}
\newcommand\othreemini{\textsc{o3-mini}\xspace}
\newcommand\deepseekrone{\textsc{DeepSeek-r1}\xspace}
\newcommand\qwen{\textsc{Qwen-2.5-7B}\xspace}
\newcommand\llama{\textsc{Llama-3.3-70B}\xspace}

\newcommand\game{Game of 24\xspace}
\newcommand\wordsorting{Word Sorting\xspace}
\newcommand\checkmate{CheckmateInOne\xspace}
\newcommand\aimefour{AIME 2024\xspace}
\newcommand\aimefive{AIME 2025\xspace}

\definecolor{gred}{HTML}{cc0200}
\definecolor{ggreen}{HTML}{418124}
\definecolor{Gray}{gray}{0.93}
\newcommand{\ua}[1]{\scriptsize\textcolor{ggreen}{\footnotesize $\uparrow$}{\color{ggreen}#1}} 
\newcommand{\da}[1]{\scriptsize\textcolor{gred}{\footnotesize $\downarrow$}{\color{gred}#1}} 

\newcolumntype{a}{>{\columncolor{Gray}}l}

\icmltitlerunning{Beyond Output Critique: Self-Correction via Task Distillation}

\begin{document}

\twocolumn[
  \icmltitle{Beyond Output Critique: Self-Correction via Task Distillation}



  \icmlsetsymbol{equal}{*}

  \begin{icmlauthorlist}
    \icmlauthor{Hossein A.~Rahmani}{sch,equal}
    \icmlauthor{Mengting Wan}{comp}
    \icmlauthor{Pei Zhou}{comp}
    \icmlauthor{Longqi Yang}{comp}
    \icmlauthor{Nick Craswell}{comp} \\
    \icmlauthor{Emine Yilmaz}{sch}
    \icmlauthor{Sujay Kumar Jauhar}{comp}
  \end{icmlauthorlist}

  \icmlaffiliation{sch}{AI Center, University College London}
  \icmlaffiliation{comp}{Microsoft. \textsuperscript{*}Work done while at Microsoft}

  \icmlcorrespondingauthor{Hossein A.~Rahmani}{hossein.rahmani.22@ucl.ac.uk}

  \icmlkeywords{Self-Correction, Large Language Model, Reasoning, Distillation}

  \vskip 0.3in
]



\printAffiliationsAndNotice{}  

\begin{abstract}
  Large language models (LLMs) have shown promising self-correction abilities, where iterative refinement improves the quality of generated responses. However, most existing approaches operate at the level of output critique, patching surface errors while often failing to correct deeper reasoning flaws. We propose \textbf{\ourmethod}, a framework that introduces an intermediate step of task abstraction before solution refinement. Given an input and an initial response, the model first distills the task into a structured template that captures key variables, constraints, and problem structure. This abstraction then guides solution instantiation, grounding subsequent responses in a clearer understanding of the task and reducing error propagation. Crucially, we show that these abstractions can be transferred across models: \textit{templates generated by larger models can serve as structured guides for smaller LLMs, which typically struggle with intrinsic self-correction}. By reusing distilled task structures, smaller models achieve more reliable refinements without heavy fine-tuning or reliance on external verifiers. Experiments across diverse reasoning tasks demonstrate that \ourmethod improves accuracy, robustness, and generalization for both large and small models, offering a scalable path toward more reliable self-correcting language systems.
\end{abstract}

\section{Introduction}
\label{sec:introduction}
Large Language Models (LLMs) have achieved remarkable progress in reasoning, problem-solving, and dialogue generation \citep[\textit{inter alia}]{brown2020language,chang2024survey,kojima2022large}. However, despite their impressive abilities, even the strongest models often produce errors such as flawed reasoning steps, factual mistakes, or inconsistent results \citep[\textit{inter alia}]{maynez2020faithfulness,gehman2020realtoxicityprompts,alkaissi2023artificial,yuan2023well}. Self-correction is a capability of LLMs that has recently emerged as a promising solution to mitigate these limitations \citep[\textit{inter alia}]{kamoi2024can,liu2024large}. Recent studies of this and other similar methods \citep[\textit{inter alia}]{madaan2023self-refine,shinn2023reflexion,welleck2022generating,chen2024teaching} show that models can critique their answers, generate feedback, and revise solutions. These methods highlight the promise of intrinsic self-correction, where models improve their own output through iterative refinement.

However, existing self-correction methods \citep{madaan2023self-refine,shinn2023reflexion,cook2024self-tick} have largely take the form of surface-level editing. A model generates an answer, evaluates that answer, and then attempts to patch errors. While effective in some cases, their efficacy in complex problem-solving, such as mathematical reasoning, remains limited. For instance, \textsc{Self-Refine} \citep{madaan2023self-refine} yields an average gain of 20\% across tasks, but only modest improvements on mathematical reasoning tasks even when aided by external signals. Similar limitations are observed in \textsc{Self-TICK} \citep{cook2024self-tick} and \textsc{ProgCo} \citep{song2025progco}, where gains on mathematics and reasoning benchmarks remain marginal compared to the other tasks.\footnote{See Appendix \ref{sec:app:baseline-results} for additional detailed analysis on results from baseline approaches.} Without a structured understanding of the task itself, corrections may be shallow, inconsistent, or fail to generalize beyond the specific example.

Moreover, current self-correction studies are primarily designed for large-scale models, relying on their extensive capacity to generate critiques and perform revisions \citep{madaan2023self-refine,cook2024self-tick,huang2023cannot-self-correct,kamoi2024can}. However, these methods often fail to extend to smaller models, which remain widely used in practice due to their efficiency, lower deployment costs, and utility in resource-constrained settings \citep{kamoi2024can,madaan2023self-refine,belcak2025small}. Despite their advantages, small models typically lack the reasoning depth and robustness of larger counterparts, and existing self-correction techniques provide little to no measurable improvement for them. This gap raises an important question of \textit{how to design self-correction mechanisms that are effective not only for frontier LLMs but also for smaller models, enabling them to benefit from iterative refinement}. Addressing this challenge is crucial for broadening the impact of self-correction beyond cutting-edge systems and enabling reliable reasoning across diverse model scales.

In this paper, we propose \textbf{\ourmethod}, a new framework for iterative self-correction that emphasizes task abstraction before refinement. Instead of immediately critiquing the output, the model first distills the problem into a structured template, identifying variables, constraints, and underlying problem types. This abstraction acts as a reusable guide that grounds subsequent reasoning. The model then instantiates this template to produce a refined solution. By separating understanding the task from solving it, our method reduces error propagation and leads to more robust corrections. Prior work \cite{sarthi2024raptor,jiang2023active,ho2023large}, performs recursive content summarization to improve retrieval and context construction. In contrast, \ourmethod performs task-level abstraction, which explicitly extracts variables, constraints, solution structure, and decision logic, independent of any supporting documents or external memory. While summarization compresses what is said and reasoning distillation compresses how a solution was obtained, task abstraction formalizes what must be solved, providing a reusable control representation that guides downstream reasoning and self-correction.


Moreover, we extend \ourmethod to smaller models through a variant called \ourmethodsmall. In this setting, we reuse the abstract templates distilled by larger, more capable models. These templates encapsulate high-level reasoning and self-correction strategies, allowing smaller models to benefit from structured guidance without requiring external verifiers or costly fine-tuning. By templatizing the problem-solving process, \ourmethodsmall enables smaller models to converge on solutions more quickly and with fewer iterative refinements. This not only improves performance but also offers a cost-saving advantage -- reusable templates reduce computational overhead and accelerate inference, making the approach more efficient and scalable across model sizes.


We evaluate our approaches on a range of LLMs, including \gptfouromini, \gptfouro, \othreemini, \deepseekrone, and open-source models \qwen and \llama, across a wide range of tasks. Our findings demonstrate that \ourmethod consistently surpasses prior techniques, obviating the need for supplementary data or training. For example, when applied to \gptfouromini, \ourmethod attains $126.30\%$ enhancements on \game, $81.82\%$ gains on \wordsorting, and a $199.85\%$ improvement on \aimefive. Similarly, on small models such as \qwen and \llama, \ourmethodsmall yields notable gains, including $154.54\%$ average improvement on \qwen and $121.42\%$ on \llama, demonstrating that task abstractions learned from large models can effectively transfer to smaller models.

Our work reframes self-correction as a process of thinking about the task rather than fixing the answer. This perspective not only enhances the performance of large models but also provides a scalable path for empowering smaller ones with structured reasoning support.
\section{\ourmethod}
Our proposed method, \ourmethod, builds upon the principle of iterative refinement but introduces a unique intermediate step, \textit{task abstraction}. This abstraction process distills the core components of a problem, transforming it into a structured template that guides the model in generating a more precise and accurate final solution. 
\ourmethod first creates a canonical representation of the task itself, ensuring that subsequent solution generation is grounded in a deep understanding of the objectives and constraints of the problem. 
See Algorithm \ref{alg:method} for a summary of \ourmethod method, and we describe \ourmethod in more detail in the following sections.

\renewcommand{\algorithmiccomment}[1]{\hfill{\(\triangleright\)~#1}\par}
\begin{algorithm}[tb]
    \caption{\textsc{\ourmethod}}
    \small
    \begin{algorithmic}
        \STATE {\bfseries Input:} task $x$, model $\mathcal{M}$, prompts $\{\Im, \wp, \Re\}$, stop condition \textsf{stop}($\cdot$), number of iterations $n$
        \STATE {\bfseries Output:} Corrected output $\hat{y}$ from $\mathcal{M}$
        \STATE Generate initial output $\hat{y}_0 \sim \mathbb{P}_{\mathcal{M}}(\cdot|\Im \oplus x)$ \COMMENT{Initialization}
        \FOR{$t=1$ {\bfseries to} $n$}
            \STATE $d_{t} \sim \mathbb{P}_{\mathcal{M}}(\cdot|\wp \oplus x \oplus \hat{y}_t)$ \COMMENT{Task Abstraction}
            \IF{$\textsf{stop}(d_{t}, t)$}
            \STATE {\bfseries return} $\hat{y}_{t}$ \COMMENT{Stopping Condition}
            \ENDIF
            \STATE $\hat{y}_{t+1} \sim \mathbb{P}_{\mathcal{M}}(\cdot|\Re \oplus x \oplus \hat{y}_{t} \oplus d_t)$ \COMMENT{Instantiation}
        \ENDFOR
        \STATE {\bfseries return} $\hat{y}_{n}$
    \end{algorithmic}
    \label{alg:method}
\end{algorithm}

\textbf{Initial Generation.}
\ourmethod generates an initial response $\hat{y}_0$ given an input $x$, prompt $\Im$, and a model $\mathcal{M}$, $\hat{y}_0 \sim \mathbb{P}_{\mathcal{M}}(\cdot | \Im \oplus x)$, where $\Im$ is a task-specific prompt for an initial generation, and $\oplus$ denotes concatenation. This is a standard generation process where the model produces a preliminary response to user input. The initial response serves as the starting point for the subsequent self-correction loop.

\begin{figure*}
    \centering
    \includegraphics[scale=0.6]{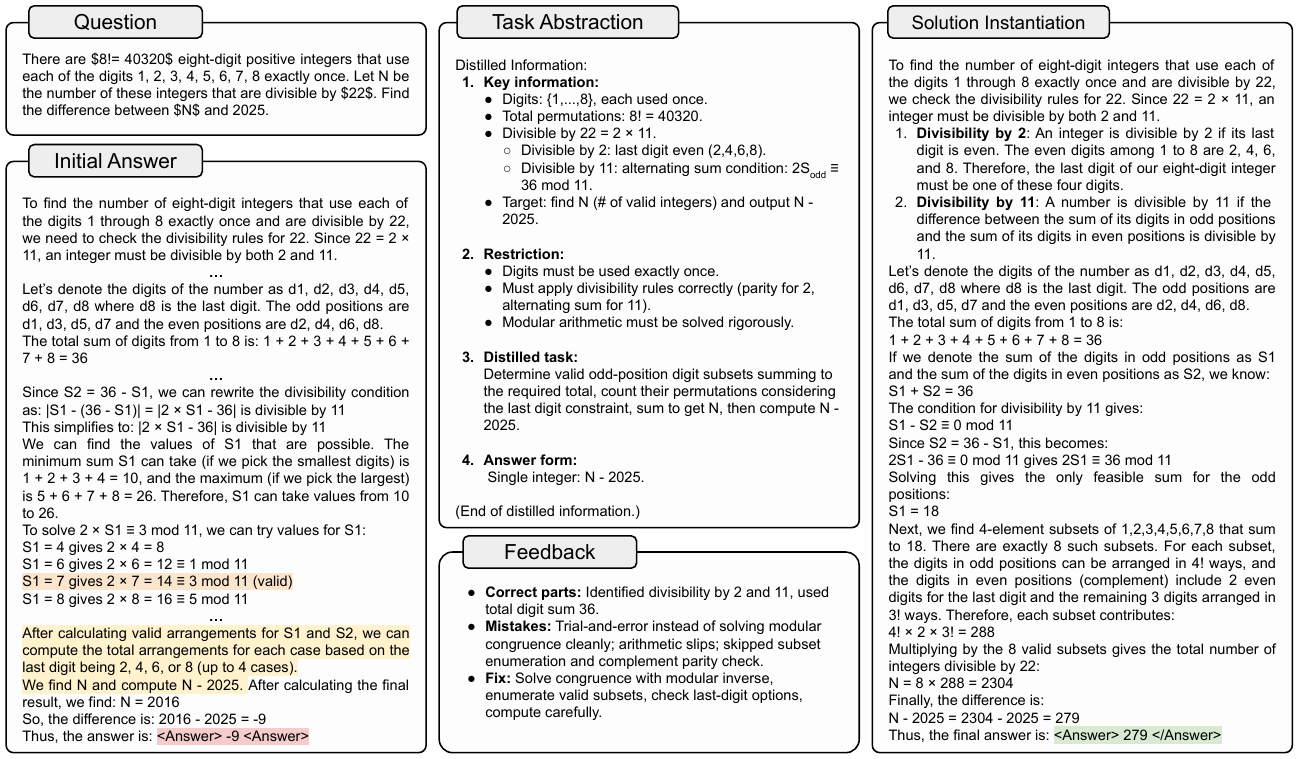}
    \caption{We present an example trace of \ourmethod self-correcting on a sample from \aimefive using \gptfouromini. The initial answer is simplified for clarity, and the full response is provided in Table \ref{tab:full-answer} in the Appendix. This initial response contains \textcolor[RGB]{200,120,50}{logical reasoning errors}, \textcolor[RGB]{230,200,60}{incomplete calculations}, and an \textcolor[RGB]{200,100,100}{incorrect final result}. By applying task abstraction, \ourmethod successfully identifies and corrects these mistakes.}
    \label{fig:placeholder}
\end{figure*}

\textbf{Task Abstraction.}
Next, \ourmethod focuses on abstracting the user's input query $x$ and the initial response $\hat{y}_0$ into a structured task representation $d$. This is guided by a structured prompt $\wp$, which facilitates the distillation of the problem into a generalized, constraint-aware template. The model $\mathcal{M}$ is instructed to act as a ``meta-distiller'' and perform a critical analysis of the user's request. This analysis systematically extracts and formalizes the essential elements of the problem: $d_{t} \sim \mathbb{P}_{\mathcal{M}}(\cdot|\wp \oplus x \oplus \hat{y}_t)$. The output, $d$, is a structured object that encapsulates the essence of the problem in a format designed to guide subsequent solution generation. An example is presented in Figure~\ref{fig:placeholder}.

Functionally, as a first step, our method extracts \emph{key information}, identifying all salient variables, values, and data points from $x$. Concurrently, it formalizes the \emph{problem restrictions}, such as mathematical operator precedence or physical laws, to ensure the solution adheres to real-world rules. These explicitly defined constraints are crucial for preventing errors. Finally, \ourmethod generalizes the problem, reframing $x$ into a higher-level, more abstract, \emph{distilled task} to ensure the solution is robust and applicable to a wider range of similar inputs. Depending on the problem, the structured task abstraction may also contain other constraints such as expected \emph{answer format}.

In concert, these steps encourage the model to identify the underlying problem \emph{type} rather than focusing solely on the specific example. The process also translates the problem into an algorithmic structure, identifying required input parameters and data types, effectively preparing the problem for a programmatic solution. This comprehensive, multi-faceted analysis in the abstraction phase ensures that the final solution in the next phase is grounded in a deep, accurate understanding of the problem's structure and constraints.

\textbf{Solution Instantiation.}
Next, \ourmethod utilizes the distilled information $d$ to generate a specific, concrete solution. The objective here is to instantiate an improved answer ($\hat{y}_{t+1}$) by applying the abstract knowledge of $d$ to the initial query $x$ and response $\hat{y}_{t}$. The model is provided with the prompt $\Re$ to act as a problem-solving expert to analyze $d$, the input query $x$, and previous output $\hat{y}_t$ to produce a refined and accurate response $\hat{y}_{t+1} \sim \mathbb{P}_{\mathcal{M}}(\cdot|\Re \oplus x \oplus \hat{y}_{t} \oplus d_{t})$. The presence of explicitly defined constraints and the abstracted task in $d$ serves as a powerful guide, significantly reducing the likelihood of errors and ensuring that the solution aligns with the true intent of the problem.

The \ourmethod method thus establishes a self-correction loop where the model's internal analysis of the problem, rather than a critique of its initial output, becomes the mechanism for refinement. This approach ensures that the final response is not merely a corrected version of an initial attempt but a well-reasoned solution derived from a foundational understanding of the problem's structure. This two-step abstraction and instantiation process leads to more robust, reliable, and consistent performance across a wide range of tasks.

\begin{table*}[!t]
\centering
\caption{A single step of self-correction performance on \game, \wordsorting, \checkmate, \aimefour, \aimefive with \gptfouromini, \gptfouro, \othreemini, and \deepseekrone. Green ({\scriptsize\textcolor{ggreen}{\footnotesize $\uparrow$}}) and red ({\scriptsize\textcolor{gred}{\footnotesize $\downarrow$}}) arrows indicate performance changes against the previous attempt (i.e., \initial). \textbf{Bold} corresponds to the best performance. We find that \ourmethod consistently yields positive gains between the first and second attempts, demonstrating stable improvements. While baseline approaches often erroneously modify a correct response into an incorrect one, \ourmethod preserves correctness and consistently improves LLM performance.}
\label{tab:large_delta_t1}
\resizebox{\textwidth}{!}{
    \setlength{\tabcolsep}{1pt}
    \begin{tabular}{llccllccllccllccllccla}
        \toprule
        \multirow{2}{*}{\textbf{Method}} & \multicolumn{3}{c}{\bf Game of 24} && \multicolumn{3}{c}{\bf Word Sorting} && \multicolumn{3}{c}{\bf CheckmateInOne} && \multicolumn{3}{c}{\bf AIME 2024} && \multicolumn{3}{c}{\bf AIME 2025} && \multicolumn{1}{c}{\bf Mean} \cr
        
        \cmidrule(lr){2-4} \cmidrule(lr){6-8} \cmidrule(lr){10-12} \cmidrule(lr){14-16} \cmidrule(lr){18-20}
        
        & \makecell{\textbf{Acc@t1}} & \makecell{$\Delta^{{i \to c}}(t_0, t_1)$} & \makecell{$\Delta^{{c \to i}}(t_0, t_1)$} && 
          \makecell{\textbf{Acc@t1}} & \makecell{$\Delta^{{i \to c}}(t_0, t_1)$} & \makecell{$\Delta^{{c \to i}}(t_0, t_1)$} &&
          \makecell{\textbf{Acc@t1}} & \makecell{$\Delta^{{i \to c}}(t_0, t_1)$} & \makecell{$\Delta^{{c \to i}}(t_0, t_1)$} &&
          \makecell{\textbf{Acc@t1}} & \makecell{$\Delta^{{i \to c}}(t_0, t_1)$} & \makecell{$\Delta^{{c \to i}}(t_0, t_1)$} &&
          \makecell{\textbf{Acc@t1}} & \makecell{$\Delta^{{i \to c}}(t_0, t_1)$} & \makecell{$\Delta^{{c \to i}}(t_0, t_1)$} &&
          \makecell{\textbf{Acc@t1}} \\
            \midrule
            & \multicolumn{21}{c}{\textbf{\gptfouromini}} \\
            \midrule
            \initial & 38.78 & - & - && 55.0 & - & - &&30.67 & - & - &&20.0 & - & - &&6.67 & - & - && 30.0 \\ 
            \hdashline\noalign{\vskip 0.4ex}
            \vanilla & 24.49 \da{14.29} & 8.16 & 22.45 && 60.0 \ua{5.0} & 7.5 & 2.5 && 9.33 \da{21.34} & 5.33 & 26.67 && 10.0 \da{10.0} & 0.0 & 10.0 && 10.0 \ua{3.33} & 3.33 & 0.0 && 23.0 \da{7.0} \\ 
            \refine & 25.51 \da{13.27} & 11.22 & 24.49 && 58.75 \ua{3.75} & 7.5 & 3.75 && 10.67 \da{20.0} & 6.67 & 26.67 && 13.33 \da{6.67} & 3.33 & 10.0 && 16.67 \ua{10.0} & 10.0 & 0.0 && 25.0 \da{5.0} \\ 
            \tick & 38.78  & 18.37 & 18.37 && 40.0 \da{15.0} & 7.5 & 22.5 && 20.0 \da{10.67} & 6.67 & 17.33 && 23.33 \ua{3.33} & 6.67 & \textbf{3.33} && 13.33 \ua{6.66} & 6.67 & 0.0 && 27.0 \da{3.0} \\ 
            \reflexion & 26.53 \da{12.25} & 9.18 & 21.43 && 60.0 \ua{5.0} & 11.25 & 6.25 && 9.33 \da{21.34} & 4.0 & 25.33 && 13.33 \da{6.67} & 6.67 & 13.33 && 6.67 & 3.33 & 3.33 && 23.0 \da{7.0} \\ 
            \ourmethod & \textbf{87.76} \textbf{\ua{48.98}} & \textbf{51.02} & \textbf{2.04} && \textbf{100.0} \textbf{\ua{45.0}} & \textbf{45.0} & \textbf{0.0} && \textbf{33.33} \textbf{\ua{2.66}} & \textbf{14.67} & \textbf{12.0} && \textbf{30.0} \textbf{\ua{10.0}} & \textbf{16.67} & 6.67 && \textbf{20.0} \textbf{\ua{13.33}} & \textbf{13.33} & 0.0 && \textbf{54.0} \textbf{\ua{24.0}} \\
            \midrule
            & \multicolumn{21}{c}{\textbf{\gptfouro}} \\
            \midrule
            \initial & 17.35 & - & - && 86.25 & - & - &&41.33 & - & - &&13.33 & - & - &&10.0 & - & - && 34.0 \\ 
            \hdashline\noalign{\vskip 0.4ex}
            \vanilla & 19.39 \ua{2.04} & 11.22 & 9.18 && 81.25 \da{5.0} & 3.75 & 8.75 && 26.67 \da{14.66} & 12.0 & 26.67 && 13.33  & 0.0 & 0.0 && 6.67 \da{3.33} & 0.0 & 3.33 && 29.0 \da{5.0} \\ 
            \refine & 33.67 \ua{16.32} & 27.55 & 11.22 && 78.75 \da{7.5} & 7.5 & 15.0 && 38.67 \da{2.66} & 13.33 & 16.0 && 20.0 \ua{6.67} & 10.0 & 3.33 && 10.0  & 6.67 & 6.67 && 36.0 \ua{2.0} \\ 
            \tick & 30.61 \ua{13.26} & 24.49 & 11.22 && 70.0 \da{16.25} & 3.75 & 20.0 && 30.67 \da{10.66} & 16.0 & 26.67 && 16.67 \ua{3.34} & 3.33 & 0.0 && 10.0  & 6.67 & 6.67 && 32.0 \da{2.0} \\ 
            \reflexion & 36.73 \ua{19.38} & 27.55 & \textbf{8.16} && 82.5 \da{3.75} & 6.25 & 10.0 && 25.33 \da{16.0} & 16.0 & 32.0 && 16.67 \ua{3.34} & 10.0 & 6.67 && 10.0  & 6.67 & 6.67 && 34.0  \\ 
            \ourmethod & \textbf{37.76} \textbf{\ua{20.41}} & \textbf{30.61} & 10.2 && \textbf{100.0} \textbf{\ua{13.75}} & \textbf{13.75} & \textbf{0.0} && \textbf{65.33} \textbf{\ua{24.0}} & \textbf{32.0} & \textbf{8.0} && \textbf{33.33} \textbf{\ua{20.0}} & \textbf{20.0} & \textbf{0.0} && \textbf{16.67} \textbf{\ua{6.67}} & \textbf{10.0} & \textbf{3.33} && \textbf{51.0} \textbf{\ua{17.0}} \\
            \midrule
            & \multicolumn{21}{c}{\textbf{\othreemini}} \\
            \midrule
            \initial & 86.73 & - & - && 90.0 & - & - &&34.67 & - & - &&80.0 & - & - &&73.33 & - & - && 73.0 \\ 
            \hdashline\noalign{\vskip 0.4ex}
            \vanilla & 83.67 \da{3.06} & 5.1 & \textbf{8.16} && 90.0  & 8.75 & 8.75 && 32.0 \da{2.67} & 5.33 & \textbf{8.0} && 80.0  & 3.33 & 3.33 && 76.67 \ua{3.34} & 3.33 & 0.0 && 72.0 \da{1.0} \\ 
            \refine & 86.73  & 11.22 & 11.22 && 87.5 \da{2.5} & 8.75 & 11.25 && 20.0 \da{14.67} & 4.0 & 18.67 && 83.33 \ua{3.33} & 10.0 & 6.67 && 73.33  & 6.67 & 6.67 && 70.0 \da{3.0} \\ 
            \tick & 0.0 \da{86.73} & 0.0 & 86.73 && 87.5 \da{2.5} & 7.5 & 10.0 && 13.33 \da{21.34} & 2.67 & 24.0 && 76.67 \da{3.33} & 3.33 & 6.67 && 66.67 \da{6.66} & \textbf{10.0} & 16.67 && 49.0 \da{24.0} \\ 
            \reflexion & 84.69 \da{2.04} & 7.14 & 9.18 && 97.5 \ua{7.5} & 7.5 & \textbf{0.0} && 32.0 \da{2.67} & 12.0 & 14.67 && 80.0  & 3.33 & 3.33 && 66.67 \da{6.66} & 3.33 & 10.0 && 72.0 \da{1.0} \\ 
            \ourmethod & \textbf{88.78} \textbf{\ua{2.05}} & 11.22 & 9.18 && 97.5 \ua{7.5} & 8.75 & 1.25 && \textbf{37.33} \textbf{\ua{2.66}} & \textbf{21.33} & 18.67 && \textbf{86.67} \textbf{\ua{6.67}} & 6.67 & \textbf{0.0} && \textbf{80.0} \ua{6.67} & 6.67 & 0.0 && \textbf{78.0} \textbf{\ua{5.0}} \\ 
            \midrule
            & \multicolumn{21}{c}{\textbf{\deepseekrone}} \\
            \midrule
            \initial & 84.69 & - & - && 97.5 & - & - &&17.33 & - & - &&80.0 & - & - &&63.33 & - & - && 69.0 \\ 
            \hdashline\noalign{\vskip 0.4ex}
            \vanilla & 64.29 \da{20.4} & 5.1 & 25.51 && 93.75 \da{3.75} & 2.5 & 6.25 && 16.0 \da{1.33} & 9.33 & 10.67 && 76.67 \da{3.33} & 0.0 & \textbf{3.33} && 63.33  & 6.67 & 6.67 && 63.0 \da{6.0} \\ 
            \refine & 52.04 \da{32.65} & 5.1 & 37.76 && 88.75 \da{8.75} & 1.25 & 10.0 && 16.0 \da{1.33} & 10.67 & 12.0 && 76.67 \da{3.33} & 3.33 & 6.67 && 70.0 \ua{6.67} & \textbf{20.0} & 13.33 && 61.0 \da{8.0} \\ 
            \tick & 17.35 \da{67.34} & 2.04 & 69.39 && 91.25 \da{6.25} & 0.0 & 6.25 && 5.33 \da{12.0} & 1.33 & 13.33 && 60.0 \da{20.0} & 0.0 & 20.0 && 53.33 \da{10.0} & 6.67 & 16.67 && 45.0 \da{24.0} \\ 
            \reflexion & 50.0 \da{34.69} & 6.12 & 40.82 && 90.0 \da{7.5} & 0.0 & 7.5 && 18.67 \ua{1.34} & \textbf{14.67} & 13.33 && 56.67 \da{23.33} & 3.33 & 26.67 && 60.0 \da{3.33} & 6.67 & 10.0 && 55.0 \da{14.0} \\ 
            \ourmethod & \textbf{85.71} \textbf{\ua{1.02}} & \textbf{12.24} & \textbf{11.22} && \textbf{100.0} \textbf{\ua{2.5}} & 2.5 & \textbf{0.0} && \textbf{20.0} \textbf{\ua{2.67}} & 12.0 & \textbf{9.33} && \textbf{80.0} & \textbf{10.0} & 10.0 && \textbf{73.33} \textbf{\ua{10.0}} & 13.33 & \textbf{3.33} && \textbf{72.0} \textbf{\ua{3.0}} \\ 
            \bottomrule
    \end{tabular}
}
\end{table*}

\textbf{Task Distillation for Smaller Models.}
While \ourmethod is model-agnostic, we extend its utility to settings where smaller language models struggle with abstraction. In such cases, we leverage the output of the \textit{Task Abstraction} step, $d$, produced by a stronger model. This distilled representation serves as a reusable template that encodes the essential problem structure, constraints, and solution strategy. Given a distilled abstraction $d$ generated by a larger model $\mathcal{M}_{L}$, a smaller model $\mathcal{M}_{S}$ can instantiate the solution as: $\hat{y}^{S}_{t+1} \sim \mathbb{P}_{\mathcal{M}_{S}}(\Re \oplus x \oplus \hat{y}^{S}_{t} \oplus d_{L} \oplus d_{S})$, where $\hat{y}^{S}_{t}$ denotes the current output of the smaller model. 

This extension effectively performs \textit{task distillation}: instead of requiring $\mathcal{M}_{S}$ to perform high-level reasoning from scratch, it inherits the abstract reasoning trace from $\mathcal{M}_{L}$. As a result, smaller models benefit from the structured guidance in $d$, leading to more accurate self-corrections without incurring the computational overhead of repeatedly prompting larger models. 

By decoupling abstraction from instantiation, \ourmethod not only improves self-correction in a single model but also provides a scalable mechanism for transferring distilled reasoning to less capable models.
\section{Experimental Setup}
\label{sec:setup}

\begin{table*}[!t]
\centering
\caption{A single step of self-correction performance on \game, \wordsorting, \checkmate, \aimefour, \aimefive with \textit{small} models, \qwen and \llama. Green ({\scriptsize\textcolor{ggreen}{\footnotesize $\uparrow$}}) and red ({\scriptsize\textcolor{gred}{\footnotesize $\downarrow$}}) arrows indicate performance changes against the previous attempt (i.e., \initial). \textbf{Bold} corresponds to the best performance. Both \ourmethod and \ourmethodsmall achieve consistent improvements, with the latter leveraging task abstractions.}
\resizebox{\textwidth}{!}{
    \setlength{\tabcolsep}{1pt}
    \begin{tabular}{llccllccllccllccllccla}
    \toprule
    \multirow{2}{*}{\textbf{Method}} & \multicolumn{3}{c}{\bf Game of 24} && \multicolumn{3}{c}{\bf Word Sorting} && \multicolumn{3}{c}{\bf CheckmateInOne} && \multicolumn{3}{c}{\bf AIME 2024} && \multicolumn{3}{c}{\bf AIME 2025} && \multicolumn{1}{c}{\bf Mean} \cr
    \cmidrule(lr){2-4} \cmidrule(lr){6-8} \cmidrule(lr){10-12} \cmidrule(lr){14-16} \cmidrule(lr){18-20}
    & \makecell{\textbf{Acc@t1}} & \makecell{$\Delta^{{i \to c}}(t_0, t_1)$} & \makecell{$\Delta^{{c \to i}}(t_0, t_1)$} && 
    \makecell{\textbf{Acc@t1}} & \makecell{$\Delta^{{i \to c}}(t_0, t_1)$} & \makecell{$\Delta^{{c \to i}}(t_0, t_1)$} &&
    \makecell{\textbf{Acc@t1}} & \makecell{$\Delta^{{i \to c}}(t_0, t_1)$} & \makecell{$\Delta^{{c \to i}}(t_0, t_1)$} &&
    \makecell{\textbf{Acc@t1}} & \makecell{$\Delta^{{i \to c}}(t_0, t_1)$} & \makecell{$\Delta^{{c \to i}}(t_0, t_1)$} &&
    \makecell{\textbf{Acc@t1}} & \makecell{$\Delta^{{i \to c}}(t_0, t_1)$} & \makecell{$\Delta^{{c \to i}}(t_0, t_1)$} &&
    \makecell{\textbf{Acc@t1}} \\
    \midrule
    & \multicolumn{21}{c}{\textbf{\qwen}} \\
    \midrule
    \initial & 8.16 & - & - && 13.75 & - & - && 2.67 & - & - && 20.0 & - & - && 10.0 & - & - && 11.0 \\ 
    \hdashline\noalign{\vskip 0.4ex}
    \vanilla & 6.12 \da{2.04} & 5.1 & 7.14 && 16.25 \ua{2.5} & 5.0 & 2.5 && 0.0 \da{2.67} & 0.0 & 2.67 && 16.67 \da{3.33} & 3.33 & 6.67 && 10.0  & 0.0 & 0.0 && 10.0 \da{1.0} \\ 
    \refine & 6.12 \da{2.04} & 5.1 & 7.14 && 23.75 \ua{10.0} & 13.75 & 3.75 && 1.33 \da{1.34} & 1.33 & 2.67 && 20.0  & 10.0 & 10.0 && 3.33 \da{6.67} & 3.33 & 10.0 && 11.0  \\ 
    \tick & 0.0 \da{8.16} & 0.0 & 8.16 && 17.5 \ua{3.75} & 13.75 & 10.0 && 0.0 \da{2.67} & 0.0 & 2.67 && 10.0 \da{10.0} & 3.33 & 13.33 && 6.67 \da{3.33} & 6.67 & 10.0 && 7.0 \da{4.0} \\ 
    \reflexion & 11.22 \ua{3.06} & 9.18 & 6.12 && 21.25 \ua{7.5} & 10.0 & 2.5 && 2.67  & 2.67 & 2.67 && 13.33 \da{6.67} & 3.33 & 10.0 && 13.33 \ua{3.33} & 6.67 & \textbf{3.33} && 12.0 \ua{1.0} \\ 
    \ourmethod & 11.22 \ua{3.06} & 8.16 & 5.1 && \textbf{66.25} \textbf{\ua{52.5}} & \textbf{57.5} & 5.0 && 4.0 \ua{1.33} & 4.0 & 2.67 && 20.0  & 3.33 & \textbf{3.33} && 10.0  & 10.0 & 10.0 && 22.0 \ua{11.0} \\ 
    \ourmethodsmall & \textbf{41.84} \textbf{\ua{33.68}} & \textbf{37.76} & \textbf{4.08} && 48.75 \ua{35.0} & 40.0 & 5.0 && \textbf{10.67} \textbf{\ua{8.0}} & \textbf{10.67} & 2.67 && \textbf{23.33} \textbf{\ua{3.33}} & \textbf{13.33} & 10.0 && 13.33 \ua{3.33} & \textbf{13.33} & 10.0 && \textbf{28.0} \textbf{\ua{17.0}} \\
    \midrule
    & \multicolumn{21}{c}{\textbf{\llama}} \\
    \midrule
    \initial & 19.39 & - & - && 75.0 & - & - && 8.0 & - & - && 33.33 & - & - && 3.33 & - & - && 28.0 \\ 
    \hdashline\noalign{\vskip 0.4ex}
    \vanilla & 42.86 \ua{23.47} & 30.61 & 7.14 && 77.5 \ua{2.5} & 12.5 & 10.0 && 1.33 \da{6.67} & 0.0 & 6.67 && 36.67 \ua{3.34} & 6.67 & 3.33 && 3.33 & 0.0 & 0.0 && 32.0 \ua{4.0} \\ 
    \refine & 33.67 \ua{14.28} & 22.45 & 8.16 && 76.25 \ua{1.25} & 11.25 & 10.0 && 5.33 \da{2.67} & 4.0 & 6.67 && 40.0 \ua{6.67} & 10.0 & 3.33 && 6.67 \ua{3.33} & 3.33 & 0.0 && 32.0 \ua{4.0} \\ 
    \tick & 14.29 \da{5.1} & 12.24 & 17.35 && 71.25 \da{3.75} & 5.0 & 8.75 && 6.67 \da{1.33} & 5.33 & 6.67 && 30.0 \da{3.33} & 6.67 & 10.0 && 6.67 \ua{3.33} & 3.33 & 0.0 && 26.0 \da{2.0} \\ 
    \reflexion & 23.47 \ua{4.08} & 19.39 & 15.31 && 76.25 \ua{1.25} & 7.5 & 6.25 && 5.33 \da{2.67} & 1.33 & \textbf{4.0} && 26.67 \da{6.66} & 3.33 & 10.0 && 3.33  & 0.0 & 0.0 && 27.0 \da{1.0} \\ 
    \ourmethod & 64.29 \ua{44.9} & 48.98 & 4.08 && 98.75 \ua{23.75} & 25.0 & 1.25 && 2.67 \da{5.33} & 1.33 & 6.67 && 36.67 \ua{3.34} & 10.0 & 6.67 && 16.67 \ua{13.33} & 13.33 & 0.0 && 44.0 \ua{16.0} \\ 
    \ourmethodsmall & \textbf{100.0} \textbf{\ua{80.61}} & \textbf{80.61} & \textbf{0.0} && \textbf{100.0} \textbf{\ua{25.0}} & 25.0 & \textbf{0.0} && \textbf{38.67} \textbf{\ua{30.67}} & \textbf{38.67} & 8.0 && \textbf{46.67} \textbf{\ua{13.33}} & \textbf{20.0} & 6.67 && \textbf{23.33} \textbf{\ua{20.0}} & \textbf{20.0} & 0.0 && \textbf{62.0} \textbf{\ua{34.0}} \\
    \bottomrule
    \end{tabular}
}
\label{tab:small_delta_t1}
\end{table*}

\textbf{Datasets.}
\label{sec:setup:datasets}
We evaluate a wide range of tasks that require varying degrees of mathematical and algorithmic reasoning, focusing on problem types where traditional self-correction methods fail. The results and analysis of existing self-correction methods are provided in Appendix~\ref{sec:app:baseline-results}. We conduct experiments on \textbf{\game}~\citep{yao2023tree}, \textbf{\checkmate}~\citep{srivastava2023bigbench}, \textbf{\wordsorting}~\citep{suzgun2023bigbenchhard}, \textbf{\aimefour}~\citep{huggingfaceh42025aime2024}, and \textbf{\aimefive}~\citep{yentinglin2025aime2025}. Additional dataset details are provided in Appendix~\ref{app:sec:datasets}.

\textbf{Baselines and Comparison.}
\label{sec:setup:baselines}
We compare our methods to relevant prior approaches based on prompting, sampling, or fine-tuning a single model for both task-solving and self-correction. For prompting, we specifically compare to \textbf{\vanilla} \citep{song2025progco}, a basic iterative refinement method where the model revises its initial output. \textbf{\refine} \citep{madaan2023self-refine} is a representative approach for eliciting self-correction behaviors. \textbf{\reflexion} \citep{shinn2023reflexion} iteratively evaluates its output, generates verbal feedback, and refines its response based on this feedback. \textbf{\tick} \citep{cook2024self-tick} generates a checklist of Yes/No questions for the task and uses any unsatisfied points as feedback to improve its output. We include \textbf{\consistency} \citep{wang2022self-consistency} as a sampling-based baseline, since recent work \citep{huang2023cannot-self-correct} shows it can outperform prompting-based self-correction methods when the number of generated samples matches the number of correction steps. Among the fine-tuning based approaches, we compare to \textbf{\supercorrect} \citep{yang2024supercorrect}, which enhances small LLM reasoning by distilling thought templates and incorporating self-correction mechanisms, \textbf{\stwor} \citep{ma2025s2r} employs reinforcement learning to teach LLMs to self-verify and self-correct during inference, and \textbf{\stasc} \citep{moskvoretskii2025selftaught} focuses on self-correction for small language models through iterative fine-tuning using solely self-generated data. Further details about baselines are provided in Appendix \ref{sec:app:setup:baselines}.

\textbf{Models.}
\label{sec:setup:models}
We evaluate our approaches on a diverse set of language models, covering both large and small models. The large models include \gptfouromini \citep{openai2024gpt4omini} and \gptfouro \citep{gpt4o}, representing strong general-purpose systems.\footnote{We note that newer model families such as \textsc{GPT-4.1} and \textsc{GPT-5} were not included, since our experiments were initiated prior to their release, and a full re-evaluation with these models would have incurred considerable additional cost.} To assess performance in resource-constrained settings, we also consider smaller models such as \qwen \citep{qwen2025qwen25technicalreport} and \llama \citep{dubey2024llama}. In addition, we include specialized models focused on reasoning, such as \othreemini \citep{openai2025o3mini} and \deepseekrone \citep{guo2025deepseek}, which are explicitly designed to handle complex problem-solving tasks.

\textbf{Evaluation Protocol and Metrics.}
We evaluate performance using different accuracy metrics tailored to the specific requirements of each tasks: \textbf{Exact Match (EM)} \citep{suzgun2024meta}, which requires the output to match the ground-truth label exactly; \textbf{Soft Match (SM)} \citep{suzgun2024meta,suzgun2025dynamic}, which accepts answers containing the correct label while ignoring minor formatting differences; and \textbf{Functionally Correct (FC)} \citep{suzgun2024meta,suzgun2025dynamic}, which considers outputs correct if they satisfy task-specific constraints even when formatting or presentation differs. Following prior work \citep{suzgun2024meta,suzgun2025dynamic,huang2023cannot-self-correct}, we use EM for \checkmate, SM for \wordsorting, and FC for \game, \aimefour, and \aimefive.

To measure self-correction performance, we report and analyze the following metrics: \textbf{(1)} \textbf{Acc@ti}: accuracy at the $i$-th attempt; \textbf{(2)} \textbf{$\Delta^{i \to c}(t_{i-1}, t_i)$}: the fraction of problems that were incorrect at attempt $i-1$ but corrected at attempt $i$, capturing how many new problems self-correction solves; and \textbf{(3)} \textbf{$\Delta^{c \to i}(t_{i-1}, t_i)$}: the fraction of problems that were correct at attempt $i-1$ but become incorrect at attempt $i$, reflecting how reliably the model preserves correct answers.

\textbf{Selecting Task Abstraction.}
For our experiments, we randomly sample a single task abstraction from the set of successful cases, i.e., those in which the large model (\gptfouromini) successfully corrected the initial output. Our idea is that abstractions associated with successful corrections provide at least one concrete example of reasoning that leads to the right solution, thereby offering a useful, though not necessarily optimal, guideline for subsequent models. The selected abstraction is then reused across smaller models, serving as structured guidance to support their self-correction. This choice allows us to test whether relatively lightweight abstractions, distilled from a mid-sized model, are sufficient to enhance the performance of smaller models without relying exclusively on the largest and most costly systems.
\section{Experiments and Results}
\label{sec:results}

\begin{figure*}[!t]
    \centering
    \begin{subfigure}{\textwidth}
        \centering
        \begin{subfigure}{0.24\textwidth}
            \includegraphics[width=\linewidth]{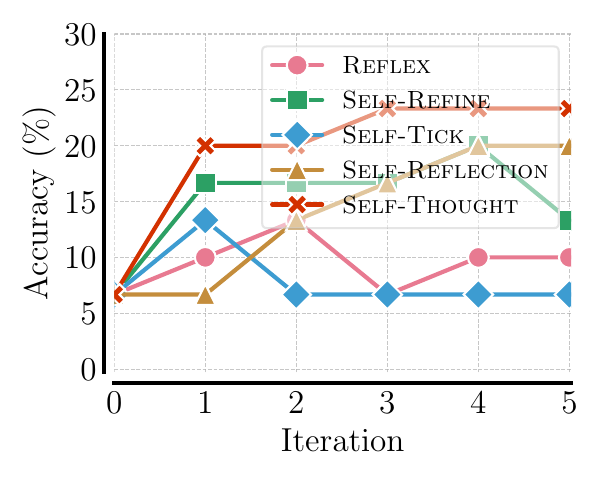}
            \subcaption{\gptfouromini}
        \end{subfigure}
        \begin{subfigure}{0.24\textwidth}
            \includegraphics[width=\linewidth]{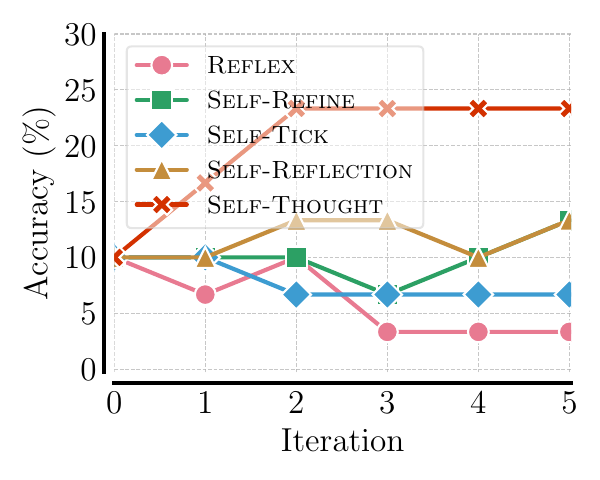}
            \subcaption{\gptfouro}
        \end{subfigure}
        \begin{subfigure}{0.24\textwidth}
            \includegraphics[width=\linewidth]{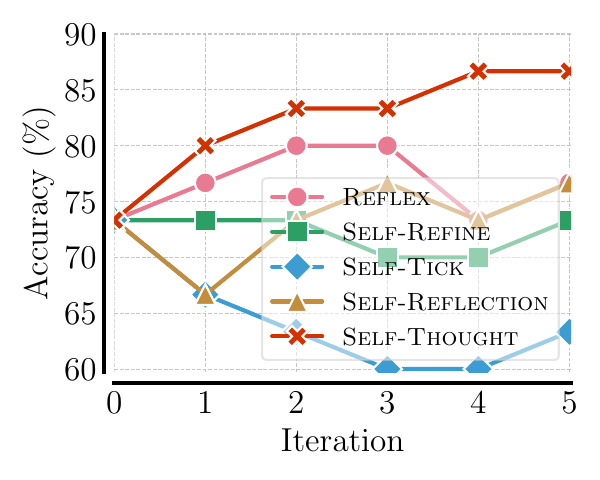}
            \subcaption{\othreemini}
        \end{subfigure}
        \begin{subfigure}{0.24\textwidth}
            \includegraphics[width=\linewidth]{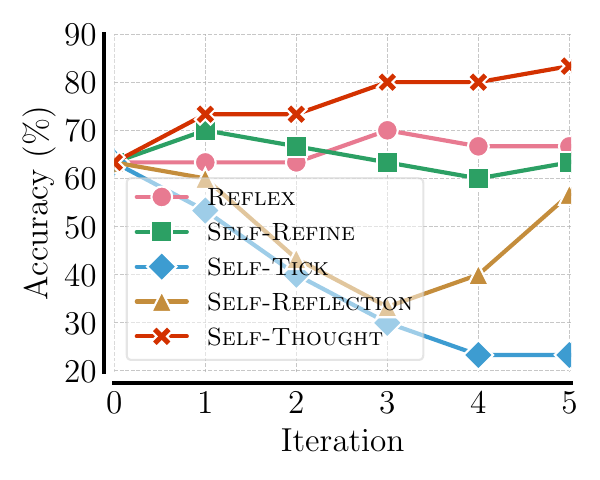}
            \subcaption{\deepseekrone}
        \end{subfigure}
    \end{subfigure}
    \begin{subfigure}{\textwidth}
        \centering
        \begin{subfigure}{0.24\textwidth}
            \includegraphics[width=\linewidth]{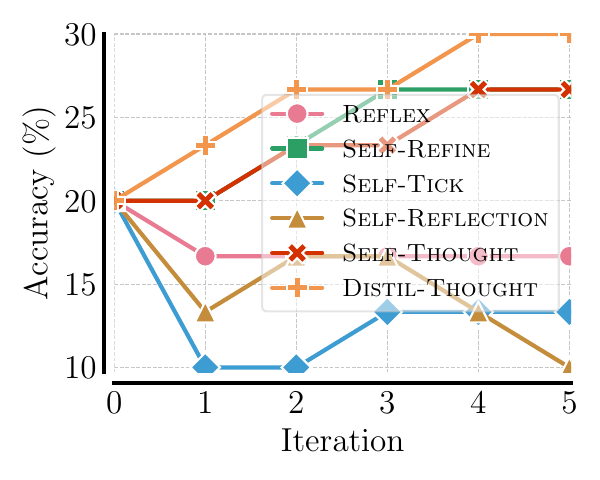}
            \subcaption{\qwen}
            \label{fig:qwen-aime2024-iter}
        \end{subfigure}
        \begin{subfigure}{0.24\textwidth}
            \includegraphics[width=\linewidth]{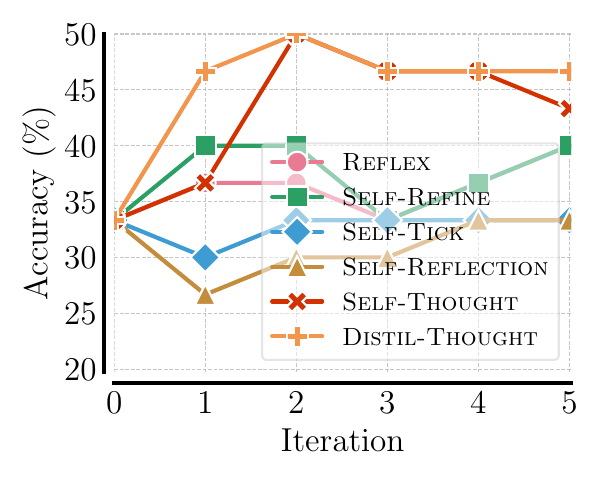}
            \subcaption{\llama}
        \end{subfigure}
        \begin{subfigure}{0.24\textwidth}
            \includegraphics[width=\linewidth]{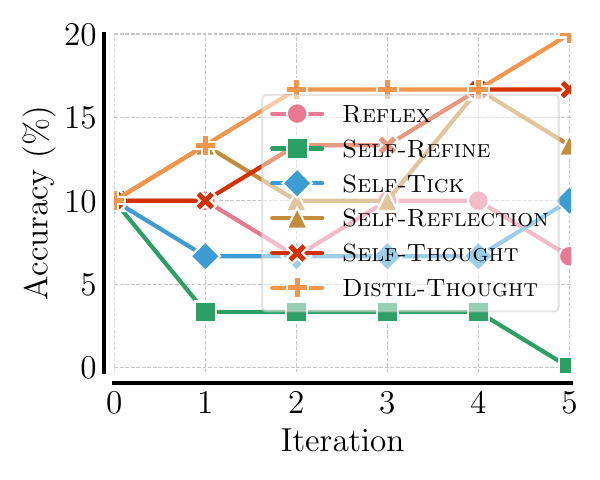}
            \subcaption{\qwen}
        \end{subfigure}
        \begin{subfigure}{0.24\textwidth}
            \includegraphics[width=\linewidth]{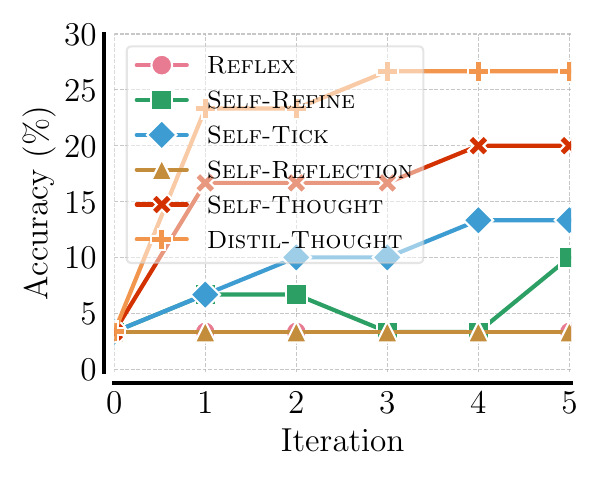}
            \subcaption{\llama}
        \end{subfigure}
    \end{subfigure}
    \caption{Accuracy over iterations with self-correction methods across models. \textbf{Top} row show results on \aimefour using large models, while \textbf{Bottom} row show results on \aimefour (subfigures e and f) and \aimefive (subfigures g and h). Please refer to Figures \ref{fig:app:iter-large} and \ref{fig:app:iter-small} in the Appendix for the iteration effect plots of other tasks.}
    \label{fig:iterations}
\end{figure*}

\textbf{Main Results.} 
\label{sec:result:main}
Table \ref{tab:large_delta_t1} shows the self-correction results across five reasoning benchmarks and four different models. We observe that \ourmethod consistently yields the highest accuracy after one round of self-correction, outperforming other intrinsic methods such as \refine, \tick, and \reflexion. We show results on iterative self-correction in the experiments, with additional analysis in Appendix~\ref{sec:app:iter-effect}. For instance, on \gptfouromini, \ourmethod improves \textbf{Acc@t1} from $38.78\%$ to $87.76\%$ on \game and from $55.0\%$ to $100.0\%$ on \wordsorting, corresponding to gains of \textcolor{ggreen}{$\uparrow$}$48.98\%$ and \textcolor{ggreen}{$\uparrow$}$45.0\%$, respectively. In comparison, \refine and \reflexion show far smaller net improvements and often increase $\Delta^{c \to i}(t_{0}, t_1)$, indicating that they mistakenly alter correct responses. This trend is most evident in reasoning-heavy tasks such as \checkmate, where \ourmethod consistently improves performance across all models, while competing approaches often reduce accuracy by altering correct answers into incorrect ones.

Looking across models, general models such as \gptfouromini and \gptfouro benefit the most from \ourmethod. On \gptfouro, it raises mean \textbf{Acc@t1} from $34.0\%$ (\initial) to $51.0\%$, with especially large gains on \checkmate and \aimefour. Reasoning models like \othreemini and \deepseekrone, which already start from stronger baselines, still see consistent positive gains, \ourmethod improves mean \textbf{Acc@t1} from $73.0\%$ to $78.0\%$ on \othreemini and from $69.0\%$ to $72.0\%$ on \deepseekrone. Importantly, the balance between $\Delta^{i \to c}(t_{0}, t_1)$ and $\Delta^{c \to i}(t_{0}, t_1)$ confirms that \ourmethod encourages conservative but effective revisions. For example, on \gptfouro, it achieves $\Delta^{i \to c}(t_{0}, t_1)$ of $13.75\%$ on \wordsorting while keeping $\Delta^{c \to i}(t_{0}, t_1)$ zero, unlike \tick, which achieves a high $\Delta^{i \to c}(t_{0}, t_1)$ but with much higher $\Delta^{c \to i}(t_{0}, t_1)$. Taken together, these results show that \ourmethod reliably improves self-correction across tasks and scales, whereas prior intrinsic methods either provide limited benefits or destabilize performance by introducing unnecessary changes.

\begin{figure*}[!t]
    \centering
    \includegraphics[width=\linewidth]{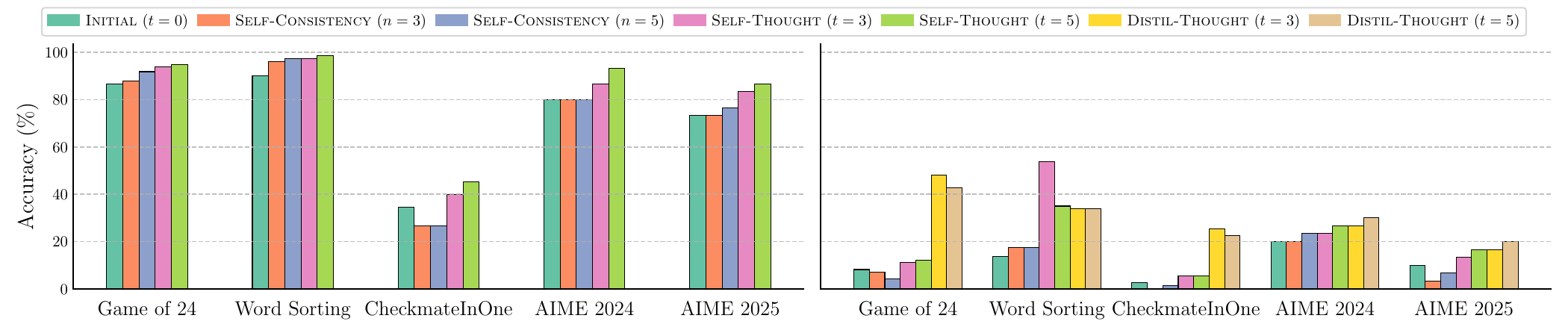}
    \caption{Comparison of \ourmethod and \ourmethodsmall with the \consistency on \othreemini and \qwen. See Figure~\ref{fig:app:self-consistency} in Appendix~\ref{sec:app:add-self-consistency} for the results on other models.}
    \label{fig:self-consistency-main}
\end{figure*}

\textbf{Results on Small Models.}
\label{sec:result:small}
Table \ref{tab:small_delta_t1} shows the results on \qwen and \llama. We observe that baseline self-correction methods often fail to deliver consistent gains and, in most cases, even degrade performance. For instance, on \qwen, \vanilla reduces mean accuracy from $20.0$ at initialization to $10.0$, while \tick drops it further to $7.0$. A similar pattern holds for \llama, where \tick lowers mean accuracy from $28.0$ to $26.0$, and \refine shows only modest improvements to $32.0$. These trends highlight that small models struggle to generate useful intrinsic feedback, often flipping correct answers into incorrect ones.

In contrast, both \ourmethod and especially \ourmethodsmall achieve substantial improvements by leveraging task abstractions distilled from larger models. On \qwen, \ourmethodsmall raises mean accuracy from $11.0$ to $28.0$, outperforming all baselines, while \ourmethod provides a moderate improvement to $22.0$. The effect is even more striking for \llama, where \ourmethodsmall boosts mean accuracy from $28.3$ to $62.0$, more than doubling performance, with particularly large gains on reasoning-heavy tasks such as \aimefour (from $33.0$ to $46.67$) and \aimefive (from $3.33$ to $23.33$). These results demonstrate that abstraction transfer offers small models a reliable pathway to self-correction, bridging the gap between their limited reasoning ability and the stronger feedback signals required for improvement.

\textbf{Effect of Iterative Correction.}
\label{sec:result:iterative}
We examine the effect of iterative correction for all tasks using different models. The results on \aimefour and \aimefive are depicted in Figure \ref{fig:iterations}, with more results on other tasks provided in Appendix \ref{sec:app:iter-effect}. We find that iterative self-correction consistently improves accuracy across models, with the largest gains in early iterations. Importantly, this early-stage improvement also translates to reduced computational cost, as models require fewer refinement cycles to reach high-quality solutions. \ourmethodsmall achieves the highest correct-flip proportions, outperforming baseline self-correction methods such as \supercorrect and \stwor by a notable margin ($10-15\%$ improvement in early rounds). While later iterations continue to provide improvements, the marginal gains diminish after 2–3 rounds. Additionally, model-only self-correction without external guidance or feedback is less effective (e.g., using external tools for feedback \citep{gou2024critic}), showing slower convergence and lower overall gains compared to approaches leveraging structured iterative updates.

\textbf{Comparison with \consistency.}
Recent work by \citet{huang2023cannot-self-correct} indicates that \consistency outperforms many existing self-correction strategies, such as multi-agent debate \citep{du2023improving,liang2024encouraging,chen2023teaching}, when applied under the same number of response samples. We therefore adopt \consistency \citep{wang2022self-consistency} as an additional baseline for comparison with our proposed methods. \consistency generates multiple candidate responses and selects the final output through majority voting. We evaluate this method using $n \in \{3,5\}$ samples, aligning with the number of self-correction iteration used in our methods. Figure~\ref{fig:self-consistency-main} shows the results across \othreemini and \qwen on five reasoning tasks. On \othreemini, \consistency shows moderate gains over the initial responses, especially on \wordsorting, \aimefour, and \aimefive. However, our methods consistently match or surpass \consistency. For example, on \checkmate, where \consistency only marginally improves performance, \ourmethod achieves a notable increase in accuracy, indicating its ability to better exploit intermediate reasoning. Similarly, on \aimefour and \aimefive, \ourmethod with $t=5$ outperforms \consistency, highlighting the effectiveness of structured task distillation in improving final answers. The difference is more noticeable on QWEN-7B. On small models, \consistency provides limited improvement, and in some cases (e.g., \checkmate), it remains close to the baseline. In contrast, both \ourmethod and \ourmethodsmall yield substantial gains. For instance, \ourmethodsmall with $t=5$ boosts accuracy by more than $20$ points on \game and by over $15$ points on \wordsorting compared to \consistency, showing the scalability of our methods even for small models. These results suggest that while \consistency can provide benefits through sampling, our approaches more effectively harness reasoning traces, yielding stronger and more stable improvements across diverse tasks.

\textbf{Comparison with Fine-Tuning Baselines.}
Table \ref{tab:main-delta-finetune} and Figure \ref{fig:main-iter-finetune} show the comparison results of our methods with fine-tuned based self-correction methods on \aimefour and \aimefour, with more results provided in Appendix \ref{sec:app:finetune}. Our methods, \ourmethod and \ourmethodsmall, achieve clear gains over all fine-tuning based baselines. \ourmethodsmall shows the strongest trends, with steady accuracy growth and the highest correct-flip rates (e.g., $13.3\%$ in iteration $1$), while \ourmethod yields consistent improvements up to $16\%$ accuracy on \aimefive. In contrast, \stwor and \stasc plateau or decline, and \supercorrect stays flat. Notably, many baselines are fine-tuned on math datasets for self-correction and benefit from training on task, whereas our methods obtain strong generalization without heavy in-domain supervision. Overall, this highlights the efficiency of our lightweight self-correction strategies compared to costly fine-tuning approaches.

\begin{table*}[!t]
\begin{minipage}{0.6\textwidth}
\caption{Results of correct and incorrect flips, comparing with fine-tuning based models on \aimefour and \aimefive. See Table \ref{tab:app:finetune} in the Appendix for detailed results on all evaluation tasks.}
\label{tab:main-delta-finetune}
\centering
\resizebox{\linewidth}{!}{%
\setlength{\tabcolsep}{0pt}
    \begin{tabular}{lccccccccccccccc}
        \toprule
        \multirow{2}{*}{\textbf{Method}} && \multicolumn{2}{c}{\bf Iteration 1} && \multicolumn{2}{c}{\bf Iteration 2} && \multicolumn{2}{c}{\bf Iteration 3} && \multicolumn{2}{c}{\bf Iteration 4} && \multicolumn{2}{c}{\bf Iteration 5} \cr
        
        \cmidrule(lr){3-4} \cmidrule(lr){6-7} \cmidrule(lr){9-10} \cmidrule(lr){12-13} \cmidrule(lr){15-16}
        
        && $\Delta^{{i \to c}}(t_0, t_1)$ & \makecell{$\Delta^{{c \to i}}(t_0, t_1)$} && \makecell{$\Delta^{{i \to c}}(t_1, t_2)$} & \makecell{$\Delta^{{c \to i}}(t_1, t_2)$} &&
        \makecell{$\Delta^{{i \to c}}(t_2, t_3)$} & \makecell{$\Delta^{{c \to i}}(t_2, t_3)$} &&
        \makecell{$\Delta^{{i \to c}}(t_3, t_4)$} & \makecell{$\Delta^{{c \to i}}(t_3, t_4)$} &&
        \makecell{$\Delta^{{i \to c}}(t_4, t_5)$} & \makecell{$\Delta^{{c \to i}}(t_4, t_5)$} \\
        \midrule
        & \multicolumn{15}{c}{\textbf{\aimefour}} \\
        \midrule
        {\large \supercorrect} && {\large 0.0} & \textbf{\large 0.0} && {\large 3.33} & {\large 0.00} && {\large 3.33} & {\large 0.00} && {\large 0.00} & {\large 0.00} && {\large 0.00} & {\large 0.00} \\ 
        {\large \stwor} && {\large 3.33} & {\large 10.00} && {\large 0.00} & {\large 0.00} && {\large 3.33} & {\large 3.33} && {\large 3.33} & {\large 3.33} && \textbf{{\large 10.00}} & {\large 3.33} \\ 
        {\large \stasc} && {\large 0.00} & \large {\large 10.00} && {\large 0.00} & {\large 0.00} && {\large 0.00} & {\large 0.00} && {\large 0.00} & {\large 0.00} && {\large 0.00} & {\large 0.00} \\ 
        {\large \ourmethod} && \large {\large 3.33} & {\large 3.33} && {\large 3.33} & {\large 0.00} && {\large 0.00} & {\large 0.00} && {\large 3.33} & {\large 0.00} && {\large 0.00} & {\large 0.00} \\ 
        {\large \ourmethodsmall} && \textbf{\large 13.33} & {\large 10.00} && {\large 3.33} & {\large 0.00} && {\large 3.33} & {\large 3.33} && \textbf{{\large 6.67}} & {\large 3.33} && {\large 0.00} & {\large 0.00} \\
        \midrule
        & \multicolumn{15}{c}{\textbf{\aimefive}} \\
        \midrule
        {\large \supercorrect} && {\large 3.33} & {\large 0.00} && {\large 6.67} & {\large 0.00} && {\large 0.00} & {\large 0.00} && {\large 3.33} & {\large 0.00} && {\large 3.33} & {\large 3.33} \\ 
        {\large \stwor} && {\large 3.33} & {\large 6.67} && {\large 3.33} & {\large 6.67} && {\large 0.00} & {\large 3.33} && {\large 0.00} & {\large 0.00} && \textbf{{\large 6.67}} & {\large 0.00} \\ 
        {\large \stasc} && {\large 0.00} & {\large 0.00} && {\large 0.00} & {\large 0.00} && {\large 0.00} & {\large 0.00} && {\large 0.00} & {\large 0.00} && {\large 0.00} & {\large 0.00} \\ 
        {\large \ourmethod} && {\large 10.00} & {\large 10.00} && {\large 3.33} & {\large 0.00} && {\large 3.33} & {\large 3.33} && {\large 3.33} & {\large 0.00} && {\large 0.00} & {\large 0.00} \\ 
        {\large \ourmethodsmall} && \textbf{{\large 13.33}} & {\large 10.00} && {\large 3.33} & {\large 0.00} && \textbf{{\large 6.67}} & {\large 6.67} && {\large 0.00} & {\large 0.00} && {\large 3.33} & {\large 0.00} \\
        \bottomrule
    \end{tabular}
}%
\end{minipage}
\hfill
\begin{minipage}{0.39\linewidth}
  \centering
  \includegraphics[scale=0.33]{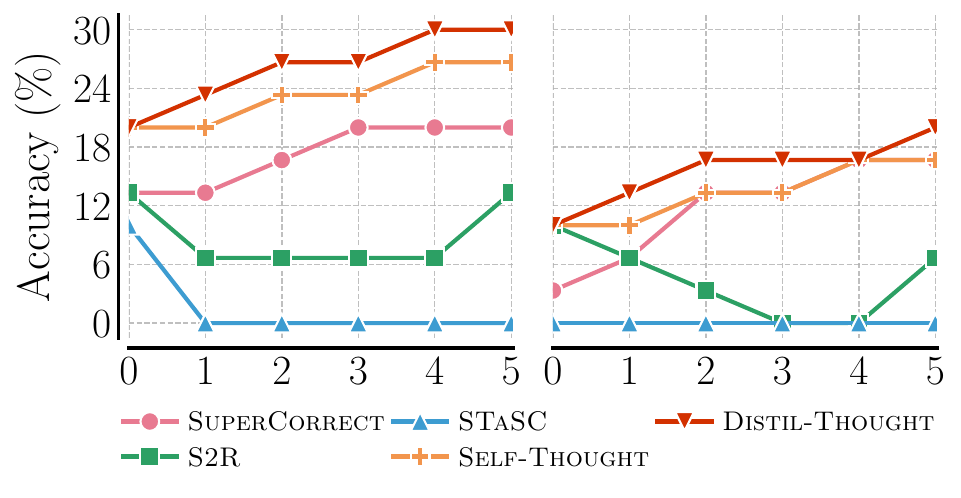}
  \captionof{figure}{Accuracy over iterations on (\textbf{Right}) \aimefour and (\textbf{Left}) \aimefive.}
  \label{fig:main-iter-finetune}
\end{minipage}
\hfill
\end{table*}

\begin{figure}
    \begin{subfigure}[b]{0.49\linewidth}
        \centering
        \includegraphics[width=.95\textwidth]{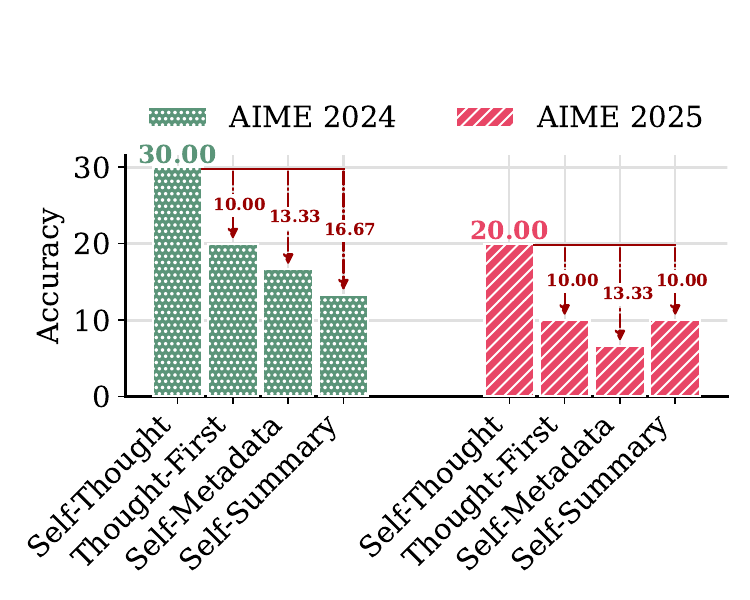}
        \caption{\gptfouromini}
        \label{fig:ablation_aime_2024}
    \end{subfigure}
    \begin{subfigure}[b]{0.49\linewidth}
        \centering
        \includegraphics[width=.95\textwidth]{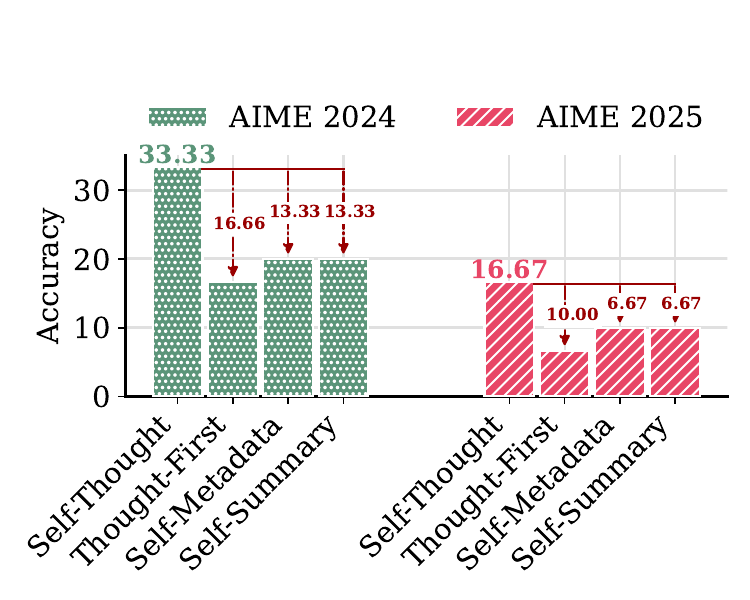}
        \caption{\gptfouro}
        \label{fig:ablation_aime_2025}
    \end{subfigure}
    \caption{Performance comparison of \ourmethod and ablated variants for \aimefour and \aimefive.}
    \label{fig:ablation}
\end{figure}

\textbf{Ablations.}
Figure \ref{fig:ablation} presents an ablation study comparing \ourmethod with three alternative designs. \ourmethod consistently achieves the highest accuracy on both benchmarks, outperforming variants that modify or remove the explicit task-distillation step. In particular, \thought, which applies the task-abstraction prompt directly to generate the final answer without distilling and reusing the abstraction as an intermediate representation, exhibits a clear performance drop relative to \ourmethod. \metadata, which extracts shallow task descriptors, and \summary, which compresses the problem into a natural-language summary, also underperform the full method. Notably, \summary performs significantly worse than \ourmethod despite using a comparable multi-step prompting budget, indicating that the gains of \ourmethod cannot be attributed solely to multi-step inference or summarization, but rather to the explicit abstraction of task structure. Full results are provided in Appendix~\ref{sec:app:ablation}.
\section{Related Work}
\label{sec:relatedwork}


\textbf{Intrinsic Self-Correction.}
Intrinsic self-correction seeks to enable models to generate and act on their own feedback during inference, without relying on external signals or additional training. Several methods have been proposed to ask LLMs to critique their initial responses and then attempt refinements \citep{kim2023language,shinn2023reflexion,madaan2023self-refine}, but recent studies highlight their limitations, where models often fail to detect reasoning errors reliably, and performance sometimes degrades when self-reflection is applied naively
\citep{huang2023cannot-self-correct,tyen2024llms,kamoi2024can}. More structured variants \citep{shinn2023reflexion,zelikman2022star} introduce iterative critique or self-distillation, showing that verbal self-feedback can improve output quality, yet these techniques remain brittle and mostly applicable to large models with strong baseline reasoning abilities. In contrast, our method introduces an explicit task abstraction step before refinement, rather than relying on unstructured critique; the model distills the input into a structured template capturing key variables and constraints. This abstraction not only improves the intrinsic correction of large models by grounding refinements in a clearer task representation, but also enables transferability---smaller models, which struggle to generate useful feedback themselves, can leverage abstractions produced by larger models as structured guidance for more reliable correction.



\textbf{Source of Feedback.}
\label{sec:relatedwork:feedback}
Feedback is crucial to improve LLM output, with humans traditionally providing corrective signals \citep{tandon2021interscript,elgohary2021nl,bai2022training}. Since human feedback is costly, alternative sources such as scalar reward functions \citep{bai2022training,liu2022rainier,welleck2022generating}, external tools like compilers or search engines \citep{yasunaga2020graph,chen2024teaching,yu2023improving}, and domain-specific knowledge bases \citep{schick2023peer} have been used. More recently, LLMs themselves have been employed to generate feedback \citep{kim2023language,madaan2023self-refine,cook2024self-tick}, allowing models to iteratively refine their own outputs. However, without structured or verified guidance, LLMs often struggle to correct deeper reasoning errors \citep{huang2023cannot-self-correct}. In contrast, our method provides feedback through explicit task abstractions distilled from the input, offering structured guidance. For large models, these abstractions ground refinements in a clearer task representation, while for smaller models they serve as reusable templates from stronger models, enabling more reliable self-correction than unverified self-critiques.
\section{Conclusion}
\label{sec:conclusion}
We introduce \ourmethod and \ourmethodsmall, two complementary approaches for enhancing self-correction in language models. \ourmethod empowers models to refine their reasoning through structured introspection, while \ourmethodsmall leverages distilled guidance from a larger model to improve correction efficiency and stability. Together, they offer a lightweight yet effective alternative to existing prompting- and fine-tuning-based methods. Our experiments across diverse reasoning benchmarks demonstrate consistent improvements in correction accuracy and flip reliability, especially in early iterations, highlighting both the generality and robustness of our framework. Beyond performance, our analysis sheds light on the limitations of current self-correction techniques and underscores the value of structured reasoning templates for building more trustworthy systems. 

\section*{Impact Statement}
This work contributes to a clearer understanding of how intermediate representations influence reasoning and self-correction in large language models. By isolating task abstraction from prompt strength and summarization, our study provides empirical evidence that explicitly distilled task representations can serve as effective control interfaces for downstream reasoning. These findings are relevant for the design and evaluation of LLM-based systems that rely on multi-step reasoning, benchmarking, and model self-improvement. More broadly, this work informs ongoing research on the structure of reasoning processes in language models and highlights the importance of evaluating not only final answers but also the representations that guide them.

\bibliography{paper}

@inproceedings{wang2022self-consistency,
title={Self-Consistency Improves Chain of Thought Reasoning in Language Models},
author={Xuezhi Wang and Jason Wei and Dale Schuurmans and Quoc V Le and Ed H. Chi and Sharan Narang and Aakanksha Chowdhery and Denny Zhou},
booktitle={The Eleventh International Conference on Learning Representations },
year={2023},
url={https://openreview.net/forum?id=1PL1NIMMrw}
}

@article{huang2023cannot-self-correct,
  title={Large language models cannot self-correct reasoning yet},
  author={Huang, Jie and Chen, Xinyun and Mishra, Swaroop and Zheng, Huaixiu Steven and Yu, Adams Wei and Song, Xinying and Zhou, Denny},
  journal={arXiv preprint arXiv:2310.01798},
  year={2023}
}

@inproceedings{yang2024supercorrect,
title={SuperCorrect: Advancing Small {LLM} Reasoning with Thought Template Distillation and Self-Correction},
author={Ling Yang and Zhaochen Yu and Tianjun Zhang and Minkai Xu and Joseph E. Gonzalez and Bin CUI and Shuicheng YAN},
booktitle={The Thirteenth International Conference on Learning Representations},
year={2025},
url={https://openreview.net/forum?id=PyjZO7oSw2}
}

@article{song2025progco,
  title={Progco: Program helps self-correction of large language models},
  author={Song, Xiaoshuai and Wu, Yanan and Wang, Weixun and Liu, Jiaheng and Su, Wenbo and Zheng, Bo},
  journal={arXiv preprint arXiv:2501.01264},
  year={2025}
}

@article{madaan2023self-refine,
  title={Self-refine: Iterative refinement with self-feedback},
  author={Madaan, Aman and Tandon, Niket and Gupta, Prakhar and Hallinan, Skyler and Gao, Luyu and Wiegreffe, Sarah and Alon, Uri and Dziri, Nouha and Prabhumoye, Shrimai and Yang, Yiming and others},
  journal={Advances in Neural Information Processing Systems},
  volume={36},
  pages={46534--46594},
  year={2023}
}

@article{shinn2023reflexion,
  title={Reflexion: Language agents with verbal reinforcement learning},
  author={Shinn, Noah and Cassano, Federico and Gopinath, Ashwin and Narasimhan, Karthik and Yao, Shunyu},
  journal={Advances in Neural Information Processing Systems},
  volume={36},
  pages={8634--8652},
  year={2023}
}

@inproceedings{cook2024self-tick,
title={{TICK}ing All the Boxes: Generated Checklists Improve {LLM} Evaluation and Generation},
author={Jonathan Cook and Tim Rockt{\"a}schel and Jakob Nicolaus Foerster and Dennis Aumiller and Alex Wang},
booktitle={Language Gamification - NeurIPS 2024 Workshop},
year={2024},
url={https://openreview.net/forum?id=Q3y6QhOUnI}
}

@article{yang2024buffer,
  title={Buffer of thoughts: Thought-augmented reasoning with large language models},
  author={Yang, Ling and Yu, Zhaochen and Zhang, Tianjun and Cao, Shiyi and Xu, Minkai and Zhang, Wentao and Gonzalez, Joseph E and Cui, Bin},
  journal={Advances in Neural Information Processing Systems},
  volume={37},
  pages={113519--113544},
  year={2024}
}

@inproceedings{yao2023tree,
title={Tree of Thoughts: Deliberate Problem Solving with Large Language Models},
author={Shunyu Yao and Dian Yu and Jeffrey Zhao and Izhak Shafran and Thomas L. Griffiths and Yuan Cao and Karthik R Narasimhan},
booktitle={Thirty-seventh Conference on Neural Information Processing Systems},
year={2023},
url={https://openreview.net/forum?id=5Xc1ecxO1h}
}

@article{srivastava2023bigbench,
title={Beyond the Imitation Game: Quantifying and extrapolating the capabilities of language models},
author={Aarohi Srivastava and Abhinav Rastogi and Abhishek Rao and Abu Awal Md Shoeb and Abubakar Abid and Adam Fisch and Adam R. Brown and Adam Santoro and Aditya Gupta and Adri{\`a} Garriga-Alonso and Agnieszka Kluska and Aitor Lewkowycz and Akshat Agarwal and Alethea Power and Alex Ray and Alex Warstadt and Alexander W. Kocurek and Ali Safaya and Ali Tazarv and Alice Xiang and Alicia Parrish and Allen Nie and Aman Hussain and Amanda Askell and Amanda Dsouza and Ambrose Slone and Ameet Rahane and Anantharaman S. Iyer and Anders Johan Andreassen and Andrea Madotto and Andrea Santilli and Andreas Stuhlm{\"u}ller and Andrew M. Dai and Andrew La and Andrew Kyle Lampinen and Andy Zou and Angela Jiang and Angelica Chen and Anh Vuong and Animesh Gupta and Anna Gottardi and Antonio Norelli and Anu Venkatesh and Arash Gholamidavoodi and Arfa Tabassum and Arul Menezes and Arun Kirubarajan and Asher Mullokandov and Ashish Sabharwal and Austin Herrick and Avia Efrat and Aykut Erdem and Ayla Karaka{\c{s}} and B. Ryan Roberts and Bao Sheng Loe and Barret Zoph and Bart{\l}omiej Bojanowski and Batuhan {\"O}zyurt and Behnam Hedayatnia and Behnam Neyshabur and Benjamin Inden and Benno Stein and Berk Ekmekci and Bill Yuchen Lin and Blake Howald and Bryan Orinion and Cameron Diao and Cameron Dour and Catherine Stinson and Cedrick Argueta and Cesar Ferri and Chandan Singh and Charles Rathkopf and Chenlin Meng and Chitta Baral and Chiyu Wu and Chris Callison-Burch and Christopher Waites and Christian Voigt and Christopher D Manning and Christopher Potts and Cindy Ramirez and Clara E. Rivera and Clemencia Siro and Colin Raffel and Courtney Ashcraft and Cristina Garbacea and Damien Sileo and Dan Garrette and Dan Hendrycks and Dan Kilman and Dan Roth and C. Daniel Freeman and Daniel Khashabi and Daniel Levy and Daniel Mosegu{\'\i} Gonz{\'a}lez and Danielle Perszyk and Danny Hernandez and Danqi Chen and Daphne Ippolito and Dar Gilboa and David Dohan and David Drakard and David Jurgens and Debajyoti Datta and Deep Ganguli and Denis Emelin and Denis Kleyko and Deniz Yuret and Derek Chen and Derek Tam and Dieuwke Hupkes and Diganta Misra and Dilyar Buzan and Dimitri Coelho Mollo and Diyi Yang and Dong-Ho Lee and Dylan Schrader and Ekaterina Shutova and Ekin Dogus Cubuk and Elad Segal and Eleanor Hagerman and Elizabeth Barnes and Elizabeth Donoway and Ellie Pavlick and Emanuele Rodol{\`a} and Emma Lam and Eric Chu and Eric Tang and Erkut Erdem and Ernie Chang and Ethan A Chi and Ethan Dyer and Ethan Jerzak and Ethan Kim and Eunice Engefu Manyasi and Evgenii Zheltonozhskii and Fanyue Xia and Fatemeh Siar and Fernando Mart{\'\i}nez-Plumed and Francesca Happ{\'e} and Francois Chollet and Frieda Rong and Gaurav Mishra and Genta Indra Winata and Gerard de Melo and Germ{\`a}n Kruszewski and Giambattista Parascandolo and Giorgio Mariani and Gloria Xinyue Wang and Gonzalo Jaimovitch-Lopez and Gregor Betz and Guy Gur-Ari and Hana Galijasevic and Hannah Kim and Hannah Rashkin and Hannaneh Hajishirzi and Harsh Mehta and Hayden Bogar and Henry Francis Anthony Shevlin and Hinrich Schuetze and Hiromu Yakura and Hongming Zhang and Hugh Mee Wong and Ian Ng and Isaac Noble and Jaap Jumelet and Jack Geissinger and Jackson Kernion and Jacob Hilton and Jaehoon Lee and Jaime Fern{\'a}ndez Fisac and James B Simon and James Koppel and James Zheng and James Zou and Jan Kocon and Jana Thompson and Janelle Wingfield and Jared Kaplan and Jarema Radom and Jascha Sohl-Dickstein and Jason Phang and Jason Wei and Jason Yosinski and Jekaterina Novikova and Jelle Bosscher and Jennifer Marsh and Jeremy Kim and Jeroen Taal and Jesse Engel and Jesujoba Alabi and Jiacheng Xu and Jiaming Song and Jillian Tang and Joan Waweru and John Burden and John Miller and John U. Balis and Jonathan Batchelder and Jonathan Berant and J{\"o}rg Frohberg and Jos Rozen and Jose Hernandez-Orallo and Joseph Boudeman and Joseph Guerr and Joseph Jones and Joshua B. Tenenbaum and Joshua S. Rule and Joyce Chua and Kamil Kanclerz and Karen Livescu and Karl Krauth and Karthik Gopalakrishnan and Katerina Ignatyeva and Katja Markert and Kaustubh Dhole and Kevin Gimpel and Kevin Omondi and Kory Wallace Mathewson and Kristen Chiafullo and Ksenia Shkaruta and Kumar Shridhar and Kyle McDonell and Kyle Richardson and Laria Reynolds and Leo Gao and Li Zhang and Liam Dugan and Lianhui Qin and Lidia Contreras-Ochando and Louis-Philippe Morency and Luca Moschella and Lucas Lam and Lucy Noble and Ludwig Schmidt and Luheng He and Luis Oliveros-Col{\'o}n and Luke Metz and L{\"u}tfi Kerem Senel and Maarten Bosma and Maarten Sap and Maartje Ter Hoeve and Maheen Farooqi and Manaal Faruqui and Mantas Mazeika and Marco Baturan and Marco Marelli and Marco Maru and Maria Jose Ramirez-Quintana and Marie Tolkiehn and Mario Giulianelli and Martha Lewis and Martin Potthast and Matthew L Leavitt and Matthias Hagen and M{\'a}ty{\'a}s Schubert and Medina Orduna Baitemirova and Melody Arnaud and Melvin McElrath and Michael Andrew Yee and Michael Cohen and Michael Gu and Michael Ivanitskiy and Michael Starritt and Michael Strube and Michal Swedrowski and Michele Bevilacqua and Michihiro Yasunaga and Mihir Kale and Mike Cain and Mimee Xu and Mirac Suzgun and Mitch Walker and Mo Tiwari and Mohit Bansal and Moin Aminnaseri and Mor Geva and Mozhdeh Gheini and Mukund Varma T and Nanyun Peng and Nathan Andrew Chi and Nayeon Lee and Neta Gur-Ari Krakover and Nicholas Cameron and Nicholas Roberts and Nick Doiron and Nicole Martinez and Nikita Nangia and Niklas Deckers and Niklas Muennighoff and Nitish Shirish Keskar and Niveditha S. Iyer and Noah Constant and Noah Fiedel and Nuan Wen and Oliver Zhang and Omar Agha and Omar Elbaghdadi and Omer Levy and Owain Evans and Pablo Antonio Moreno Casares and Parth Doshi and Pascale Fung and Paul Pu Liang and Paul Vicol and Pegah Alipoormolabashi and Peiyuan Liao and Percy Liang and Peter W Chang and Peter Eckersley and Phu Mon Htut and Pinyu Hwang and Piotr Mi{\l}kowski and Piyush Patil and Pouya Pezeshkpour and Priti Oli and Qiaozhu Mei and Qing Lyu and Qinlang Chen and Rabin Banjade and Rachel Etta Rudolph and Raefer Gabriel and Rahel Habacker and Ramon Risco and Rapha{\"e}l Milli{\`e}re and Rhythm Garg and Richard Barnes and Rif A. Saurous and Riku Arakawa and Robbe Raymaekers and Robert Frank and Rohan Sikand and Roman Novak and Roman Sitelew and Ronan Le Bras and Rosanne Liu and Rowan Jacobs and Rui Zhang and Russ Salakhutdinov and Ryan Andrew Chi and Seungjae Ryan Lee and Ryan Stovall and Ryan Teehan and Rylan Yang and Sahib Singh and Saif M. Mohammad and Sajant Anand and Sam Dillavou and Sam Shleifer and Sam Wiseman and Samuel Gruetter and Samuel R. Bowman and Samuel Stern Schoenholz and Sanghyun Han and Sanjeev Kwatra and Sarah A. Rous and Sarik Ghazarian and Sayan Ghosh and Sean Casey and Sebastian Bischoff and Sebastian Gehrmann and Sebastian Schuster and Sepideh Sadeghi and Shadi Hamdan and Sharon Zhou and Shashank Srivastava and Sherry Shi and Shikhar Singh and Shima Asaadi and Shixiang Shane Gu and Shubh Pachchigar and Shubham Toshniwal and Shyam Upadhyay and Shyamolima Shammie Debnath and Siamak Shakeri and Simon Thormeyer and Simone Melzi and Siva Reddy and Sneha Priscilla Makini and Soo-Hwan Lee and Spencer Torene and Sriharsha Hatwar and Stanislas Dehaene and Stefan Divic and Stefano Ermon and Stella Biderman and Stephanie Lin and Stephen Prasad and Steven Piantadosi and Stuart Shieber and Summer Misherghi and Svetlana Kiritchenko and Swaroop Mishra and Tal Linzen and Tal Schuster and Tao Li and Tao Yu and Tariq Ali and Tatsunori Hashimoto and Te-Lin Wu and Th{\'e}o Desbordes and Theodore Rothschild and Thomas Phan and Tianle Wang and Tiberius Nkinyili and Timo Schick and Timofei Kornev and Titus Tunduny and Tobias Gerstenberg and Trenton Chang and Trishala Neeraj and Tushar Khot and Tyler Shultz and Uri Shaham and Vedant Misra and Vera Demberg and Victoria Nyamai and Vikas Raunak and Vinay Venkatesh Ramasesh and vinay uday prabhu and Vishakh Padmakumar and Vivek Srikumar and William Fedus and William Saunders and William Zhang and Wout Vossen and Xiang Ren and Xiaoyu Tong and Xinran Zhao and Xinyi Wu and Xudong Shen and Yadollah Yaghoobzadeh and Yair Lakretz and Yangqiu Song and Yasaman Bahri and Yejin Choi and Yichi Yang and Sophie Hao and Yifu Chen and Yonatan Belinkov and Yu Hou and Yufang Hou and Yuntao Bai and Zachary Seid and Zhuoye Zhao and Zijian Wang and Zijie J. Wang and Zirui Wang and Ziyi Wu},
journal={Transactions on Machine Learning Research},
issn={2835-8856},
year={2023},
url={https://openreview.net/forum?id=uyTL5Bvosj},
}

@inproceedings{suzgun2023bigbenchhard,
    title = "Challenging {BIG}-Bench Tasks and Whether Chain-of-Thought Can Solve Them",
    author = {Suzgun, Mirac  and
      Scales, Nathan  and
      Sch{\"a}rli, Nathanael  and
      Gehrmann, Sebastian  and
      Tay, Yi  and
      Chung, Hyung Won  and
      Chowdhery, Aakanksha  and
      Le, Quoc  and
      Chi, Ed  and
      Zhou, Denny  and
      Wei, Jason},
    booktitle = "Findings of the Association for Computational Linguistics: ACL 2023",
    year = "2023",
    publisher = "Association for Computational Linguistics",
    url = "https://aclanthology.org/2023.findings-acl.824/",
    doi = "10.18653/v1/2023.findings-acl.824",
    pages = "13003--13051",
}

@inproceedings{gou2024critic,
title={{CRITIC}: Large Language Models Can Self-Correct with Tool-Interactive Critiquing},
author={Zhibin Gou and Zhihong Shao and Yeyun Gong and yelong shen and Yujiu Yang and Nan Duan and Weizhu Chen},
booktitle={The Twelfth International Conference on Learning Representations},
year={2024},
url={https://openreview.net/forum?id=Sx038qxjek}
}

@article{suzgun2025dynamic,
  title={Dynamic cheatsheet: Test-time learning with adaptive memory},
  author={Suzgun, Mirac and Yuksekgonul, Mert and Bianchi, Federico and Jurafsky, Dan and Zou, James},
  journal={arXiv preprint arXiv:2504.07952},
  year={2025}
}

@article{suzgun2024meta,
  title={Meta-prompting: Enhancing language models with task-agnostic scaffolding},
  author={Suzgun, Mirac and Kalai, Adam Tauman},
  journal={arXiv preprint arXiv:2401.12954},
  year={2024}
}

@article{tandon2021interscript,
  title={Interscript: A dataset for interactive learning of scripts through error feedback},
  author={Tandon, Niket and Madaan, Aman and Clark, Peter and Sakaguchi, Keisuke and Yang, Yiming},
  journal={arXiv preprint arXiv:2112.07867},
  year={2021}
}

@article{elgohary2021nl,
  title={NL-EDIT: Correcting semantic parse errors through natural language interaction},
  author={Elgohary, Ahmed and Meek, Christopher and Richardson, Matthew and Fourney, Adam and Ramos, Gonzalo and Awadallah, Ahmed Hassan},
  journal={arXiv preprint arXiv:2103.14540},
  year={2021}
}

@article{bai2022training,
  title={Training a helpful and harmless assistant with reinforcement learning from human feedback},
  author={Bai, Yuntao and Jones, Andy and Ndousse, Kamal and Askell, Amanda and Chen, Anna and DasSarma, Nova and Drain, Dawn and Fort, Stanislav and Ganguli, Deep and Henighan, Tom and others},
  journal={arXiv preprint arXiv:2204.05862},
  year={2022}
}

@article{liu2022rainier,
  title={Rainier: Reinforced knowledge introspector for commonsense question answering},
  author={Liu, Jiacheng and Hallinan, Skyler and Lu, Ximing and He, Pengfei and Welleck, Sean and Hajishirzi, Hannaneh and Choi, Yejin},
  journal={arXiv preprint arXiv:2210.03078},
  year={2022}
}

@article{welleck2022generating,
  title={Generating sequences by learning to self-correct},
  author={Welleck, Sean and Lu, Ximing and West, Peter and Brahman, Faeze and Shen, Tianxiao and Khashabi, Daniel and Choi, Yejin},
  journal={arXiv preprint arXiv:2211.00053},
  year={2022}
}

@inproceedings{yasunaga2020graph,
  title={Graph-based, self-supervised program repair from diagnostic feedback},
  author={Yasunaga, Michihiro and Liang, Percy},
  booktitle={International Conference on Machine Learning},
  pages={10799--10808},
  year={2020},
  organization={PMLR}
}

@inproceedings{
chen2024teaching,
title={Teaching Large Language Models to Self-Debug},
author={Xinyun Chen and Maxwell Lin and Nathanael Sch{\"a}rli and Denny Zhou},
booktitle={The Twelfth International Conference on Learning Representations},
year={2024},
url={https://openreview.net/forum?id=KuPixIqPiq}
}

@article{yu2023improving,
  title={Improving language models via plug-and-play retrieval feedback},
  author={Yu, Wenhao and Zhang, Zhihan and Liang, Zhenwen and Jiang, Meng and Sabharwal, Ashish},
  journal={arXiv preprint arXiv:2305.14002},
  year={2023}
}

@inproceedings{schick2023peer,
title={{PEER}: A Collaborative Language Model},
author={Timo Schick and Jane A. Yu and Zhengbao Jiang and Fabio Petroni and Patrick Lewis and Gautier Izacard and Qingfei You and Christoforos Nalmpantis and Edouard Grave and Sebastian Riedel},
booktitle={The Eleventh International Conference on Learning Representations },
year={2023},
url={https://openreview.net/forum?id=KbYevcLjnc}
}

@article{kim2023language,
  title={Language models can solve computer tasks},
  author={Kim, Geunwoo and Baldi, Pierre and McAleer, Stephen},
  journal={Advances in Neural Information Processing Systems},
  volume={36},
  pages={39648--39677},
  year={2023}
}

@inproceedings{tyen2024llms,
  title={LLMs cannot find reasoning errors, but can correct them given the error location},
  author={Tyen, Gladys and Mansoor, Hassan and C{\u{a}}rbune, Victor and Chen, Yuanzhu Peter and Mak, Tony},
  booktitle={Findings of the Association for Computational Linguistics ACL 2024},
  pages={13894--13908},
  year={2024}
}

@article{kamoi2024can,
  title={When can llms actually correct their own mistakes? a critical survey of self-correction of llms},
  author={Kamoi, Ryo and Zhang, Yusen and Zhang, Nan and Han, Jiawei and Zhang, Rui},
  journal={Transactions of the Association for Computational Linguistics},
  volume={12},
  pages={1417--1440},
  year={2024}
}

@inproceedings{zelikman2022star,
title={{ST}aR: Bootstrapping Reasoning With Reasoning},
author={Eric Zelikman and Yuhuai Wu and Jesse Mu and Noah Goodman},
booktitle={Advances in Neural Information Processing Systems},
editor={Alice H. Oh and Alekh Agarwal and Danielle Belgrave and Kyunghyun Cho},
year={2022},
url={https://openreview.net/forum?id=_3ELRdg2sgI}
}

@article{brown2020language,
  title={Language models are few-shot learners},
  author={Brown, Tom and Mann, Benjamin and Ryder, Nick and Subbiah, Melanie and Kaplan, Jared D and Dhariwal, Prafulla and Neelakantan, Arvind and Shyam, Pranav and Sastry, Girish and Askell, Amanda and others},
  journal={Advances in neural information processing systems},
  volume={33},
  pages={1877--1901},
  year={2020}
}

@article{chang2024survey,
  title={A survey on evaluation of large language models},
  author={Chang, Yupeng and Wang, Xu and Wang, Jindong and Wu, Yuan and Yang, Linyi and Zhu, Kaijie and Chen, Hao and Yi, Xiaoyuan and Wang, Cunxiang and Wang, Yidong and others},
  journal={ACM transactions on intelligent systems and technology},
  volume={15},
  number={3},
  pages={1--45},
  year={2024},
  publisher={ACM New York, NY}
}

@article{kojima2022large,
  title={Large language models are zero-shot reasoners},
  author={Kojima, Takeshi and Gu, Shixiang Shane and Reid, Machel and Matsuo, Yutaka and Iwasawa, Yusuke},
  journal={Advances in neural information processing systems},
  volume={35},
  pages={22199--22213},
  year={2022}
}

@inproceedings{maynez2020faithfulness,
  title={On Faithfulness and Factuality in Abstractive Summarization},
  author={Maynez, Joshua and Narayan, Shashi and Bohnet, Bernd and McDonald, Ryan},
  booktitle={Proceedings of the 58th Annual Meeting of the Association for Computational Linguistics},
  pages={1906--1919},
  year={2020}
}

@inproceedings{gehman2020realtoxicityprompts,
  title={RealToxicityPrompts: Evaluating Neural Toxic Degeneration in Language Models},
  author={Gehman, Samuel and Gururangan, Suchin and Sap, Maarten and Choi, Yejin and Smith, Noah A},
  booktitle={Findings of the Association for Computational Linguistics: EMNLP 2020},
  pages={3356--3369},
  year={2020}
}

@article{alkaissi2023artificial,
  title={Artificial hallucinations in ChatGPT: implications in scientific writing},
  author={Alkaissi, Hussam and McFarlane, Samy I},
  journal={Cureus},
  volume={15},
  number={2},
  year={2023},
  publisher={Cureus}
}

@article{yuan2023well,
  title={How well do large language models perform in arithmetic tasks?},
  author={Yuan, Zheng and Yuan, Hongyi and Tan, Chuanqi and Wang, Wei and Huang, Songfang},
  journal={arXiv preprint arXiv:2304.02015},
  year={2023}
}

@article{liu2024large,
  title={Large language models have intrinsic self-correction ability},
  author={Liu, Dancheng and Nassereldine, Amir and Yang, Ziming and Xu, Chenhui and Hu, Yuting and Li, Jiajie and Kumar, Utkarsh and Lee, Changjae and Qin, Ruiyang and Shi, Yiyu and others},
  journal={arXiv preprint arXiv:2406.15673},
  year={2024}
}

@article{belcak2025small,
  title={Small Language Models are the Future of Agentic AI},
  author={Belcak, Peter and Heinrich, Greg and Diao, Shizhe and Fu, Yonggan and Dong, Xin and Muralidharan, Saurav and Lin, Yingyan Celine and Molchanov, Pavlo},
  journal={arXiv preprint arXiv:2506.02153},
  year={2025}
}

@inproceedings{moskvoretskii2025selftaught,
title={Self-Taught Self-Correction for Small Language Models},
author={Viktor Moskvoretskii and Chris Biemann and Irina Nikishina},
booktitle={Scaling Self-Improving Foundation Models without Human Supervision - ICLR 2025 Workshop},
year={2025},
url={https://openreview.net/forum?id=vKE0StOgZ6}
}

@inproceedings{du2023improving,
  title={Improving factuality and reasoning in language models through multiagent debate},
  author={Du, Yilun and Li, Shuang and Torralba, Antonio and Tenenbaum, Joshua B and Mordatch, Igor},
  booktitle={Forty-first International Conference on Machine Learning},
  year={2023}
}

@inproceedings{liang2024encouraging,
  title={Encouraging Divergent Thinking in Large Language Models through Multi-Agent Debate},
  author={Liang, Tian and He, Zhiwei and Jiao, Wenxiang and Wang, Xing and Wang, Yan and Wang, Rui and Yang, Yujiu and Shi, Shuming and Tu, Zhaopeng},
  booktitle={Proceedings of the 2024 Conference on Empirical Methods in Natural Language Processing},
  pages={17889--17904},
  year={2024}
}

@inproceedings{chen2023teaching,
  title={Teaching Large Language Models to Self-Debug},
  author={Chen, Xinyun and Lin, Maxwell and Schaerli, Nathanael and Zhou, Denny},
  booktitle={The 61st Annual Meeting Of The Association For Computational Linguistics},
  year={2023}
}

@inproceedings{ma2025s2r,
    title = "{S}$^2${R}: Teaching {LLM}s to Self-verify and Self-correct via Reinforcement Learning",
    author = "Ma, Ruotian  and
      Wang, Peisong  and
      Liu, Cheng  and
      Liu, Xingyan  and
      Chen, Jiaqi  and
      Zhang, Bang  and
      Zhou, Xin  and
      Du, Nan  and
      Li, Jia",
    booktitle = "Proceedings of the 63rd Annual Meeting of the Association for Computational Linguistics (Volume 1: Long Papers)",
    url = "https://aclanthology.org/2025.acl-long.1104/",
    doi = "10.18653/v1/2025.acl-long.1104",
    pages = "22632--22654",
    year = "2025"
}

@misc{huggingfaceh42025aime2024,
  author       = {AIME},
  title        = {AIME 2024},
  year = {2024},
  publisher    = {HuggingFace},
  url          = {https://huggingface.co/datasets/HuggingFaceH4/aime_2024},
  note         = {Accessed: 2025}
}

@misc{yentinglin2025aime2025,
  author       = {AIME},
  title        = {AIME 2025},
  year = {2025},
  publisher    = {HuggingFace},
  url          = {https://huggingface.co/datasets/yentinglin/aime_2025},
  note         = {Accessed: 2025}
}

@misc{gpt4o,
  title={{Hello GPT-4o}},
  author={OpenAI},
  url={https://openai.com/index/hello-gpt-4o/},
  year={2024}
}

@misc{openai2024gpt4omini,
  author       = {OpenAI},
  title        = {{GPT-4o mini: Advancing Cost-Efficient Intelligence}},
  year         = {2024},
  url          = {https://openai.com/index/gpt-4o-mini-advancing-cost-efficient-intelligence/}
}

@misc{openai2025o3mini,
  author       = {OpenAI},
  title        = {{OpenAI o3-mini System Card}},
  year         = {2025},
  url          = {https://openai.com/index/o3-mini-system-card/},
}

@article{guo2025deepseek,
  author       = {Daya Guo and Dejian Yang and Haowei Zhang and Junxiao Song and Peiyi Wang and Qihao Zhu and Runxin Xu and Ruoyu Zhang and Shirong Ma and Xiao Bi and Xiaokang Zhang and Xingkai Yu and Yu Wu and Z. F. Wu and Zhibin Gou and Zhihong Shao and Zhuoshu Li and Ziyi Gao and Aixin Liu and Bing Xue and Bingxuan Wang and Bochao Wu and Bei Feng and Chengda Lu and Chenggang Zhao and Chengqi Deng and Chong Ruan and Damai Dai and Deli Chen and Dongjie Ji and Erhang Li and Fangyun Lin and Fucong Dai and Fuli Luo and Guangbo Hao and Guanting Chen and Guowei Li and H. Zhang and Hanwei Xu and Honghui Ding and Huazuo Gao and Hui Qu and Hui Li and Jianzhong Guo and Jiashi Li and Jingchang Chen and Jingyang Yuan and Jinhao Tu and Junjie Qiu and Junlong Li and J. L. Cai and Jiaqi Ni and Jian Liang and Jin Chen and Kai Dong and Kai Hu and Kaichao You and Kaige Gao and Kang Guan and Kexin Huang and Kuai Yu and Lean Wang and Lecong Zhang and Liang Zhao and Litong Wang and Liyue Zhang and Lei Xu and Leyi Xia and Mingchuan Zhang and Minghua Zhang and Minghui Tang and Mingxu Zhou and Meng Li and Miaojun Wang and Mingming Li and Ning Tian and Panpan Huang and Peng Zhang and Qiancheng Wang and Qinyu Chen and Qiushi Du and Ruiqi Ge and Ruisong Zhang and Ruizhe Pan and Runji Wang and R. J. Chen and R. L. Jin and Ruyi Chen and Shanghao Lu and Shangyan Zhou and Shanhuang Chen and Shengfeng Ye and Shiyu Wang and Shuiping Yu and Shunfeng Zhou and Shuting Pan and S. S. Li and Shuang Zhou and Shaoqing Wu and Tao Yun and Tian Pei and Tianyu Sun and T. Wang and Wangding Zeng and Wen Liu and Wenfeng Liang and Wenjun Gao and Wenqin Yu},
  title        = {DeepSeek-R1 incentivizes reasoning in LLMs through reinforcement learning},
  journal      = {Nature},
  volume       = {648},
  pages        = {633--639},
  year         = {2025},
  doi          = {10.1038/s41586-025-09422-z},
  url          = {https://www.nature.com/articles/s41586-025-09422-z}
}

@article{qwen2025qwen25technicalreport,
      title={Qwen2.5 Technical Report}, 
      author={Qwen},
      year={2025},
      journal={arXiv preprint arXiv:2412.15115},
}

@article{dubey2024llama,
  title={The llama 3 herd of models},
  author={Llama Team, AI@Meta},
  journal={arXiv e-prints},
  pages={arXiv--2407},
  year={2024}
}

@inproceedings{sarthi2024raptor,
  title={Raptor: Recursive abstractive processing for tree-organized retrieval},
  author={Sarthi, Parth and Abdullah, Salman and Tuli, Aditi and Khanna, Shubh and Goldie, Anna and Manning, Christopher D},
  booktitle={The Twelfth International Conference on Learning Representations},
  year={2024}
}

@inproceedings{jiang2023active,
  title={Active retrieval augmented generation},
  author={Jiang, Zhengbao and Xu, Frank F and Gao, Luyu and Sun, Zhiqing and Liu, Qian and Dwivedi-Yu, Jane and Yang, Yiming and Callan, Jamie and Neubig, Graham},
  booktitle={Proceedings of the 2023 Conference on Empirical Methods in Natural Language Processing},
  pages={7969--7992},
  year={2023}
}

@inproceedings{ho2023large,
  title={Large language models are reasoning teachers},
  author={Ho, Namgyu and Schmid, Laura and Yun, Se-Young},
  booktitle={Proceedings of the 61st annual meeting of the association for computational linguistics (volume 1: long papers)},
  pages={14852--14882},
  year={2023}
}
\bibliographystyle{icml2026}

\newpage
\appendix
\onecolumn
\clearpage
\raggedbottom
\section{Limitations \& Future Work}
\label{set:limit}

\paragraph{Task Abstraction Selection.}
Our current approach relies on randomly selecting task abstractions from successful \gptfouromini cases. While this provides a practical starting point, it does not guarantee that the chosen abstraction is the most effective one for a given problem. Future work could explore more principled strategies, such as embedding-based similarity measures to align problems with the most relevant abstractions. Additionally, abstractions could be generated from stronger models (e.g., \gptfouro, \othreemini, or \deepseekrone), which may yield richer reasoning patterns. A systematic comparison across source models would clarify whether larger and more capable models produce abstractions that better generalize across tasks.

\section{Additional Experiment Details}
\label{app:sec:setup}

\subsection{Datasets and Tasks}
\label{app:sec:datasets}
To evaluate the efficacy of our proposed approach compared to other state-of-the-art proposed self-correction baselines, we consider a wide range of tasks and datasets that require various degrees of mathematical and algorithmic reasoning. The introduction to the evaluation datasets is as follows:

\begin{itemize}
    \item \textbf{\game} \citep{yao2023tree}: A mathematical reasoning challenge where the objective is to form an expression that evaluates to $24$ using four given numbers exactly once. For instance, if the input values were ``$7 \ 7 \ 8 \ 11$,'' one valid answer would be ``$8 * (7 + 7 - 11)$.'' This task emphasizes systematic search, strategic reasoning, and pattern recognition. We use the $99$ examples from \citep{yang2024buffer} to evaluate models capacity for refining computational heuristics and strategy over manual attempts.
    \item \textbf{\checkmate} \citep{srivastava2023bigbench}: A chess reasoning challenge where the objective is to identify the move, expressed in Standard Algebraic Notation (SAN), that delivers checkmate in a given position. The input consists of a sequence of prior moves leading to a state where a one-move checkmate is possible. For instance, after the sequence ``1. e4 e5 2. Qh5 Nc6 3. Bc4 Nf6,'' the correct output is ``Qxf7\#.'' This task probes spatial reasoning, rule application, and tactical foresight. We use 3,500 curated game positions to evaluate models’ ability to achieve exact match accuracy in identifying checkmating moves.
    \item \textbf{\wordsorting} \citep{suzgun2023bigbenchhard}: A linguistic reasoning challenge where the model must sort a given list of words according to a specified criterion, such as alphabetical order, length, or semantic attributes. For example, sorting ``cat, elephant, dog'' by length yields ``cat, dog, elephant.'' This task tests systematic application of sorting rules, attention to fine-grained instructions, and consistency in following multi-step language-based procedures.
    \item \textbf{\aimefour and \aimefive}: The American Invitational Mathematics Examination (AIME) is a prestigious high-school competition featuring complex problems across algebra, combinatorics, number theory, geometry, and probability. These questions require deep mathematical reasoning and multi-step problem-solving. We consider two subsets that are shown to be challenging for large language models \citep{suzgun2025dynamic}, namely, AIME 2024\footnote{\url{https://huggingface.co/datasets/HuggingFaceH4/aime_2024}} and AIME 2025\footnote{\url{https://huggingface.co/datasets/yentinglin/aime\_2025}}, where each subset has 30 questions.
\end{itemize}


\subsection{Baselines}
\label{sec:app:setup:baselines}
Here, we introduce the details of the baseline methods for comparison with our proposed method:

\begin{itemize}
    \item \textbf{\vanilla}: \vanilla is a basic iterative refinement method in which the LLM reflects on its initial output and generates a revised response. We include \vanilla as one of our comparison baselines, following the recent work of \cite{song2025progco}, which showed that this simple approach can improve performance when using large base models such as GPT-4o (refer to Table 1 in \cite{song2025progco} for detailed results).
    \item \textbf{\refine} \citep{madaan2023self-refine}: \refine iteratively reviews its own output to generate feedback and proposes refinements based on the feedback from the previous step, continuing this process until no errors are detected or a maximum number of iterations is reached.
    \item \textbf{\reflexion} \citep{shinn2023reflexion}: \reflexion is an iterative approach where the model first evaluates its output, then generates verbal feedback about its previous output based on the evaluation and uses this feedback to refine its output. \cite{shinn2023reflexion} uses ground truth labels about answer correctness for evaluation to guide the self-correction process. However, in our experiments, we rely on the Chain-of-Thought (CoT) generated by the model itself, because we assume that the ground truth context or an external API is not available (see Section 4.2 in \cite{shinn2023reflexion}).
    \item \textbf{\tick} \citep{cook2024self-tick}: \tick first generates a checklist, i.e., Yes/No questions, for the input task, and then verifies whether the generated response satisfies all the questions from the checklist one by one. Any unsatisfied verification points will be used as feedback to refine and improve its own output.
    \item \textbf{\consistency} \citep{wang2022self-consistency}: \consistency is a decoding strategy that samples diverse reasoning trajectories from the model and selects the most consistent answer based on majority voting among these reasoning traces.
    \item \textbf{SuperCorrect} \citep{yang2024supercorrect}: SuperCorrect is a two-stage framework in which a large teacher model supervises the reasoning and self-correction processes of a smaller student. First, the reasoning trajectories generated by the teacher model are used to perform Supervised Fine-Tuning (SFT) on the student model to enhance its reasoning capabilities. Then, the teacher model provides corrections for the hierarchical reasoning trajectories generated by the SFT-fine-tuned student model, and a collaborative Direct Preference Optimization (DPO) technique is applied to improve the ability of the student model to refine its outputs based on these correction traces. \cite{yang2024supercorrect} used \texttt{o1-mini} or \texttt{GPT-4o-mini} as the teacher model and \texttt{Qwen-2.5-Math-7B-Instruct} as the student model.
    \item \textbf{\stwor} \citep{ma2025s2r}: \stwor introduces a framework that enhances LLM reasoning by teaching models to self-verify and self-correct during inference. It begins by initializing LLMs with iterative self-verification and self-correction behaviors through supervised fine-tuning on curated data. These skills are further strengthened by both outcome-level and process-level reinforcement learning, with minimized resource requirements, enabling the model to adaptively refine its reasoning process during inference. Experimental results demonstrate significant accuracy improvements, showcasing the effectiveness of \stwor in enhancing LLM reasoning capabilities.
    \item \textbf{\stasc} \citep{moskvoretskii2025selftaught}: \stasc focuses on self-correction in small language models through iterative fine-tuning using solely self-generated data. The Self-Taught Self-Correction (\stasc) algorithm incorporates multiple algorithmic design choices, allowing models to improve their outputs without external supervision. Experimental results on a question-answering task demonstrate that \stasc effectively learns self-correction, leading to significant performance improvements. The study provides insights into the mechanisms of self-correction and the impact of different design choices on learning dynamics and overall performance.
\end{itemize}

\subsubsection{Compression against \consistency}
Why should we compare our approach with \consistency \citep{wang2022self-consistency}? \consistency prompts models to generate multiple responses and select the most consistent responses by performing majority voting. A recent study \citep{huang2023cannot-self-correct} shows that \consistency outperforms the multi-agent debate approach with the equivalent number of responses. A recent study compared majority voting with the other techniques and showed that it outperforms other aggregation functions \citep{song2025progco}.

\subsubsection{Compression against \supercorrect}
Why should we compare our approach with \supercorrect \citep{yang2024supercorrect}? \supercorrect is a recent self-correcting model that leverages distillation from larger models. A model that is fine-tuned in two stages, SFT and DPO. In the SFT stage, the model is fine-tuned on reasoning traces generated by a larger model on math datasets, and in the DPO stage, it is fine-tuned on a preference pair dataset of corrected reasoning trajectories generated by the large model and the small model.

\subsection{Answer Extraction Protocol}
To keep the evaluation consistent and reliable, all models are asked to write their final answers in a structured and machine-readable format. Each answer is expected to be wrapped in the following XML-style tags:

\begin{center}
    \texttt{<\textcolor{darkgreen}{\textbf{Answer}}> Your Final Answer Here </\textcolor{darkgreen}{\textbf{Answer}}>}
\end{center}

This specific format makes it easy to correctly read and process the answers, avoiding mistakes from extra text or ambiguous outputs. After being extracted, the final answers are evaluated using the accuracy measure for each specific task.

\subsection{Evaluation Protocol}
Given the diversity of tasks, we use different accuracy metrics tailored to the specific requirements of each benchmark:

\begin{itemize}
    \item \textbf{Exact Match (EM)}. EM is a strict metric that marks an answer as correct only if it matches the ground-truth label exactly, without extra text or formatting differences.
    \item \textbf{Soft Match (SM)}. SM is a lenient metric that marks an answer as correct if the ground-truth label appears in the model’s output, ignoring minor formatting differences such as punctuation or whitespace. Unlike EM, SM does not require the output to match the label verbatim.
    \item \textbf{Functionally Correct (FC)}. FC is a flexible metric that marks an answer as correct if it satisfies task-specific constraints, even when the exact wording, numeral presentation, or formatting differs from the reference solution.
\end{itemize}

We apply \textbf{EM} for \checkmate, \textbf{SM} for \wordsorting, and \textbf{FC} for \game, \aimefour, and \aimefive benchmarks.

To measure self-correction performance, we report and analyze the following metrics: \textbf{(1)} \textbf{Acc@ti}: accuracy at the $i$-th attempt; \textbf{(2)} \textbf{$\Delta^{i \to c}(t_{i-1}, t_i)$}: the fraction of problems that were incorrect at attempt $i-1$ but corrected at attempt $i$, capturing how many new problems self-correction solves; and \textbf{(3)} \textbf{$\Delta^{c \to i}(t_{i-1}, t_i)$}: the fraction of problems that were correct at attempt $i-1$ but become incorrect at attempt $i$, reflecting how reliably the model preserves correct answers.
\section{Prompts}

\subsection{Initial Generation}
\begin{table}[H]
\centering
\begin{tcolorbox}[colback=gray!5!white,colframe=black!100,title=\game,label={appendix:refinement}]
Let's play a game called 24. You'll be given four integers, and your objective is to use each number only once, combined with any of the four arithmetic operations (addition, subtraction, multiplication, and division) and parentheses, to achieve a total of 24. For example, if the input is 4, 7, 8, and 8, the output could be (7 * 8) - (4 * 8). You only need to find one feasible solution!

Input: \textit{\{question\}}. Please provide the final answer within \verb|<|Answer\verb|>| Your Final Answer Here \verb|<|/Answer\verb|>|.
\end{tcolorbox}
\end{table}
\begin{table}[!h]
\centering
\begin{tcolorbox}[colback=gray!5!white,colframe=black!100,title=\wordsorting,label={appendix:refinement}]

Sort a list of words alphabetically, placing them in a single line of text separated by spaces.

Input: \textit{\{question\}}. Please provide the final answer within \verb|<|Answer\verb|>| Your Final Answer Here \verb|<|/Answer\verb|>|.

\end{tcolorbox}
\end{table}
\begin{table}[!h]
\centering
\begin{tcolorbox}[colback=gray!5!white,colframe=black!100,title=\checkmate,label={appendix:refinement}]

Given a series of chess moves written in Standard Algebraic Notation (SAN), determine the next move that will result in a checkmate.

Input: \textit{\{question\}}. Please provide the final answer within \verb|<|Answer\verb|>| Your Final Answer Here \verb|<|/Answer\verb|>|.

\end{tcolorbox}
\end{table}
\begin{table}[!ht]
\centering
\begin{tcolorbox}[colback=gray!5!white,colframe=black!100,title=\aimefour,label={appendix:refinement}]

Given the input question, your task is to provide the answer to the question. 

Input: \textit{\{question\}}. Please provide the final answer within \verb|<|Answer\verb|>| Your Final Answer Here \verb|<|/Answer\verb|>|.

\end{tcolorbox}
\end{table}
\begin{table}[!ht]
\centering
\begin{tcolorbox}[colback=gray!5!white,colframe=black!100,title=\aimefive,label={appendix:refinement}]

Given the input question, your task is to provide the answer to the question. 

Input: \textit{\{question\}}. Please provide the final answer within \verb|<|Answer\verb|>| Your Final Answer Here \verb|<|/Answer\verb|>|.

\end{tcolorbox}
\end{table}

\subsection{Task Abstraction}
\begin{table}[H]
\centering
\begin{tcolorbox}[colback=gray!5!white,colframe=black!100,title=Task Abstraction and Distillation,label={appendix:prompt_task_distillation}]
As a highly professional and intelligent expert in information distillation, you excel at extracting essential information to solve problems from user input queries. You adeptly transform this extracted information into a suitable format based on the respective type of issue. If the problem can be generalized to a higher level to solve multiple issues, further analysis and explanation will be provided upon your next response. Please categorize and extract the crucial information required to solve the problem from the user's input query. Combining these two elements will generate distilled information. The distilled information should include: \\

1. Values and information of key variables extracted from user input, which will be handed over to the respective expert for task resolution, ensuring all essential information required to solve the problem is provided. \\
2. The objective of the problem and corresponding constraints. \\
3. Extend the problem based on 1 and 2, propose a meta problem that can address the user query and handle more input and output variations. Incorporate the real-world scenario of the extended problem along with the types of key variables and information constraints from the original problem to restrict the key variables in the extended problem. After that, use the user query input key information as input to solve the problem, as an example. \\
4. Try to transform the problem into a Python algorithm problem, and provide the input parameters. \\
5. Your task is to distill the problem; you shouldn't give the final result or possible solution in your response. \\

Please distill the information following the format below and cease responding after the output of the distilled information. \\

Meta distiller Respond: \\

Distilled Information: \\

1. Key information: \\

2. Restriction: (It should be noted that the answer should strictly follow the real-world rule, such as in an arithmetic equation, the priority of operators, the need for parentheses, etc. So, according to the distilled information, emphasize the real-world rules that need to be followed within the problem.) \\

3. Distilled task: \\

4. Python transformation: \\
Input parameters: (The names of each variable should be clear and not confusing, and correspond to the entity names in the problem) \\
variable1\_name = x \\
variable2\_name = y \\
..... \\
variableN\_name = z \\

5. Answer form: (Optional, skip when there is no specific answer form) \\

** Note: The generation ends here. Do not show this message in your answer! **
\end{tcolorbox}
\label{prompt:task-abstraction}
\end{table}

\subsection{Solution Instantiation}
\begin{table}[H]
\centering
\begin{tcolorbox}[colback=gray!5!white,colframe=black!100,title=Solution Instantiation and Refinement,label={appendix:refinement}]
You are an expert in problem analysis and can apply previous problem-solving approaches to new issues. The user will provide an input query and a specific task description. Your goal is to analyze the user's query and generate a specific solution based on the task description. If the solution does not involve code, provide a final answer that is easy to extract from the text. \\

Distilled information: \\
\textit{\{distilled\_information\}} \\
User Input: \\
\textit{\{user\_input\}} \\

Instantiated Solution: \\
Please analyze the above user task description and thought template, and generate a specific, detailed solution. Please provide a clear and extractable final answer within \verb|<|Answer\verb|>| Your Final Answer Here \verb|<|/Answer\verb|>|.

\end{tcolorbox}
\label{tab:appendix_improver_minimal_prompt}
\end{table}
\begin{table}[H]
\centering
\begin{tcolorbox}[colback=gray!5!white,colframe=black!100,title=Solution Instantiation and Refinement (Small Models),label={appendix:refinement}]
You are an expert in problem analysis and can apply previous problem-solving approaches to new issues. The user will provide an input query and a specific task description. Your goal is to analyze the user's query and generate a specific solution based on the task description. If the solution does not involve code, provide a final answer that is easy to extract from the text. \\

Distilled information: \\
\textit{\{distilled\_information\}} \\
User Input: \\
\textit{\{user\_input\}} \\
Thought Template: \\
\textit{\{task\_abstraction\}} \\

Instantiated Solution: \\
Please analyze the above user task description and thought template, and generate a specific, detailed solution. Please provide a clear and extractable final answer within \verb|<|Answer\verb|>| Your Final Answer Here \verb|<|/Answer\verb|>|.
\end{tcolorbox}
\label{tab:appendix_improver_minimal_prompt}
\end{table}

\subsection{Ablations}

\begin{table}[H]
\centering
\begin{tcolorbox}[colback=gray!5!white,colframe=black!100,title=Metadata Extraction,label={appendix:refinement}]
You are given the following task: \\
\\
\textit{\{Task-specific Initial Generation Prompt\}} \\
\\
Extract the following metadata from the task: \\
- Task objective\\
- Required inputs\\
- Expected outputs\\
- Constraints or assumptions\\
- Relevant entities or variables\\
\\
Present the extracted metadata in a structured bullet-point format.
\end{tcolorbox}
\label{tab:appendix_improver_minimal_prompt}
\end{table}
\begin{table}[H]
\centering
\begin{tcolorbox}[colback=gray!5!white,colframe=black!100,title=Generating Answer using Metadata,label={appendix:refinement}]
You are given the original task and its extracted metadata: \\
\\
Task:\\
\textit{\{Task-specific Initial Generation Prompt\}} \\
\\
Extracted metadata: \\
\textit{\{metadata\}} \\

Using the metadata above, carefully reason step by step and produce the final answer. Please provide the final answer within \verb|<|Answer\verb|>| Your Final Answer Here \verb|<|/Answer\verb|>|.
\end{tcolorbox}
\label{tab:appendix_improver_minimal_prompt}
\end{table}

\begin{table}[H]
\centering
\begin{tcolorbox}[colback=gray!5!white,colframe=black!100,title=Summary Extraction,label={appendix:refinement}]
You are given the following task:\\
\\
\textit{\{Task-specific Initial Generation Prompt\}} \\

Summarize the task in your own words, focusing on the most important information.
\end{tcolorbox}
\label{tab:appendix_improver_minimal_prompt}
\end{table}
\begin{table}[H]
\centering
\begin{tcolorbox}[colback=gray!5!white,colframe=black!100,title=Generating Answer using Summary,label={appendix:refinement}]
You are given the original task and its summary: \\
\\
Task:\\
\textit{\{Task-specific Initial Generation Prompt\}} \\
\\
Summary:\\
\textit{\{task\_summary\}} \\
\\
Using the summary above, reason step by step and provide the final answer. Please provide the final answer within \verb|<|Answer\verb|>| Your Final Answer Here \verb|<|/Answer\verb|>|.
\end{tcolorbox}
\label{tab:appendix_improver_minimal_prompt}
\end{table}
\section{Additional Comparison with Fine-tuning Baselines}
\label{sec:app:finetune}

\begin{table}[H]
\centering
\caption{Self-correction performance on \game, \wordsorting, \checkmate, \aimefour, and \aimefive using fine-tuning based baselines (\supercorrect, \stwor, \stasc) and our methods (\ourmethod, \ourmethodsmall). \textbf{Bold} indicates the best performance.}
\label{tab:app:finetune}
\resizebox{\textwidth}{!}{
\setlength{\tabcolsep}{1pt}
    \begin{tabular}{lcccccccccccccccccccccc}
        \toprule
        \multirow{2}{*}{\textbf{Method}} && \multirow{2}{*}{\textbf{Acc@0}} && \multicolumn{3}{c}{\bf Iteration 1} && \multicolumn{3}{c}{\bf Iteration 2} && \multicolumn{3}{c}{\bf Iteration 3} && \multicolumn{3}{c}{\bf Iteration 4} && \multicolumn{3}{c}{\bf Iteration 5} \cr
        
        \cmidrule(lr){5-7} \cmidrule(lr){9-11} \cmidrule(lr){13-15} \cmidrule(lr){17-19} \cmidrule(lr){21-23}
        
        && && \makecell{\textbf{Acc}} & \makecell{$\Delta^{{i \to c}}(t_0, t_1)$} & \makecell{$\Delta^{{c \to i}}(t_0, t_1)$} && 
          \makecell{\textbf{Acc}} & \makecell{$\Delta^{{i \to c}}(t_1, t_2)$} & \makecell{$\Delta^{{c \to i}}(t_1, t_2)$} &&
          \makecell{\textbf{Acc}} & \makecell{$\Delta^{{i \to c}}(t_2, t_3)$} & \makecell{$\Delta^{{c \to i}}(t_2, t_3)$} &&
          \makecell{\textbf{Acc}} & \makecell{$\Delta^{{i \to c}}(t_3, t_4)$} & \makecell{$\Delta^{{c \to i}}(t_3, t_4)$} &&
          \makecell{\textbf{Acc}} & \makecell{$\Delta^{{i \to c}}(t_4, t_5)$} & \makecell{$\Delta^{{c \to i}}(t_4, t_5)$} \\
            \midrule
            & \multicolumn{21}{c}{\textbf{\game}} \\
            \midrule
            SuperCorrect    && 11.22 && 10.2 & 0.0 & 1.02 && 7.14 & 1.02 & 4.08 && 5.1 & 1.02 & 3.06 && 12.24 & 8.16 & 1.02 && 12.24 & 2.04 & 2.04 \\ 
            S2R             && \textbf{20.41} && 14.29 & 5.1 & 11.22 && 9.18 & 1.02 & 6.12 && 10.2 & 5.1 & 4.08 && 8.16 & 4.08 & 6.12 && 6.12 & 3.06 & 5.1 \\ 
            STaSC           && 2.04 && 0.0 & 0.0 & \textbf{2.04} && 0.0 & 0.0 & 0.0 && 0.0 & 0.0 & 0.0 && 0.0 & 0.0 & \textbf{0.0} && 1.02 & 1.02 & \textbf{0.0} \\ 
            \ourmethod      && 8.16 && 11.22 & 8.16 & 5.1 && 12.24 & 1.02 & 0.0 && 11.22 & 1.02 & 2.04 && 12.24 & 2.04 & 1.02 && 12.24 & 0.0 & 0.0 \\ 
            \ourmethodsmall && 8.16 && \textbf{41.84} & \textbf{37.76} & 4.08 && \textbf{52.04} & \textbf{23.47} & 13.27 && \textbf{47.96} & \textbf{9.18} & \textbf{13.27} && \textbf{51.02} & \textbf{11.22} & 8.16 && \textbf{42.86} & \textbf{10.2} & 18.37 \\
            \midrule
            & \multicolumn{21}{c}{\textbf{\wordsorting}} \\
            \midrule
            SuperCorrect && 2.5 && 1.25 & 0.0 & 1.25 && 0.0 & 0.0 & 1.25 && 0.0 & 0.0 & 0.0 && 0.0 & 0.0 & 0.0 && 0.0 & 0.0 & 0.0 \\ 
            S2R && 16.25 && 3.75 & 0.0 & 12.5 && 3.75 & 1.25 & 1.25 && 1.25 & 0.0 & 2.5 && 1.25 & 0.0 & 0.0 && 0.0 & 0.0 & 1.25 \\ 
            STaSC && \textbf{18.75} && 10.0 & 0.0 & 8.75 && 7.5 & 1.25 & 3.75 && 6.25 & 0.0 & 1.25 && 3.75 & 0.0 & 2.5 && 3.75 & 0.0 & 0.0 \\ 
            \ourmethod && 13.75 && \textbf{66.25} & \textbf{57.5} & 5.0 && \textbf{61.25} & 5.0 & 10.0 && \textbf{53.75} & 1.25 & 8.75 && 45.0 & 2.5 & 11.25 && 35.0 & 3.75 & 13.75 \\ 
            \ourmethodsmall && 13.75 && 48.75 & 40.0 & 5.0 && 32.5 & \textbf{8.75} & 25.0 && 33.75 & \textbf{10.0} & 8.75 && \textbf{33.75} & \textbf{8.75} & 8.75 && \textbf{33.75} & \textbf{8.75} & 8.75 \\
            \midrule
            & \multicolumn{21}{c}{\textbf{\checkmate}} \\
            \midrule
            SuperCorrect && 0.0 && 0.0 & 0.0 & 0.0 && 0.0 & 0.0 & 0.0 && 0.0 & 0.0 & 0.0 && 0.0 & 0.0 & 0.0 && 0.0 & 0.0 & 0.0 \\ 
            S2R && 0.0 && 0.0 & 0.0 & 0.0 && 0.0 & 0.0 & 0.0 && 0.0 & 0.0 & 0.0 && 0.0 & 0.0 & 0.0 && 0.0 & 0.0 & 0.0 \\ 
            STaSC && 1.33 && 0.0 & 0.0 & 1.33 && 0.0 & 0.0 & 0.0 && 0.0 & 0.0 & 0.0 && 0.0 & 0.0 & 0.0 && 0.0 & 0.0 & 0.0 \\ 
            \ourmethod && 2.67 && 4.0 & 4.0 & 2.67 && 5.33 & 1.33 & 0.0 && 5.33 & 0.0 & 0.0 && 6.67 & 1.33 & 0.0 && 5.33 & 1.33 & 2.67 \\ 
            \ourmethodsmall && 2.67 && \textbf{10.67} & \textbf{10.67} & 2.67 && \textbf{25.33} & \textbf{18.67} & 4.0 && \textbf{25.33} & \textbf{12.0} & 12.0 && \textbf{29.33} & \textbf{16.0} & 12.0 && \textbf{22.67} & \textbf{10.67} & 17.33 \\
            \midrule
            & \multicolumn{21}{c}{\textbf{\aimefour}} \\
            \midrule
            SuperCorrect && 13.33 && 13.33 & 0.0 & \textbf{0.0} && 16.67 & 3.33 & 0.0 && 20.0 & 3.33 & 0.0 && 20.0 & 0.0 & 0.0 && 20.0 & 0.0 & 0.0 \\ 
            S2R && 13.33 && 6.67 & 3.33 & 10.0 && 6.67 & 0.0 & 0.0 && 6.67 & 3.33 & 3.33 && 6.67 & 3.33 & 3.33 && 13.33 & \textbf{10.0} & 3.33 \\ 
            STaSC && 10.0 && 0.0 & 0.0 & 10.0 && 0.0 & 0.0 & 0.0 && 0.0 & 0.0 & 0.0 && 0.0 & 0.0 & 0.0 && 0.0 & 0.0 & 0.0 \\ 
            \ourmethod && 20.0 && 20.0 & 3.33 & 3.33 && 23.33 & 3.33 & 0.0 && 23.33 & 0.0 & 0.0 && 26.67 & 3.33 & 0.0 && 26.67 & 0.0 & 0.0 \\ 
            \ourmethodsmall && \textbf{20.0} && \textbf{23.33} & \textbf{13.33} & 10.0 && \textbf{26.67} & 3.33 & 0.0 && \textbf{26.67} & 3.33 & 3.33 && \textbf{30.0} & \textbf{6.67} & 3.33 && \textbf{30.0} & 0.0 & 0.0 \\
            \midrule
            & \multicolumn{21}{c}{\textbf{\aimefive}} \\
            \midrule
            SuperCorrect && 3.33 && 6.67 & 3.33 & 0.0 && 13.33 & 6.67 & 0.0 && 13.33 & 0.0 & 0.0 && 16.67 & 3.33 & 0.0 && 16.67 & 3.33 & 3.33 \\ 
            S2R && 10.0 && 6.67 & 3.33 & 6.67 && 3.33 & 3.33 & 6.67 && 0.0 & 0.0 & 3.33 && 0.0 & 0.0 & 0.0 && 6.67 & \textbf{6.67} & 0.0 \\ 
            STaSC && 0.0 && 0.0 & 0.0 & 0.0 && 0.0 & 0.0 & 0.0 && 0.0 & 0.0 & 0.0 && 0.0 & 0.0 & 0.0 && 0.0 & 0.0 & 0.0 \\ 
            \ourmethod && 10.0 && 10.0 & 10.0 & 10.0 && 13.33 & 3.33 & 0.0 && 13.33 & 3.33 & 3.33 && 16.67 & 3.33 & 0.0 && 16.67 & 0.0 & 0.0 \\ 
            \ourmethodsmall && 10.0 && \textbf{13.33} & \textbf{13.33} & 10.0 && \textbf{16.67} & 3.33 & 0.0 && \textbf{16.67} & \textbf{6.67} & 6.67 && \textbf{16.67} & 0.0 & 0.0 && \textbf{20.0} & 3.33 & 0.0 \\
            \bottomrule
    \end{tabular}
}
\end{table}

\section{Additional Comparison with \consistency}
\label{sec:app:add-self-consistency}

\begin{figure}[H]
    \centering
    \subfloat[\centering \gptfouromini]{{\includegraphics[width=0.5\textwidth]{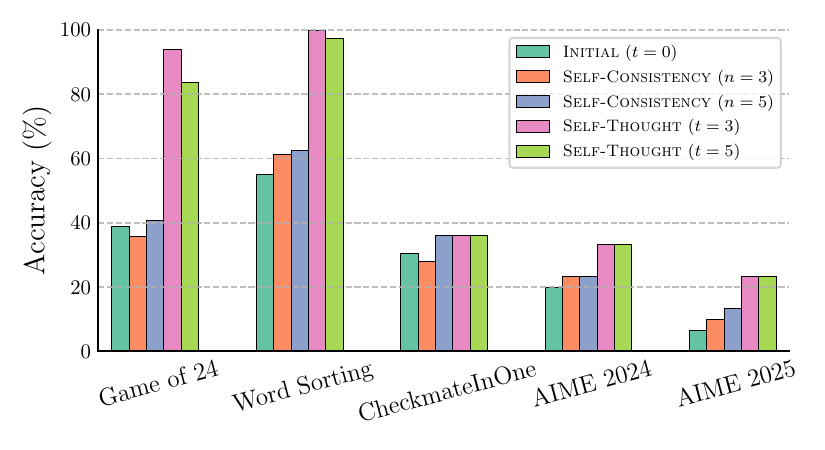}}}
    \subfloat[\centering \gptfouro]{{\includegraphics[width=0.5\textwidth]{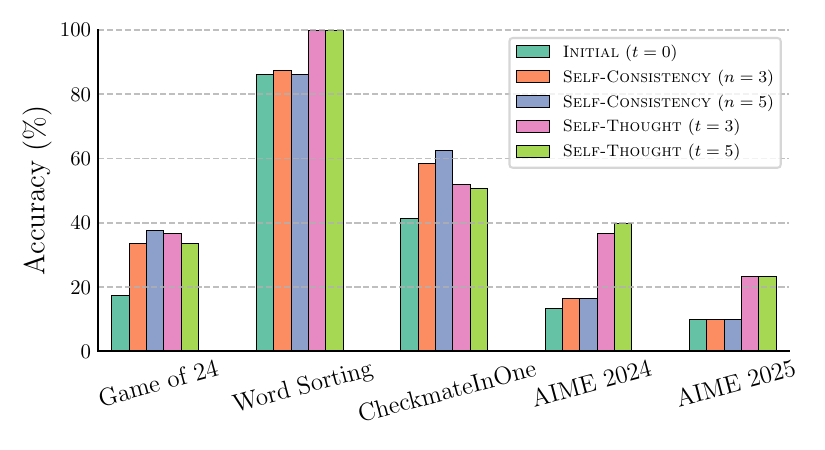}}}
    \qquad
    \subfloat[\centering \othreemini]{{\includegraphics[width=0.5\textwidth]{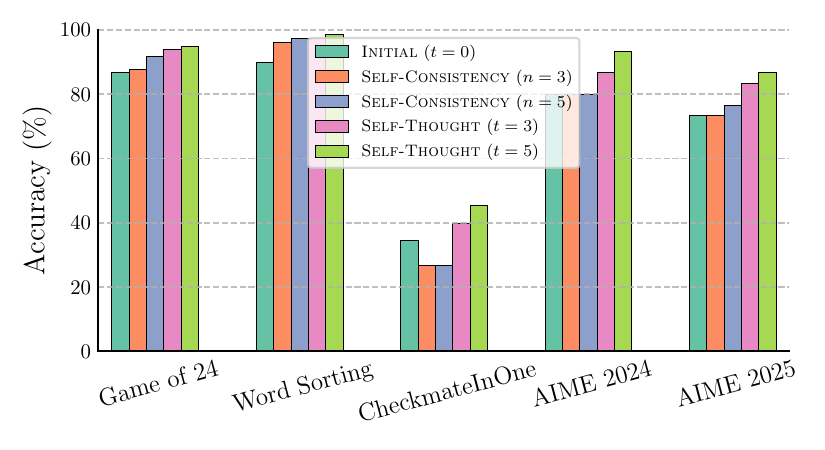}}}
    \subfloat[\centering \deepseekrone]{{\includegraphics[width=0.5\textwidth]{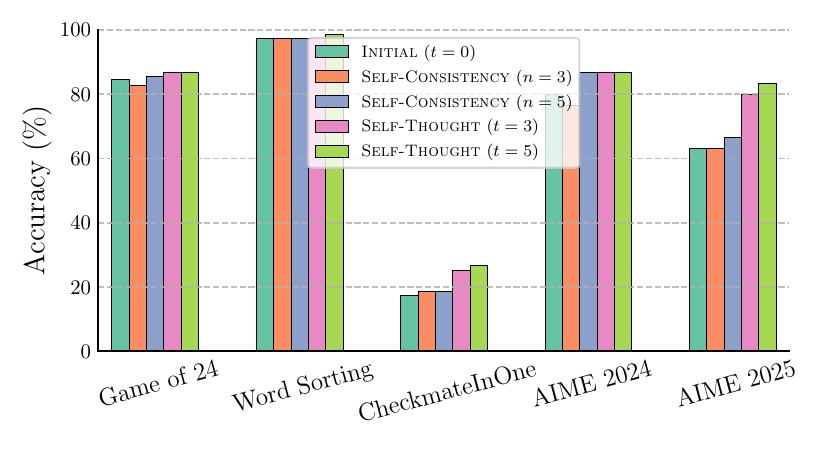}}}
    \qquad
    \subfloat[\centering \qwen]{\includegraphics[width=0.5\textwidth]{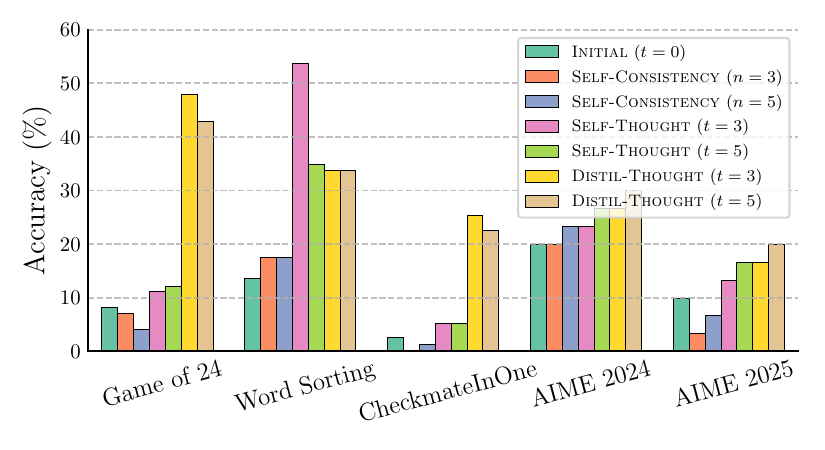}}
    \subfloat[\centering \llama]{{\includegraphics[width=0.5\textwidth]{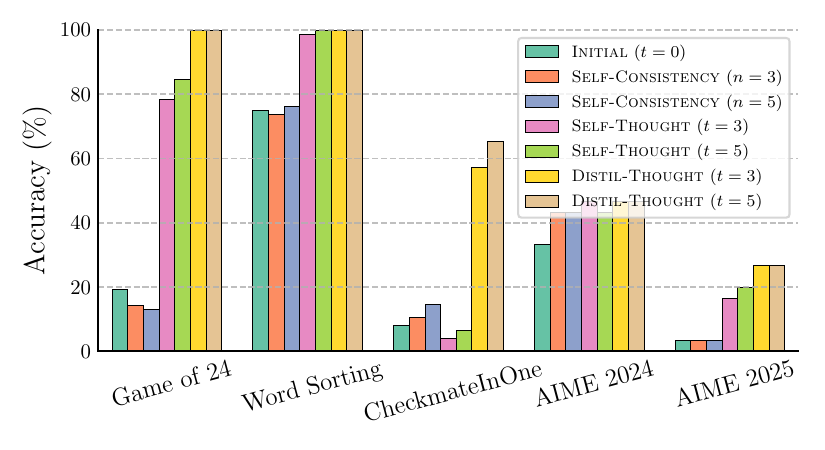}}}
    \caption{Comparison of \ourmethod and \ourmethodsmall with the \consistency.}
    \label{fig:app:self-consistency}
\end{figure}
\section{Additional Experiments and Results}
\label{sec:app:results}

\subsection{Full Initial Answer for An Example}
\begin{longtable}{p{0.96\textwidth}}
\caption{Full Initial Answer for Provided Example} \label{tab:full-answer} \\
\toprule
To find the number of eight-digit integers that use each of the digits 1 through 8 exactly
once and are divisible by 22, we need to check the divisibility rules for 22. Since 22 = 2 ×
11, an integer must be divisible by both 2 and 11. \\
\\
1. \textbf{Divisibility by 2:} An integer is divisible by 2 if its last digit is even. The even digits among 1 to 8 are 2, 4, 6, and 8. Therefore, the last digit of our eight-digit integer
must be one of these four digits. \\
2. \textbf{Divisibility by 11:} A number is divisible by 11 if the absolute difference between
the sum of its digits in odd positions and the sum of its digits in even positions is
either 0 or a multiple of 11. \\
\\
Let's denote the digits of the number as d1, d2, d3, d4, d5, d6, d7, d8 where d8 is the
last digit. The odd positions are d1, d3, d5, d7 and the even positions are d2, d4, d6, d8. \\
The total sum of digits from 1 to 8 is: 1 + 2 + 3 + 4 + 5 + 6 + 7 + 8 = 36 \\
If we denote the sum of the digits in odd positions as S1 and the sum of the digits in
even positions as S2, we know: S1 + S2 = 36 \\
The condition for divisibility by 11 gives us: $|S1 - S2|$ is divisible by 11 \\
Since $S2 = 36 - S1$, we can rewrite the divisibility condition as: $|S1 - (36 - S1)| = |2 \times S1 -
36|$ is divisible by 11 \\
This simplifies to: $|2 \times S1 - 36|$ is divisible by 11 \\
We can find the values of S1 that are possible. The minimum sum S1 can take (if we
pick the smallest digits) is 1 + 2 + 3 + 4 = 10, and the maximum (if we pick the largest) is
5 + 6 + 7 + 8 = 26. Therefore, S1 can take values from 10 to 26. \\
Next, we find S1 such that: 2 × S1 is divisible by 11 and congruent to 36 modulo 11 \\
Calculating 36 modulo 11 gives us 36 $\equiv$ 3 mod 11. Thus, we need: 2 × S1 $\equiv$ 3 mod 11 \\
To solve 2 × S1 $\equiv$ 3 mod 11, we can try values for S1:
\begin{itemize}
    \item $S1 = 2~~gives~~2 \times 2 = 4$
    \item $S1 = 3~~gives~~2 \times 3 = 6$
    \item $S1 = 4~~gives~~2 \times 4 = 8$
    \item $S1 = 5~~gives~~2 \times 5 = 10$
    \item $S1 = 6~~gives~~2 \times 6 = 12 \equiv 1~~mod~~11$
    \item $S1 = 7~~gives~~2 \times 7 = 14 \equiv 3~~mod~~11$ (valid)
    \item $S1 = 8~~gives~~2 \times 8 = 16 \equiv 5~~mod~~11$
    \item $S1 = 9~~gives~~2 \times 9 = 18 \equiv 7~~mod~~11$
\end{itemize}
Continuing this, we find:
\begin{itemize}
    \item $S1 = 14~~gives~~2 \times 14 = 28 \equiv 6~~mod~~11$
    \item $S1 = 15~~gives~~2 \times 15 = 30 \equiv 8~~mod~~11$
    \item $S1 = 16~~gives~~2 \times 16 = 32 \equiv 10~~mod~~11$
    \item $S1 = 17~~gives~~2 \times 17 = 34 \equiv 1~~mod~~11$
    \item $S1 = 18~~gives~~2 \times 18 = 36 \equiv 3~~mod~~11$ (valid)
\end{itemize}
Continuing this gives us all valid S1 values.

After calculating valid arrangements for S1 and S2, we can compute the total
arrangements for each case based on the last digit being 2, 4, 6, or 8 (up to 4 cases). \\
We find N and compute N - 2025. \\
After calculating the final result, we find:\\
N = 2016\\
So, the difference is: 2016 - 2025 = -9\\
Thus, the answer is: \textless Answer\textgreater\ -9 \textless/Answer\textgreater \\
\bottomrule
\end{longtable}

\subsection{Results on Second Step of Self-Correction}
\begin{table}[H]
\centering
\caption{The second attempt of self-correction performance on \game, \wordsorting, \checkmate, \aimefour, \aimefive with \gptfouromini, \gptfouro, \othreemini, and \deepseekrone. Green ({\scriptsize\textcolor{ggreen}{\footnotesize $\uparrow$}}) and red ({\scriptsize\textcolor{gred}{\footnotesize $\downarrow$}}) arrows
indicate performance changes against the previous attempt (i.e., \initial). \textbf{Bold} corresponds to the best performance.}
\label{tab:t2_delta_large}
\resizebox{\textwidth}{!}{
\setlength{\tabcolsep}{1pt}
    \begin{tabular}{llccllccllccllccllccla}
        \toprule
        \multirow{2}{*}{\textbf{Method}} & \multicolumn{3}{c}{\bf Game of 24} && \multicolumn{3}{c}{\bf Word Sorting} && \multicolumn{3}{c}{\bf CheckmateInOne} && \multicolumn{3}{c}{\bf AIME 2024} && \multicolumn{3}{c}{\bf AIME 2025} && \multicolumn{1}{c}{\bf Mean} \cr
        \cmidrule(lr){2-4} \cmidrule(lr){6-8} \cmidrule(lr){10-12} \cmidrule(lr){14-16} \cmidrule(lr){18-20}
        & \makecell{\textbf{Acc@t2}} & \makecell{$\underset{i \to c}{\Delta}(t_1, t_2)$} & \makecell{$\underset{c \to i}{\Delta}(t_1, t_2)$} && 
          \makecell{\textbf{Acc@t2}} & \makecell{$\underset{i \to c}{\Delta}(t_1, t_2)$} & \makecell{$\underset{c \to i}{\Delta}(t_1, t_2)$} &&
          \makecell{\textbf{Acc@t2}} & \makecell{$\underset{i \to c}{\Delta}(t_1, t_2)$} & \makecell{$\underset{c \to i}{\Delta}(t_1, t_2)$} &&
          \makecell{\textbf{Acc@t2}} & \makecell{$\underset{i \to c}{\Delta}(t_1, t_2)$} & \makecell{$\underset{c \to i}{\Delta}(t_1, t_2)$} &&
          \makecell{\textbf{Acc@t2}} & \makecell{$\underset{i \to c}{\Delta}(t_1, t_2)$} & \makecell{$\underset{c \to i}{\Delta}(t_1, t_2)$} &&
          \makecell{\textbf{Acc@t2}} \\
        \midrule
        & \multicolumn{21}{c}{\textbf{\gptfouromini}} \\
        \midrule
        \initial & 38.78 & - & - && 55.0 & - & - &&30.67 & - & - &&20.0 & - & - &&6.67 & - & - && 0.3 \\ 
        \hdashline\noalign{\vskip 0.4ex}
        \vanilla & 28.57 \ua{4.08} & 15.31 & 11.22 && 61.25 \ua{1.25} & 5.0 & 3.75 && 13.33 \ua{4.0} & 10.67 & 6.67 && 10.0  & 3.33 & 3.33 && 13.33 \ua{3.33} & 3.33 & 0.0 && 0.25 \ua{2.0} \\ 
        \refine & 19.39 \da{6.12} & 9.18 & 15.31 && 68.75 \ua{10.0} & 10.0 & 0.0 && 16.0 \ua{5.33} & 8.0 & 2.67 && 16.67 \ua{3.34} & 6.67 & 3.33 && 16.67  & 0.0 & 0.0 && 0.27 \ua{2.0} \\ 
        \tick & 39.8 \ua{1.02} & 3.06 & 2.04 && 26.25 \da{13.75} & 3.75 & 17.5 && 17.33 \da{2.67} & 0.0 & 2.67 && 20.0 \da{3.33} & 0.0 & 3.33 && 6.67 \da{6.66} & 0.0 & 6.67 && 0.22 \da{5.0} \\ 
        \reflexion & 31.63 \ua{5.1} & 14.29 & 9.18 && 63.75 \ua{3.75} & 6.25 & 2.5 && 17.33 \ua{8.0} & 14.67 & 6.67 && 10.0 \da{3.33} & 0.0 & 3.33 && 13.33 \ua{6.66} & 6.67 & 0.0 && 0.27 \ua{4.0} \\ 
        \ourmethod & 87.76  & 12.24 & 12.24 && 100.0  & 0.0 & 0.0 && 36.0 \ua{2.67} & 4.0 & 1.33 && 33.33 \ua{3.33} & 6.67 & 3.33 && 20.0  & 3.33 & 3.33 && 0.55 \ua{1.0} \\ 
        \midrule
        & \multicolumn{21}{c}{\textbf{\gptfouro}} \\
        \midrule
        \initial & 17.35 & - & - && 86.25 & - & - &&41.33 & - & - &&13.33 & - & - &&10.0 & - & - && 0.34 \\ 
        \hdashline\noalign{\vskip 0.4ex}
        \vanilla & 7.14 \da{12.25} & 2.04 & 14.29 && 86.25 \ua{5.0} & 10.0 & 5.0 && 26.67  & 16.0 & 16.0 && 10.0 \da{3.33} & 0.0 & 3.33 && 10.0 \ua{3.33} & 3.33 & 0.0 && 0.28 \da{1.0} \\ 
        \refine & 29.59 \da{4.08} & 10.2 & 14.29 && 88.75 \ua{10.0} & 12.5 & 2.5 && 40.0 \ua{1.33} & 18.67 & 17.33 && 20.0  & 3.33 & 3.33 && 10.0  & 0.0 & 0.0 && 0.38 \ua{2.0} \\ 
        \tick & 16.33 \da{14.28} & 0.0 & 14.29 && 70.0  & 2.5 & 2.5 && 20.0 \da{10.67} & 0.0 & 10.67 && 23.33 \ua{6.66} & 6.67 & 0.0 && 6.67 \da{3.33} & 0.0 & 3.33 && 0.27 \da{5.0} \\ 
        \reflexion & 30.61 \da{6.12} & 8.16 & 14.29 && 81.25 \da{1.25} & 3.75 & 5.0 && 32.0 \ua{6.67} & 13.33 & 6.67 && 20.0 \ua{3.33} & 3.33 & 0.0 && 13.33 \ua{3.33} & 3.33 & 0.0 && 0.35 \ua{1.0} \\ 
        \ourmethod & 38.78 \ua{1.02} & 6.12 & 5.1 && 100.0 & 0.0 & 0.0 && 64.0 \da{1.33} & 2.67 & 4.0 && 36.67 \ua{3.34} & 13.33 & 10.0 && 23.33 \ua{6.66} & 6.67 & 0.0 && 0.53 \ua{2.0} \\ 
        \midrule
        & \multicolumn{21}{c}{\textbf{\othreemini}} \\
        \midrule
        \initial & 86.73 & - & - && 90.0 & - & - &&34.67 & - & - &&80.0 & - & - &&73.33 & - & - && 0.73 \\ 
        \hdashline\noalign{\vskip 0.4ex}
        \vanilla & 84.69 \ua{1.02} & 6.12 & 5.1 && 85.0 \da{5.0} & 6.25 & 11.25 && 30.67 \da{1.33} & 2.67 & 4.0 && 83.33 \ua{3.33} & 3.33 & 0.0 && 80.0 \ua{3.33} & 3.33 & 0.0 && 0.73 \ua{1.0} \\ 
            \refine & 89.8 \ua{3.07} & 8.16 & 5.1 && 86.25 \da{1.25} & 8.75 & 10.0 && 25.33 \ua{5.33} & 5.33 & 0.0 && 86.67 \ua{3.34} & 3.33 & 0.0 && 73.33  & 6.67 & 6.67 && 0.72 \ua{2.0} \\ 
            \tick & 0.0  & 0.0 & 0.0 && 76.25 \da{11.25} & 3.75 & 15.0 && 13.33  & 1.33 & 1.33 && 76.67  & 0.0 & 0.0 && 63.33 \da{3.34} & 0.0 & 3.33 && 0.46 \da{3.0} \\ 
            \reflexion & 85.71 \ua{1.02} & 8.16 & 7.14 && 96.25 \da{1.25} & 1.25 & 2.5 && 30.67 \da{1.33} & 4.0 & 5.33 && 83.33 \ua{3.33} & 3.33 & 0.0 && 73.33 \ua{6.66} & 6.67 & 0.0 && 0.74 \ua{2.0} \\ 
            \ourmethod & 91.84 \ua{3.06} & 3.06 & 0.0 && 96.25 \da{1.25} & 0.0 & 1.25 && 38.67 \ua{1.34} & 2.67 & 1.33 && 86.67  & 0.0 & 0.0 && 83.33 \ua{3.33} & 10.0 & 6.67 && 0.79 \ua{1.0} \\ 
            \midrule
            & \multicolumn{21}{c}{\textbf{\deepseekrone}} \\
            \midrule
            \initial & 84.69 & - & - && 97.5 & - & - &&17.33 & - & - &&80.0 & - & - &&63.33 & - & - && 0.69 \\ 
            \hdashline\noalign{\vskip 0.4ex}
            \vanilla & 65.31 \ua{1.02} & 13.27 & 12.24 && 95.0 \ua{1.25} & 5.0 & 3.75 && 20.0 \ua{4.0} & 10.67 & 6.67 && 70.0 \da{6.67} & 3.33 & 10.0 && 63.33  & 3.33 & 3.33 && 0.63  \\ 
            \refine & 35.71 \da{16.33} & 2.04 & 18.37 && 91.25 \ua{2.5} & 8.75 & 6.25 && 13.33 \da{2.67} & 5.33 & 8.0 && 63.33 \da{13.34} & 6.67 & 20.0 && 66.67 \da{3.33} & 13.33 & 16.67 && 0.54 \da{7.0} \\ 
            \tick & 7.14 \da{10.21} & 0.0 & 10.2 && 91.25  & 1.25 & 1.25 && 0.0 \da{5.33} & 0.0 & 5.33 && 33.33 \da{26.67} & 6.67 & 33.33 && 40.0 \da{13.33} & 0.0 & 13.33 && 0.34 \da{11.0} \\ 
            \reflexion & 35.71 \da{14.29} & 8.16 & 22.45 && 85.0 \da{5.0} & 5.0 & 10.0 && 10.67 \da{8.0} & 4.0 & 12.0 && 66.67 \ua{10.0} & 10.0 & 0.0 && 43.33 \da{16.67} & 3.33 & 20.0 && 0.48 \da{7.0} \\ 
            \ourmethod & 86.73 \ua{1.02} & 2.04 & 1.02 && 100.0  & 0.0 & 0.0 && 22.67 \ua{2.67} & 6.67 & 4.0 && 83.33 \ua{3.33} & 6.67 & 3.33 && 73.33  & 3.33 & 3.33 && 0.73 \ua{1.0} \\ 
            \bottomrule
    \end{tabular}
}
\end{table}
\begin{table}[!ht]
\centering
\caption{The second attempt of self-correction performance on \game, \wordsorting, \checkmate, \aimefour, \aimefive with \textit{small} models \qwen and \llama. Green ({\scriptsize\textcolor{ggreen}{\footnotesize $\uparrow$}}) and red ({\scriptsize\textcolor{gred}{\footnotesize $\downarrow$}}) arrows
indicate performance changes against the previous attempt (i.e., \initial). \textbf{Bold} corresponds to the best performance.}
\label{tab:t2_delta_small}
\resizebox{\textwidth}{!}{
\setlength{\tabcolsep}{1pt}
    \begin{tabular}{llccllccllccllccllccla}
    \toprule
    \multirow{2}{*}{\textbf{Method}} & \multicolumn{3}{c}{\bf Game of 24} && \multicolumn{3}{c}{\bf Word Sorting} && \multicolumn{3}{c}{\bf CheckmateInOne} && \multicolumn{3}{c}{\bf AIME 2024} && \multicolumn{3}{c}{\bf AIME 2025} && \multicolumn{1}{c}{\bf Mean} \cr
    \cmidrule(lr){2-4} \cmidrule(lr){6-8} \cmidrule(lr){10-12} \cmidrule(lr){14-16} \cmidrule(lr){18-20}
    & \makecell{\textbf{Acc@t2}} & \makecell{$\underset{i \to c}{\Delta}(t_1, t_2)$} & \makecell{$\underset{c \to i}{\Delta}(t_1, t_2)$} && 
      \makecell{\textbf{Acc@t2}} & \makecell{$\underset{i \to c}{\Delta}(t_1, t_2)$} & \makecell{$\underset{c \to i}{\Delta}(t_1, t_2)$} &&
      \makecell{\textbf{Acc@t2}} & \makecell{$\underset{i \to c}{\Delta}(t_1, t_2)$} & \makecell{$\underset{c \to i}{\Delta}(t_1, t_2)$} &&
      \makecell{\textbf{Acc@t2}} & \makecell{$\underset{i \to c}{\Delta}(t_1, t_2)$} & \makecell{$\underset{c \to i}{\Delta}(t_1, t_2)$} &&
      \makecell{\textbf{Acc@t2}} & \makecell{$\underset{i \to c}{\Delta}(t_1, t_2)$} & \makecell{$\underset{c \to i}{\Delta}(t_1, t_2)$} &&
      \makecell{\textbf{Acc@t2}} \\
    \midrule
    & \multicolumn{21}{c}{\textbf{\qwen}} \\
    \midrule
    \initial & 8.16 & - & - && 13.75 & - & - &&2.67 & - & - &&20.0 & - & - &&10.0 & - & - && 0.11 \\ 
            \hdashline\noalign{\vskip 0.4ex}
            \vanilla & 7.14 \ua{1.02} & 4.08 & 3.06 && 12.5 \da{3.75} & 1.25 & 5.0 && 0.0  & 0.0 & 0.0 && 16.67  & 3.33 & 3.33 && 6.67 \da{3.33} & 3.33 & 6.67 && 0.09 \da{1.0} \\ 
            \refine & 7.14 \ua{1.02} & 2.04 & 1.02 && 26.25 \ua{2.5} & 7.5 & 5.0 && 0.0 \da{1.33} & 0.0 & 1.33 && 23.33 \ua{3.33} & 6.67 & 3.33 && 3.33  & 0.0 & 0.0 && 0.12 \ua{1.0} \\ 
            \tick & 0.0  & 0.0 & 0.0 && 13.75 \da{3.75} & 3.75 & 7.5 && 0.0  & 0.0 & 0.0 && 10.0  & 0.0 & 0.0 && 6.67  & 0.0 & 0.0 && 0.06 \da{1.0} \\ 
            \reflexion & 10.2 \da{1.02} & 4.08 & 5.1 && 17.5 \da{3.75} & 3.75 & 7.5 && 2.67  & 1.33 & 1.33 && 16.67 \ua{3.34} & 3.33 & 0.0 && 10.0 \da{3.33} & 0.0 & 3.33 && 0.11 \da{1.0} \\ 
            \ourmethod & 12.24 \ua{1.02} & 1.02 & 0.0 && 61.25 \da{5.0} & 5.0 & 10.0 && 5.33 \ua{1.33} & 1.33 & 0.0 && 23.33 \ua{3.33} & 3.33 & 0.0 && 13.33 \ua{3.33} & 3.33 & 0.0 && 0.23 \ua{1.0} \\ 
            \ourmethodsmall & 52.04 \ua{10.2} & 23.47 & 13.27 && 32.5 \da{16.25} & 8.75 & 25.0 && 25.33 \ua{14.66} & 18.67 & 4.0 && 26.67 \ua{3.34} & 3.33 & 0.0 && 16.67 \ua{3.34} & 3.33 & 0.0 && 0.31 \ua{3.0} \\
            \midrule
            & \multicolumn{21}{c}{\textbf{\llama}} \\
            \midrule
            \initial & 19.39 & - & - && 75.0 & - & - &&8.0 & - & - &&33.33 & - & - &&3.33 & - & - && 0.28 \\ 
            \hdashline\noalign{\vskip 0.4ex}
            \vanilla & 62.24 \ua{19.38} & 24.49 & 5.1 && 80.0 \ua{2.5} & 7.5 & 5.0 && 9.33 \ua{8.0} & 8.0 & 0.0 && 36.67  & 0.0 & 0.0 && 3.33  & 0.0 & 0.0 && 0.38 \ua{6.0} \\ 
            \refine & 42.86 \ua{9.19} & 17.35 & 8.16 && 71.25 \da{5.0} & 5.0 & 10.0 && 2.67 \da{2.66} & 0.0 & 2.67 && 40.0  & 0.0 & 0.0 && 6.67  & 0.0 & 0.0 && 0.33 \ua{1.0} \\ 
            \tick & 11.22 \da{3.07} & 4.08 & 7.14 && 63.75 \da{7.5} & 1.25 & 8.75 && 5.33 \da{1.34} & 0.0 & 1.33 && 33.33 \ua{3.33} & 3.33 & 0.0 && 10.0 \ua{3.33} & 3.33 & 0.0 && 0.25 \da{1.0} \\ 
            \reflexion & 45.92 \ua{22.45} & 25.51 & 3.06 && 72.5 \da{3.75} & 5.0 & 8.75 && 6.67 \ua{1.34} & 4.0 & 2.67 && 30.0 \ua{3.33} & 3.33 & 0.0 && 3.33  & 0.0 & 0.0 && 0.32 \ua{5.0} \\ 
            \ourmethod & 71.43 \ua{7.14} & 12.24 & 5.1 && 98.75  & 0.0 & 0.0 && 4.0 \ua{1.33} & 1.33 & 0.0 && 50.0 \ua{13.33} & 16.67 & 3.33 && 16.67  & 0.0 & 0.0 && 0.48 \ua{4.0} \\ 
            \ourmethodsmall & 100.0  & 0.0 & 0.0 && 98.75 \da{1.25} & 0.0 & 1.25 && 50.67 \ua{12.0} & 22.67 & 10.67 && 50.0 \ua{3.33} & 3.33 & 0.0 && 23.33  & 3.33 & 3.33 && 0.65 \ua{3.0} \\ 
            \bottomrule
    \end{tabular}
}
\end{table}




\subsection{Additional Results on Effect of Iterative Correction}
\label{sec:app:iter-effect}
\begin{figure}[H]
    \centering
    \begin{subfigure}{\textwidth}
        \centering
        \begin{subfigure}{0.24\textwidth}
            \includegraphics[width=\linewidth]{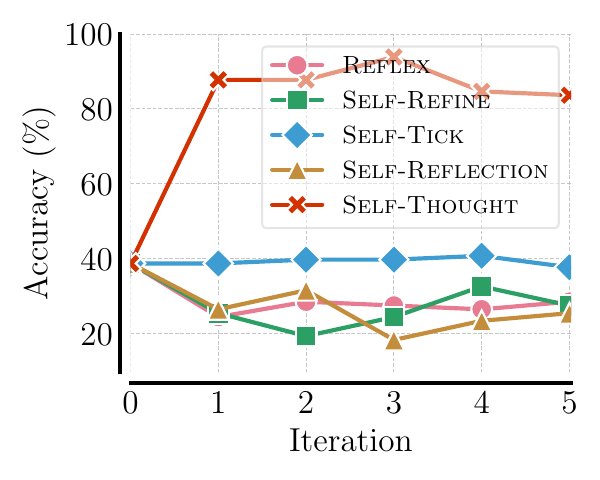}
            \subcaption{\gptfouromini}
        \end{subfigure}
        \begin{subfigure}{0.24\textwidth}
            \includegraphics[width=\linewidth]{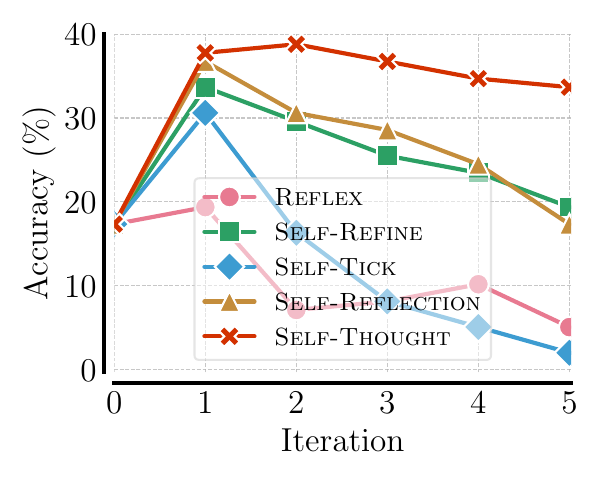}
            \subcaption{\gptfouro}
        \end{subfigure}
        \begin{subfigure}{0.24\textwidth}
            \includegraphics[width=\linewidth]{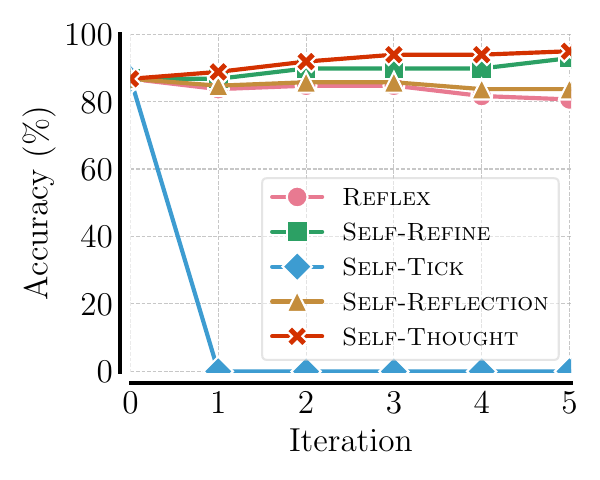}
            \subcaption{\othreemini}
        \end{subfigure}
        \begin{subfigure}{0.24\textwidth}
            \includegraphics[width=\linewidth]{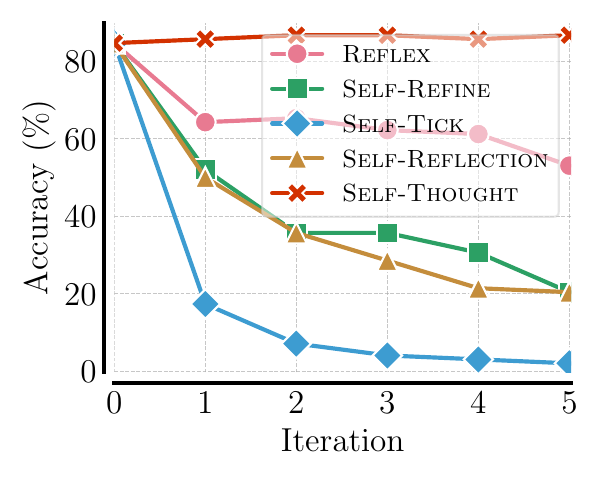}
            \subcaption{\deepseekrone}
        \end{subfigure}
        \subcaption{\game}
    \end{subfigure}
    \begin{subfigure}{\textwidth}
        \centering
        \begin{subfigure}{0.24\textwidth}
            \includegraphics[width=\linewidth]{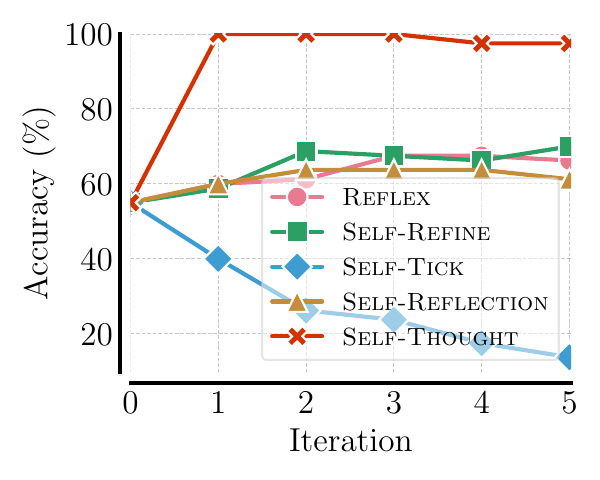}
            \subcaption{\gptfouromini}
        \end{subfigure}
        \begin{subfigure}{0.24\textwidth}
            \includegraphics[width=\linewidth]{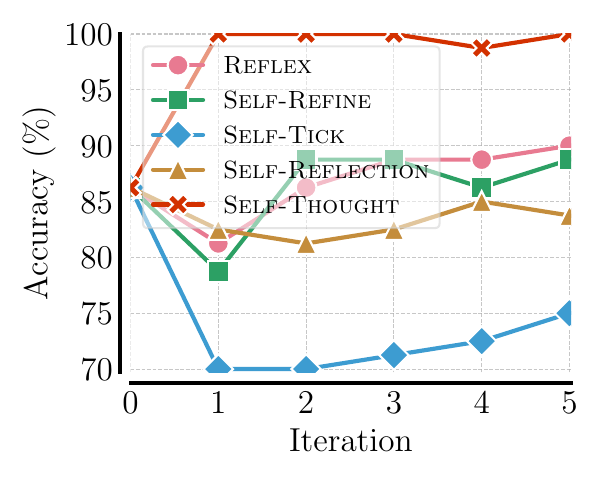}
            \subcaption{\gptfouro}
        \end{subfigure}
        \begin{subfigure}{0.24\textwidth}
            \includegraphics[width=\linewidth]{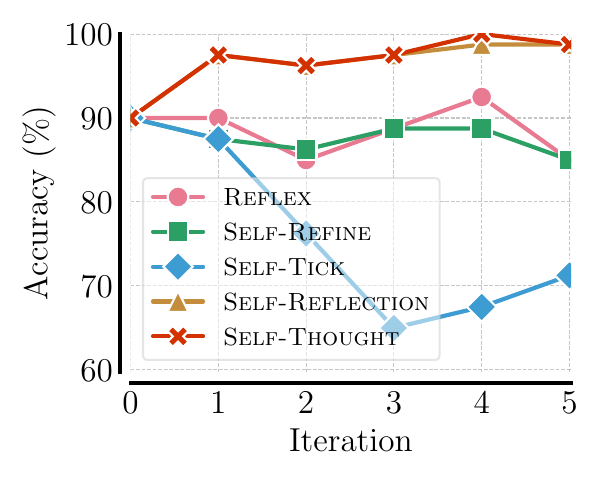}
            \subcaption{\othreemini}
        \end{subfigure}
        \begin{subfigure}{0.24\textwidth}
            \includegraphics[width=\linewidth]{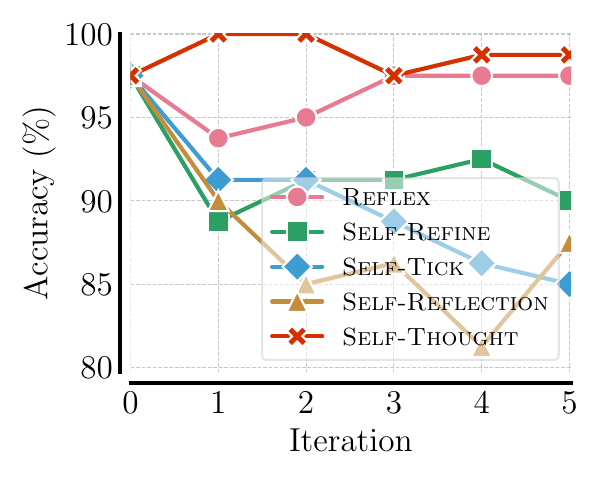}
            \subcaption{\deepseekrone}
        \end{subfigure}
        \subcaption{\wordsorting}
    \end{subfigure}
    \begin{subfigure}{\textwidth}
        \centering
        \begin{subfigure}{0.24\textwidth}
            \includegraphics[width=\linewidth]{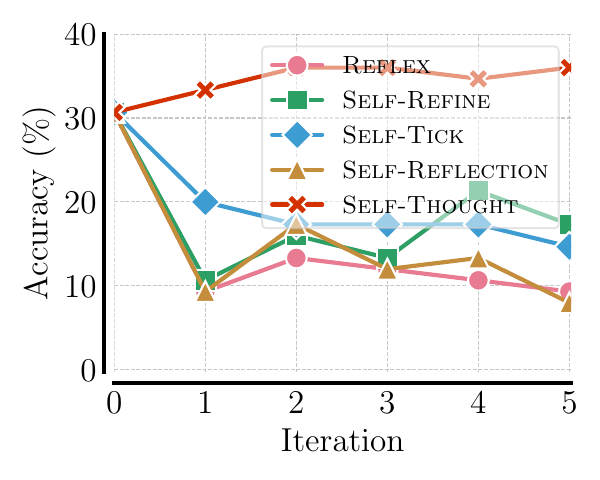}
            \subcaption{\gptfouromini}
        \end{subfigure}
        \begin{subfigure}{0.24\textwidth}
            \includegraphics[width=\linewidth]{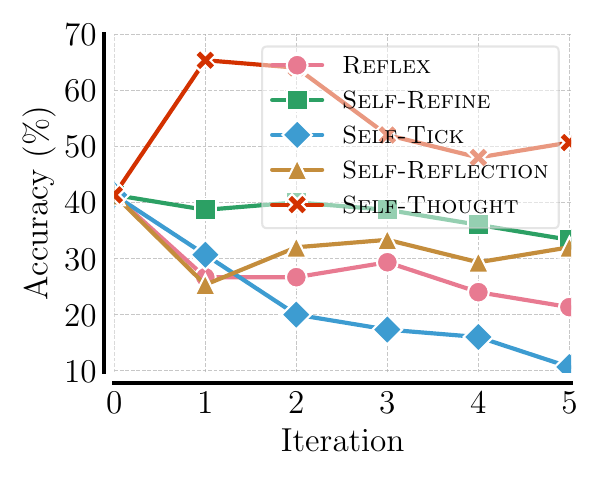}
            \subcaption{\gptfouro}
        \end{subfigure}
        \begin{subfigure}{0.24\textwidth}
            \includegraphics[width=\linewidth]{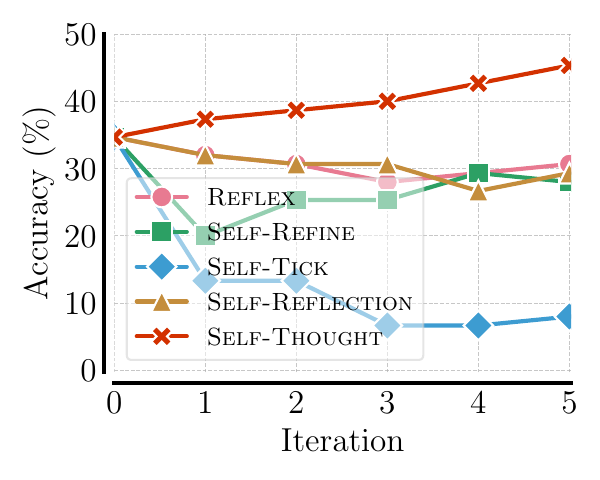}
            \subcaption{\othreemini}
        \end{subfigure}
        \begin{subfigure}{0.24\textwidth}
            \includegraphics[width=\linewidth]{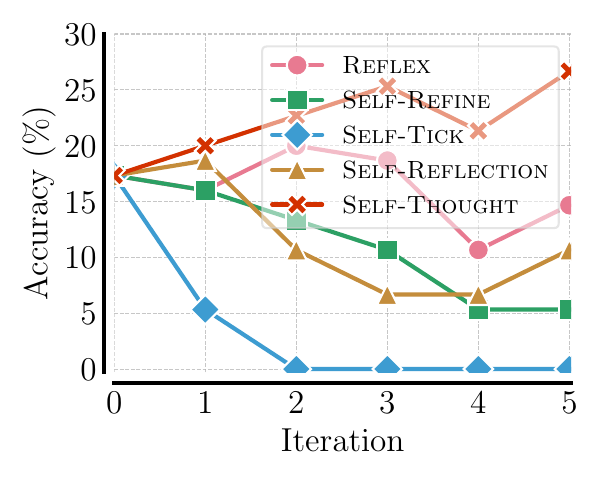}
            \subcaption{\deepseekrone}
        \end{subfigure}
        \subcaption{\checkmate}
    \end{subfigure}
    \begin{subfigure}{\textwidth}
        \centering
        \begin{subfigure}{0.24\textwidth}
            \includegraphics[width=\linewidth]{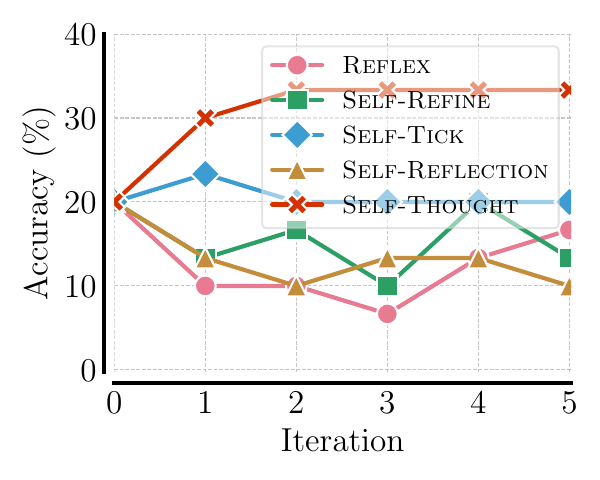}
            \subcaption{\gptfouromini}
        \end{subfigure}
        \begin{subfigure}{0.24\textwidth}
            \includegraphics[width=\linewidth]{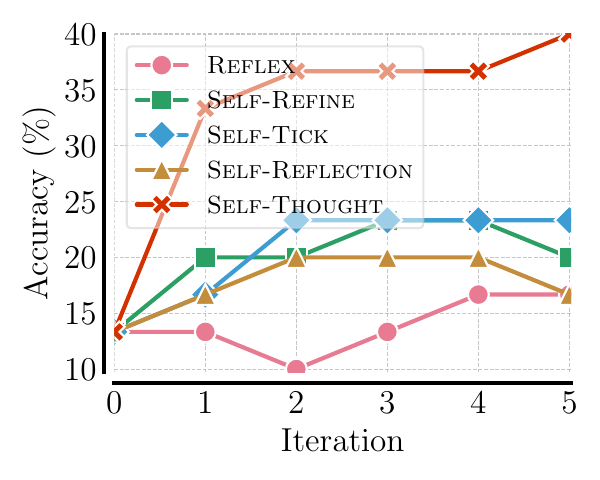}
            \subcaption{\gptfouro}
        \end{subfigure}
        \begin{subfigure}{0.24\textwidth}
            \includegraphics[width=\linewidth]{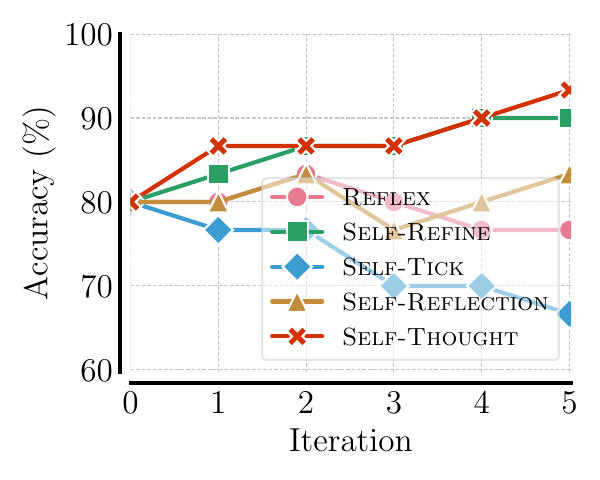}
            \subcaption{\othreemini}
        \end{subfigure}
        \begin{subfigure}{0.24\textwidth}
            \includegraphics[width=\linewidth]{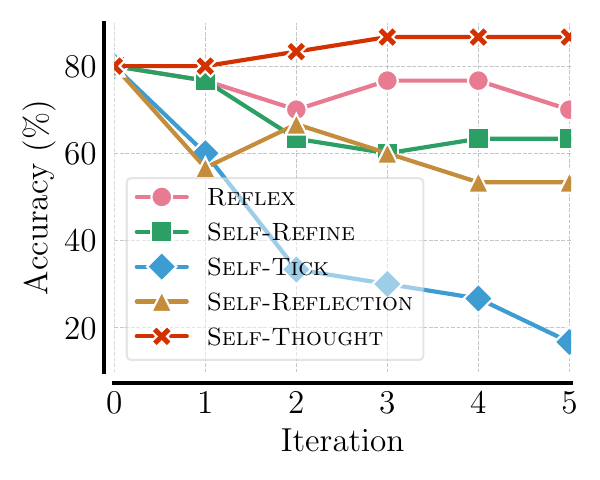}
            \subcaption{\deepseekrone}
        \end{subfigure}
        \subcaption{\aimefour}
    \end{subfigure}
    \caption{Accuracy over iterations with self-correction methods across models.}
    \label{fig:app:iter-large}
\end{figure}
\begin{figure}[H]
    \centering
    \begin{subfigure}{\textwidth}
        \centering
        \begin{subfigure}{0.27\textwidth}
            \includegraphics[width=\linewidth]{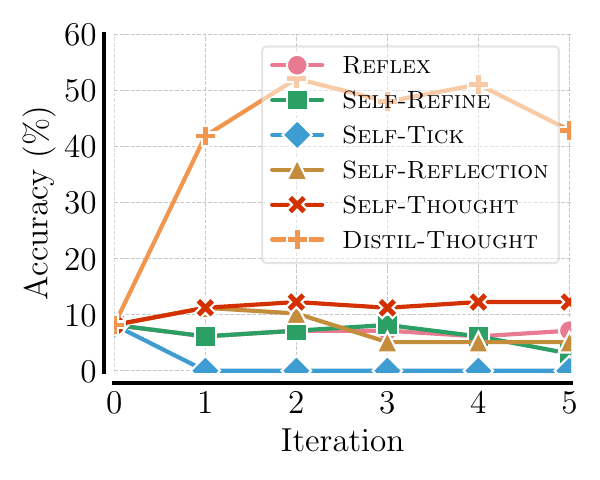}
            \subcaption{\qwen}
        \end{subfigure}
        \begin{subfigure}{0.27\textwidth}
            \includegraphics[width=\linewidth]{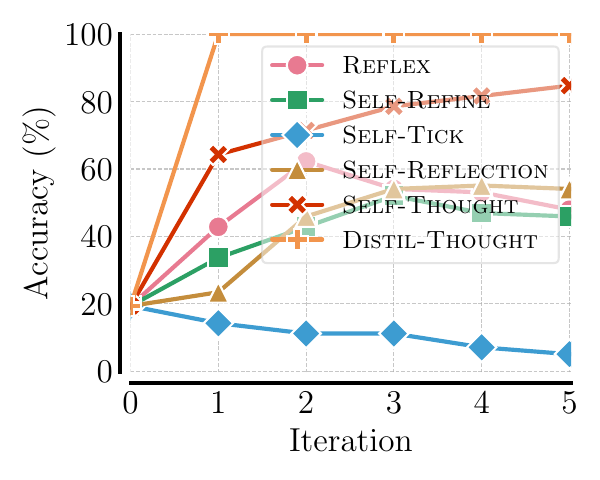}
            \subcaption{\llama}
        \end{subfigure}
        \subcaption{\game}
    \end{subfigure}
    \begin{subfigure}{\textwidth}
        \centering
        \begin{subfigure}{0.27\textwidth}
            \includegraphics[width=\linewidth]{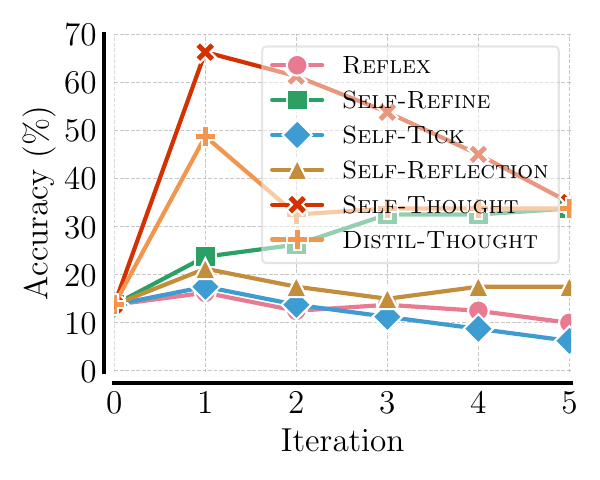}
            \subcaption{\qwen}
        \end{subfigure}
        \begin{subfigure}{0.27\textwidth}
            \includegraphics[width=\linewidth]{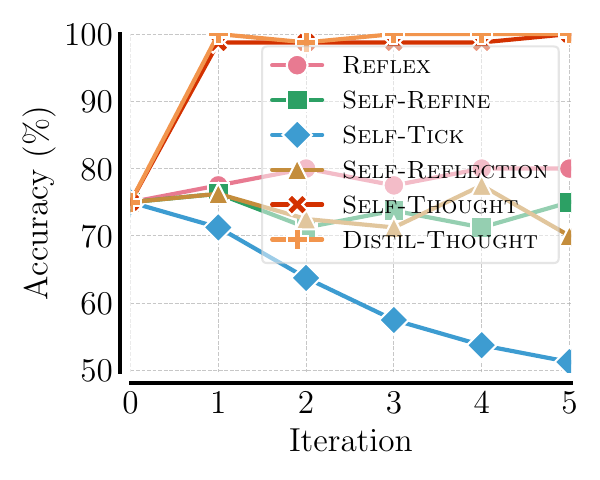}
            \subcaption{\llama}
        \end{subfigure}
        \subcaption{\wordsorting}
    \end{subfigure}
    \begin{subfigure}{\textwidth}
        \centering
        \begin{subfigure}{0.27\textwidth}
            \includegraphics[width=\linewidth]{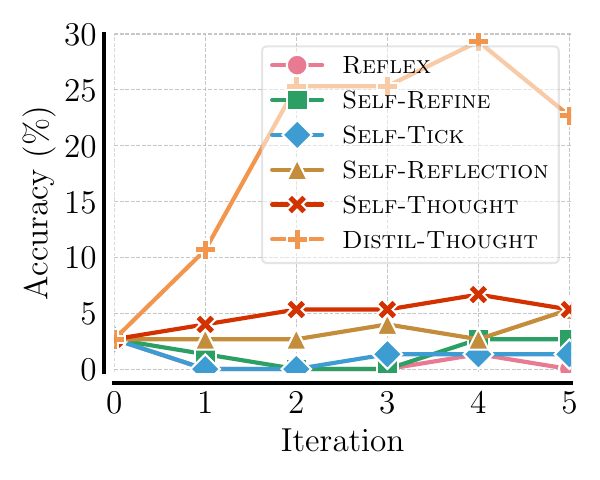}
            \subcaption{\qwen}
        \end{subfigure}
        \begin{subfigure}{0.27\textwidth}
            \includegraphics[width=\linewidth]{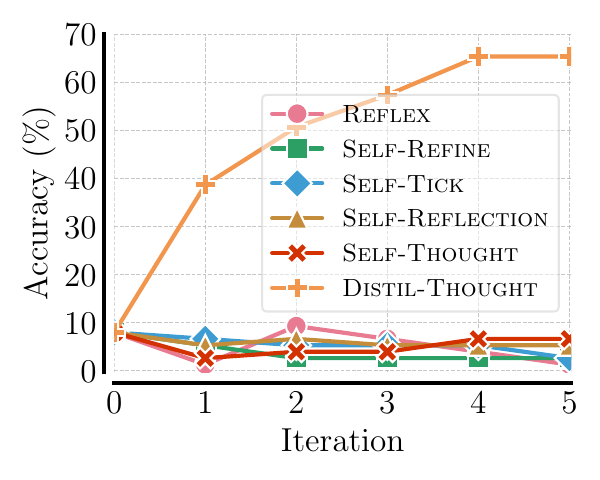}
            \subcaption{\llama}
        \end{subfigure}
        \subcaption{\checkmate}
    \end{subfigure}
    \caption{Accuracy over iterations with self-correction methods across models.}
    \label{fig:app:iter-small}
\end{figure}



\section{Results from Baseline Studies}
\label{sec:app:baseline-results}

\subsection{Additional Analysis on Results from \refine}
\begin{table}[H]
\centering
\caption{\refine results on various tasks using GPT-3.5, ChatGPT, and GPT-4 as base LLM. While \refine achieves substantial improvements on general tasks such as Dialogue Response Generation, Sentiment Reversal, and Acronym Generation, its gains on reasoning tasks are more modest. Results reported from Table 1 in \cite{madaan2023self-refine}.}
\label{tab:self-refine-results}
\resizebox{0.9\textwidth}{!}{
    \begin{tabular}{@{}lcccccc@{}}
    \toprule
     & \multicolumn{2}{c}{GPT-3.5} & \multicolumn{2}{c}{ChatGPT} & \multicolumn{2}{c}{GPT-4} \\
    \cmidrule(lr){2-3} \cmidrule(lr){4-5} \cmidrule(lr){6-7}
    Task & Base & +SELF-REFINE & Base & +SELF-REFINE & Base & +SELF-REFINE \\
    \midrule
    Sentiment Reversal & 8.8 & \textbf{30.4} ($\uparrow$21.6) & 11.4 & \textbf{43.2} ($\uparrow$31.8) & 3.8 & \textbf{36.2} ($\uparrow$32.4) \\
    Dialogue Response & 36.4 & \textbf{63.6} ($\uparrow$27.2) & 40.1 & \textbf{59.9} ($\uparrow$19.8) & 25.4 & \textbf{74.6} ($\uparrow$49.2) \\
    Code Optimization & 14.8 & \textbf{23.0} ($\uparrow$8.2) & 23.9 & \textbf{27.5} ($\uparrow$3.6) & 27.3 & \textbf{36.0} ($\uparrow$8.7) \\
    Code Readability & 37.4 & \textbf{51.3} ($\uparrow$13.9) & 27.7 & \textbf{63.1} ($\uparrow$35.4) & 27.4 & \textbf{56.2} ($\uparrow$28.8) \\
    \rowcolor{red!20} Math Reasoning & 64.1 & \textbf{64.1} (0) & 74.8 & \textbf{75.0} ($\uparrow$0.2) & 92.9 & \textbf{93.1} ($\uparrow$0.2) \\
    Acronym Generation & 41.6 & \textbf{56.4} ($\uparrow$14.8) & 27.2 & \textbf{37.2} ($\uparrow$10.0) & 30.4 & \textbf{56.0} ($\uparrow$25.6) \\
    Constrained Generation & 28.0 & \textbf{37.0} ($\uparrow$9.0) & 44.0 & \textbf{67.0} ($\uparrow$23.0) & 15.0 & \textbf{45.0} ($\uparrow$30.0) \\
    \bottomrule
    \end{tabular}%
}
\end{table}

Table~\ref{tab:self-refine-results} shows results from \refine~\citep{madaan2023self-refine}. These results indicate that \refine achieves substantial gains on preference-based tasks such as Dialogue Response Generation, Sentiment Reversal, and Acronym Generation. However, its performance improvements on reasoning tasks are more modest, which can be attributed to the limited ability of the model to accurately identify errors. Moreover, the gains on Math Reasoning increase by only $5\%$ when an external source is available to indicate whether the current answer is incorrect (See results in Appendix H.1 from \cite{madaan2023self-refine}).

\subsection{Additional Analysis on Results from \tick}
\begin{table}[H]
\centering
\caption{\tick results on a single step of self-refinement on different tasks with Command-R+ and GPT-4o. \tick consistently improves overall performance compared to both base models and \refine, with modest gains on reasoning-related tasks. Results reported from Table 1 in \cite{cook2024self-tick}.}
\label{tab:self-tick-results}
\resizebox{0.9\textwidth}{!}{
    \begin{tabular}{@{}lcccccc@{}}
    \toprule
     & \multicolumn{3}{c}{Command-R+} & \multicolumn{3}{c}{GPT-4o} \\ 
    \cmidrule(lr){2-4} \cmidrule(lr){5-7}
    Tasks & Base & \refine & \tick & Base & \refine & \tick \\ 
    \midrule
    Overall & 32.0 & 23.7 ($\downarrow$ 8.3) & \textbf{35.8} ($\uparrow$ 3.8) & 55.4 & 47.1 ($\downarrow$ 8.3) & \textbf{56.2} ($\uparrow$ 0.8) \\
    Coding & 18.8 & 9.1 ($\downarrow$ 9.7) & \textbf{22.7} ($\uparrow$ 3.9) & 50.4 & 36.4 ($\downarrow$ 14.0) & \textbf{51.6} ($\uparrow$ 1.2) \\
    Data Analysis & 25.9 & 5.3 ($\downarrow$ 20.6) & \textbf{29.8} ($\uparrow$ 3.9) & 52.4 & 27.2 ($\downarrow$ 25.2) & \textbf{52.5} ($\uparrow$ 0.1) \\
    Instructions & 69.6 & 60.5 ($\downarrow$ 9.1) & \textbf{75.8} ($\uparrow$ 6.2) & 73.3 & 62.8 ($\downarrow$ 10.5) & \textbf{76.2} ($\uparrow$ 2.9) \\
    \rowcolor{red!20}
    Language & \textbf{24.6} & 13.8 ($\downarrow$ 9.8) & 24.1 ($\downarrow$ 0.5) & 50.9 & \textbf{51.4} ($\uparrow$ 0.5) & 50.4 ($\downarrow$ 0.5) \\
    \rowcolor{red!20}
    Mathematics & 23.7 & 23.6 ($\downarrow$ 0.1) & \textbf{25.5} ($\uparrow$ 1.8) & 52.3 & 51.8 ($\downarrow$ 0.5) & \textbf{53.1} ($\uparrow$ 0.8) \\
    \rowcolor{red!20}
    Reasoning & 29.2 & 30.0 ($\uparrow$ 0.8) & \textbf{37.0} ($\uparrow$ 7.8) & \textbf{53.3} & 52.7 ($\downarrow$ 0.6) & 53.3 (0) \\
    \bottomrule
    \end{tabular}
}
\end{table}
Table~\ref{tab:self-tick-results} shows results from \tick \citep{cook2024self-tick} for a single step of self-refinement on various tasks with Command-R+ and GPT-4o. The results indicate that \tick consistently improves overall performance compared to both the base models and \refine, with the largest gains observed in preference-based and instruction-following tasks. For example, improvements on Coding, Data Analysis, and Instructions range from $1.2\%$ to $6.2\%$ across the models. In contrast, gains on reasoning-related tasks such as Language, Mathematics, and Reasoning are more modest, highlighting that even with \tick, these tasks remain challenging.
\section{Quantitative Results}
\begin{longtable}{p{0.96\textwidth}}
\caption{Success case for \ourmethod on \aimefour using \gptfouromini: \textbf{Correcting missing logic in initial response}} \label{tab:prompt_routing} \\
\toprule
\textbf{Task} \\
\midrule
Jen enters a lottery by picking $4$ distinct numbers from $S={1,2,3,\cdots,9,10}$. $4$ numbers are randomly chosen from $S.$ She wins a prize if at least two of her numbers were $2$ of the randomly chosen numbers, and wins the grand prize if all four of her numbers were the randomly chosen numbers. The probability of her winning the grand prize given that she won a prize is $\tfrac{m}{n}$ where $m$ and $n$ are relatively prime positive integers. Find $m+n$. \\
\midrule
\textbf{Initial Answer} \\
\midrule
Let $S = \{1, 2, \dots, 10\}$. The lottery chooses $4$ distinct numbers. Jen also chooses $4$ distinct numbers. \\
\\
We need to find the probability that Jen wins the grand prize (matches all 4) given that she wins some prize (matches at least 2).\\
\\
$P(\text{Grand Prize} \mid \text{Prize}) = \frac{\text{Ways to match 4}}{\text{Total ways to match 2, 3, or 4}}$ \\
\\
\textbf{Step 1: Count the winning combinations} \\
We calculate the number of ways Jen can match the winning numbers from the $4$ numbers selected by the lottery. \\
\\
\hlc{Matching exactly 2 numbers: \\
She needs to choose 2 numbers from the 4 winning numbers: $\binom{4}{2} = \frac{4 \times 3}{2} = 6$} \\
\\
\hlc{Matching exactly 3 numbers: \\
She needs to choose 3 numbers from the 4 winning numbers: $\binom{4}{3} = 4$} \\
\\
Matching exactly 4 numbers (Grand Prize): \\
She needs to choose all 4 winning numbers.
$\binom{4}{4} = 1$ \\
\\
\textbf{Step 2: Calculate the total winning scenarios} \\
\hlc{To find the total number of ways to win a prize, we simply sum the possibilities calculated above: \\
\\
$\text{Total Ways} = 6 + 4 + 1 = 11$} \\
\\
\textbf{Step 3: Calculate the probability} \\
The probability is the single way to win the Grand Prize divided by the total ways to win any prize: $\frac{1}{11}$ \\
\\
Thus, $m = 1$ and $n = 11$. These are relatively prime. \\
\\
The sum is: \\
\\
$m + n = 1 + 11 = 12$ \\
\\
\textless Answer\textgreater\ 12 \textless/Answer\textgreater \\
\midrule
\textbf{Task Abstraction} \\
\midrule
Distilled Information: \\
\\
1. Key information: \\   
- Set of numbers (S = {1, 2, 3, \ldots, 10}) \\  
- Jen picks (4) distinct numbers from (S). \\    
- A prize is won if at least (2) of Jen's numbers match (2) of the randomly chosen numbers. \\    
- The grand prize is won if all (4) of Jen's numbers match the randomly chosen numbers. \\
\\
2. Restriction: \\ 
- The probability of winning the grand prize must be calculated given that a prize has already been won (i.e., at least (2) matches). \\ 
- The total number of ways to choose (4) numbers from a set of (10) must be considered. \\ 
- The combinations must adhere to the rules of probability and combinatorics. \\
\\
3. Distilled task: \\ 
- Calculate the conditional probability ($P(\text{grand prize} \mid \text{prize})$). \\ 
- Identify the number of combinations that lead to winning the grand prize and those that lead to winning any prize. \\
\\
4. Python transformation: \\ 
Input parameters: \\ 
- total\_numbers = 10  (size of set (S)) \\ 
- numbers\_chosen = 4  (numbers Jen picks) \\ 
- prize\_condition = 2  (minimum matches for a prize) \\ 
- grand\_prize\_condition = 4  (exact matches for grand prize) \\
\\
5. Answer form: \\   
- The final answer should be in the form of a reduced fraction ($\frac{m}{n}$) where (m) and (n) are relatively prime positive integers. \\
\midrule
\textbf{Solution Instantiation} \\
\midrule
To solve the problem, we need to calculate the conditional probability $P(\text{grand prize} \mid \text{prize})$. \\
\\
Let's break down the steps needed to find this probability: \\
\\
1. \textbf{Total combinations}: The total number of ways to choose 4 numbers from the set (S = {1, 2, ..., 10}) is given by the binomial coefficient: $\binom{10}{4} = 210$ \\
\\
2. \textbf{Winning the grand prize}: Jen wins the grand prize if all 4 of her numbers match the 4 randomly drawn numbers. There is only 1 way for this to happen since there is only one specific set of 4 numbers that can be chosen out of her choices. Thus: Ways to win grand prize = 1 \\
\\
3. \textbf{Winning a prize}: Jen wins a prize if at least 2 of her numbers are among the 4 randomly chosen numbers. To find this, we will calculate the total ways to have at least 2 matches. \\
\\
We can find the winning scenarios by breaking it down into cases where she matches exactly 2, exactly 3, and all 4 of her numbers. \\
\\
\hlcorrect{- \textbf{Case 1}: Exactly 2 matches \\
- Choose 2 numbers from Jen's picks (which is ($\binom{4}{2} = 6$)). \\
- Choose 2 numbers from the remaining 6 numbers (which is ($\binom{6}{2} = 15$)). \\
- Total ways for this case: $6 * 15 = 90$} \\
\\
\hlcorrect{- \textbf{Case 2}: Exactly 3 matches \\
- Choose 3 numbers from Jen's picks (which is ($\binom{4}{3} = 4$)). \\
- Choose 1 number from the remaining 6 numbers (which is ($\binom{6}{1} = 6$)). \\
- Total ways for this case: $4 * 6 = 24$} \\
\\
- \textbf{Case 3}: Exactly 4 matches (grand prize) \\
- There is 1 way to choose all 4 numbers from Jen's picks. \\
- Total ways for this case: $1$ \\
\\
Adding these cases together, the total ways Jen can win a prize is: $90 + 24 + 1 = 115$ \\
\\
\hlcorrect{4. \textbf{Conditional probability}: We want to find the probability of winning the grand prize given that she has won a prize:
$P(\text{grand prize} \mid \text{prize}) = \frac{P(\text{grand prize})}{P(\text{prize})} = \frac{1}{115}$} \\
\\
Here, ($P(\text{grand prize}) = 1$) (the number of ways to win the grand prize) and ($P(\text{prize}) = 115$) (the number of ways to win a prize). \\
\\
5. \textbf{Final fraction}: The probability can be expressed as: $P(\text{grand prize} \mid \text{prize}) = \frac{1}{115}$ \\
\\
This fraction is in simplest form with (m = 1) and (n = 115). Thus, (m+n = 1 + 115 = 116). \\
\\
Therefore, the final answer is: \\
\\
\textless Answer\textgreater\ 116 \textless/Answer\textgreater \\
\bottomrule
\end{longtable}
\begin{longtable}{p{0.96\textwidth}}
\caption{Initial failure and success in second iteration for \ourmethod on \game using \gptfouromini: \textbf{Correcting logic and answer format in initial response}} \label{tab:prompt_routing2} \\
\toprule
\textbf{Task} \\
\midrule
4 8 11 13 \\
\midrule
\textbf{Initial Answer} \\
\midrule
One possible solution using the numbers 4, 8, 11, and 13 to achieve 24 is:\\
\\
\hlerror{$(13 - 11) \times 8 + 4 = 24$} \\
\\
Thus, the final answer is: \\
\\
\hlerror{\textless Answer\textgreater~24 \textless/Answer\textgreater} \\
\midrule
\textbf{Task Abstraction \#1} \\
\midrule
Distilled Information:\\
\\
1. Key information:\\
- Given integers: 4, 8, 11, 13\\
\\
2. Restriction:\\
- Each number must be used exactly once.\\
- Only the four basic arithmetic operations (addition, subtraction, multiplication, division) can be used.\\
- Parentheses can be used to change the order of operations.\\
- The objective is to achieve a total of 24.\\
\\
3. Distilled task: \\
- Find a combination of the numbers 4, 8, 11, and 13 using the allowed arithmetic operations to achieve the total of 24. \\
\\
4. Python transformation: \\
Input parameters: \\
- number1 = 4 \\
- number2 = 8 \\
- number3 = 11 \\
- number4 = 13 \\
\\
5. Answer form:
- A mathematical expression that results in 24 using the given integers and operations. 
\\
\hlerror{\textless/Answer\textgreater~24 \textless/Answer\textgreater} \\
\midrule
\textbf{Solution Instantiation \#1} \\
\midrule
One possible solution using the numbers 4, 8, 11, and 13 to achieve 24 is:\\
\\
\hlerror{$(13 - 11) \times 8 + 4 = 24$} \\
\\
Thus, the final answer is: \\
\\
\textless Answer\textgreater\ (13 - 11) * 8 + 4 \textless/Answer\textgreater \\
\midrule
\textbf{Task Abstraction \#2} \\
\midrule
Distilled Information:\\
\\
1. Key information:\\
- Given integers: 4, 8, 11, 13\\
\\
2. Restriction:\\
- Each number must be used exactly once.\\
- Only the arithmetic operations: addition (+), subtraction (-), multiplication (*), and division (/) are allowed.\\
- Parentheses can be used to dictate operation order.\\
- The objective is to achieve a total of 24.\\
\\
3. Distilled task:\\
- Find a valid mathematical expression using the numbers 4, 8, 11, and 13 that equals 24.\\
\\
4. Python transformation:\\
Input parameters:\\
number1 = 4\\
number2 = 8\\
number3 = 11\\
number4 = 13\\
\\
5. Answer form:\\
- The output should be a mathematical expression that results in 24 using the provided numbers. \\
\midrule
\textbf{Solution Instantiation \#2} \\
\midrule
One possible solution using the numbers 4, 8, 11, and 13 to achieve a total of 24 is:\\
\\
\hlcorrect{$(13 - 11) \times (8 + 4) = 2 \times 12 = 24$} \\
\\
Thus, the final answer is: \\
\\
\textless Answer\textgreater\ ((13 - 11) * (8 + 4)) \textless/Answer\textgreater \\
\bottomrule
\end{longtable}
\begin{longtable}{p{0.96\textwidth}}
\caption{Initial failure and success in second iteration for \ourmethod on \game using \gptfouromini: \textbf{Correcting logic in initial response and the first correction iteration}} \label{tab:prompt_routing2} \\
\toprule
\textbf{Task} \\
\midrule
2 3 4 7 \\
\midrule
\textbf{Initial Answer} \\
One possible solution using the numbers 2, 3, 4, and 7 to achieve 24 is:\\
\\
\hlerror{$(7 - (3 / 2)) \times 4 = 24$} \\
\\
Thus, the final answer is: \\
\\
\hlerror{\textless Answer\textgreater~(7 - (3 / 2)) * 4 = 24 \textless/Answer\textgreater} \\
\midrule
\textbf{Task Abstraction \#1} \\
\midrule
Distilled Information: \\
\\
1. Key information:\\
- Given integers: 2, 3, 4, 7\\
\\
2. Restriction:\\
- Each number must be used exactly once.\\
- Only the arithmetic operations of addition (+), subtraction (-), multiplication (*), and division (/) can be employed.\\
- Parentheses can be used to dictate the order of operations.\\
- The final result must equal 24.\\
\\
3. Distilled task:\\
- Find a combination of the four given integers using arithmetic operations and parentheses to achieve a total of 24.\\
\\
4. Python transformation:\\
Input parameters:\\
- number1 = 2\\
- number2 = 3\\
- number3 = 4\\
- number4 = 7\\
\\
5. Answer form:\\
- The solution should be presented in a mathematical expression format that equals 24.\\ 
\\
\textless Answer\textgreater\ The solution must be in the specified mathematical expression format. \textless/Answer\textgreater \\
\midrule
\textbf{Solution Instantiation \#1} \\
\midrule
One possible solution using the numbers 2, 3, 4, and 7 to achieve a total of 24 is:\\
\\
$(7 - 2) \times (4 + 3)$ \\
\\
Calculating this step-by-step: \\
1. ($7 - 2 = 5$) \\
2. ($4 + 3 = 7$) \\
3. ($5 \times 7 = 35$) \\
\\
This does not yield 24, so let's try again. \\
\\
Another approach is:\\
\\
$(4 \times 7) - (3 \times 2)$ \\
\\
Calculating this: \\
1. ($4 \times 7 = 28$)\\
2. ($3 \times 2 = 6$)\\
3. ($28 - 6 = 22$)\\
\\
Again, this does not yield 24. \\
\\
After trying several combinations, here's a correct expression that works:\\
\\
$(3 \times 7) + (4 - 2)$ \\
\\
Calculating this: \\
1. ($3 \times 7 = 21$)\\
2. ($4 - 2 = 2$)\\
3. ($21 + 2 = 23$)\\
\\
Let's try:\\
\\
$7 \times 4 - 2 \times 3$ \\
\\
Calculating this: \\
1. ($7 \times 4 = 28$)\\
2. ($2 \times 3 = 6$)\\
3. ($28 - 6 = 22$)\\
\\
Finally, let's use: \\
\\
$(7 - 2) \times 4 - 3$ \\
\\
Calculating this: \\
1. ($7 - 2 = 5$) \\
2. ($5 \times 4 = 20$) \\
3. ($20 - 3 = 17$) \\
\\
After further trials, we arrive at:\\
\\
$(4 \times (7 - 2)) - 3)$ \\
\\
Calculating this: \\
1. ($7 - 2 = 5$) \\
2. ($4 \times 5 = 20$) \\
3. ($20 - 3 = 17$) \\
\\
Finally, let's try:\\
$(2 \times 7) + (4 \times 3)$\\
\\
Calculating:\\
1. ($2 \times 7 = 14$)\\
2. ($4 \times 3 = 12$)\\
3. ($14 + 12 = 26$)\\
\\
After multiple trials, here is the final expression that correctly yields 24:\\
\\
$(7 - 2) \times (4 + 3) - 7$\\
\\
Calculating:\\
1. ($7 - 2 = 5$)\\
2. ($4 + 3 = 7$)\\
3. ($5 \times 7 = 35$)\\
4. ($35 - 7 = 28$)\\
\\
Eventually, we found the solution:\\
\\
$(2 + 3) \times 4 - 7$\\
\\
Calculating:\\
1. ($2 + 3 = 5$)\\
2. ($5 \times 4 = 20$)\\
3. ($20 - 7 = 13$)\\
\\
After searching and recalculating the combinations, I present the working expression for 24 correctly:\\
\\
$(4 \times 3) \times (2 - 1)$\\
\\
Calculating:\\
1. ($4 \times 3 = 12$)\\
2. ($12 \times 2 = 24$)\\
\\
Thus, the final answer is:\\
\\
\hlerror{\textless Answer\textgreater~$(4 \times 7) - (3 \times 2) = 24$ \textless/Answer\textgreater} \\
\midrule
\textbf{Task Abstraction \#2} \\
\midrule
Distilled Information:\\
\\
1. Key information:\\
- Input numbers: 2, 3, 4, 7\\
- Target value: 24\\
- Allowed operations: addition, subtraction, multiplication, division\\
- Each number must be used exactly once\\
\\
2. Restriction:\\
- The arithmetic operations must follow standard precedence rules.\\
- Each of the four integers must be used exactly once in any valid mathematical expression.\\
\\
3. Distilled task:\\
- Find a mathematical expression using the numbers 2, 3, 4, and 7 with the allowed operations to yield a total of 24.\\
\\
4. Python transformation:\\
Input parameters:\\
number1 = 2\\
number2 = 3\\
number3 = 4\\
number4 = 7\\
\\
5. Answer form:\\
- The output should be a valid expression that equals 24, formatted as a string (e.g., ``$(2 + 3) \times 4$''). \\
\midrule
\textbf{Solution Instantiation \#2} \\
\midrule
\hlcorrect{\textless Answer\textgreater~$(7 - (3 - 2)) \times 4 = 24$ \textless/Answer\textgreater} \\
\bottomrule
\end{longtable}
\section{Additional Results on Ablation}
\label{sec:app:ablation}

\begin{itemize}
    \item \textbf{\thought}: \thought applies the task-abstraction prompt directly to generate the final answer, bypassing the explicit task distillation and reuse steps of \ourmethod. This ablation isolates whether task abstraction functions merely as a stronger prompt or whether its explicit distillation as an intermediate representation is necessary to improve downstream reasoning.
    \item \textbf{\metadata}: This method first extracts shallow task metadata, including inputs, constraints, output format, relevant entities, and known assumptions, using a dedicated metadata extraction prompt. The extracted metadata is then provided as context to a separate reasoning prompt for answer generation. This ablation tests whether explicit task distillation provides benefits beyond standard information extraction pipelines.
    \item \textbf{\summary}: Self-Summary replaces task distillation with summarization-based intermediate representations. This ablation evaluates whether the gains of \ourmethod arise from compression alone or from explicitly structuring task semantics.
\end{itemize}

\begin{table}[H]
\centering
\caption{Self-correction performance comparison of \ourmethod and ablated variants, \thought, \metadata, and \summary for \game, \wordsorting, \checkmate, \aimefour, and \aimefive using \gptfouromini. \textbf{Bold} indicates the best performance.}
\label{tab:app:ablation-gpt4o-mini}
\resizebox{\textwidth}{!}{
\setlength{\tabcolsep}{1pt}
    \begin{tabular}{lcccccccccccccccccccc}
        \toprule
        \multirow{2}{*}{\textbf{Method}} && \multicolumn{3}{c}{\bf Iteration 1} && \multicolumn{3}{c}{\bf Iteration 2} && \multicolumn{3}{c}{\bf Iteration 3} && \multicolumn{3}{c}{\bf Iteration 4} && \multicolumn{3}{c}{\bf Iteration 5} \cr
        
        \cmidrule(lr){3-5} \cmidrule(lr){7-9} \cmidrule(lr){11-13} \cmidrule(lr){15-17} \cmidrule(lr){19-21}
        
        && \makecell{\textbf{Acc}} & \makecell{$\Delta^{{i \to c}}(t_0, t_1)$} & \makecell{$\Delta^{{c \to i}}(t_0, t_1)$} && 
          \makecell{\textbf{Acc}} & \makecell{$\Delta^{{i \to c}}(t_1, t_2)$} & \makecell{$\Delta^{{c \to i}}(t_1, t_2)$} &&
          \makecell{\textbf{Acc}} & \makecell{$\Delta^{{i \to c}}(t_2, t_3)$} & \makecell{$\Delta^{{c \to i}}(t_2, t_3)$} &&
          \makecell{\textbf{Acc}} & \makecell{$\Delta^{{i \to c}}(t_3, t_4)$} & \makecell{$\Delta^{{c \to i}}(t_3, t_4)$} &&
          \makecell{\textbf{Acc}} & \makecell{$\Delta^{{i \to c}}(t_4, t_5)$} & \makecell{$\Delta^{{c \to i}}(t_4, t_5)$} \\
            \midrule
            & \multicolumn{19}{c}{\textbf{\game}} \\
            \midrule
            \thought && 11.22 & 3.06 & 30.61 && 19.39 & 12.24 & 4.08 && 14.29 & 4.08 & 9.18 && 14.29 & 5.1 & 5.1 && 9.18 & 2.04 & 7.14 \\ 
            \metadata && 22.45 & 9.18 & 25.51 && 17.35 & 9.18 & 14.29 && 25.51 & 19.39 & 11.22 && 22.45 & 15.31 & 18.37 && 18.37 & 10.2 & 14.29 \\ 
            \summary && 26.53 & 10.2 & 22.45 && 28.57 & 9.18 & 7.14 && 31.63 & 11.22 & 8.16 && 30.61 & 10.2 & 11.22 && 26.53 & 9.18 & 13.27 \\ 
            \ourmethod && \textbf{87.76} & 51.02 & 2.04 && \textbf{87.76} & 12.24 & 12.24 && \textbf{93.88} & 10.2 & 4.08 && \textbf{84.69} & 3.06 & 12.24 && \textbf{83.67} & 11.22 & 12.24 \\
            \midrule
            & \multicolumn{19}{c}{\textbf{\wordsorting}} \\
            \midrule
            \thought && 58.33 & 1.25 & 1.25 && 50.0 & 0.0 & 8.33 && 50.0 & 0.0 & 0.0 && 58.33 & 8.33 & 0.0 && 58.33 & 0.0 & 0.0 \\ 
            \metadata && 57.5 & 15.0 & 12.5 && 63.75 & 13.75 & 7.5 && 62.5 & 7.5 & 8.75 && 62.5 & 8.75 & 8.75 && 65.0 & 6.25 & 3.75 \\ 
            \summary && 57.5 & 11.25 & 8.75 && 61.25 & 3.75 & 0.0 && 60.0 & 0.0 & 1.25 && 60.0 & 0.0 & 0.0 && 60.0 & 0.0 & 0.0 \\ 
            \ourmethod && \textbf{100.0} & 45.0 & 0.0 && \textbf{100.0} & 0.0 & 0.0 && \textbf{100.0} & 0.0 & 0.0 && \textbf{97.5} & 0.0 & 2.5 && \textbf{97.5} & 2.5 & 2.5 \\
            \midrule
            & \multicolumn{19}{c}{\textbf{\checkmate}} \\
            \midrule
            \thought && 14.67 & 8.0 & 24.0 && 18.67 & 5.33 & 1.33 && 17.33 & 0.0 & 1.33 && 18.67 & 2.67 & 1.33 && 18.67 & 2.67 & 2.67 \\ 
            \metadata && 10.67 & 5.33 & 25.33 && 12.0 & 6.67 & 5.33 && 12.0 & 5.33 & 5.33 && 12.0 & 2.67 & 2.67 && 14.67 & 6.67 & 4.0 \\ 
            \summary && 20.0 & 9.33 & 20.0 && 20.0 & 0.0 & 0.0 && 18.67 & 0.0 & 1.33 && 18.67 & 0.0 & 0.0 && 18.67 & 0.0 & 0.0 \\ 
            \ourmethod && \textbf{33.33} & 14.67 & 12.0 && \textbf{36.0} & 4.0 & 1.33 && \textbf{36.0} & 1.33 & 1.33 && \textbf{34.67} & 0.0 & 1.33 && \textbf{36.0} & 1.33 & 0.0 \\
            \midrule
            & \multicolumn{19}{c}{\textbf{\aimefour}} \\
            \midrule
            \thought && 20.0 & 0.0 & 0.0 && 23.33 & 6.67 & 3.33 && 23.33 & 3.33 & 3.33 && 20.0 & 3.33 & 6.67 && 20.0 & 0.0 & 0.0 \\ 
            \metadata && 16.67 & 0.0 & 3.33 && 13.33 & 0.0 & 3.33 && 13.33 & 0.0 & 0.0 && 16.67 & 3.33 & 0.0 && 13.33 & 3.33 & 6.67 \\ 
            \summary && 13.33 & 0.0 & 6.67 && 10.0 & 0.0 & 3.33 && 10.0 & 0.0 & 0.0 && 13.33 & 3.33 & 0.0 && 13.33 & 0.0 & 0.0 \\ 
            \ourmethod && \textbf{30.0} & 16.67 & 6.67 && \textbf{33.33} & 6.67 & 3.33 && \textbf{33.33} & 6.67 & 6.67 && \textbf{33.33} & 3.33 & 3.33 && \textbf{33.33} & 3.33 & 3.33 \\
            \midrule
            & \multicolumn{19}{c}{\textbf{\aimefive}} \\
            \midrule
            \thought && 10.0 & 3.33 & 0.0 && 10.0 & 3.33 & 3.33 && 13.33 & 6.67 & 3.33 && 10.0 & 3.33 & 6.67 && 13.33 & 3.33 & 0.0 \\ 
            \metadata && 6.67 & 0.0 & 0.0 && 10.0 & 3.33 & 0.0 && 10.0 & 0.0 & 0.0 && 6.67 & 0.0 & 3.33 && 3.33 & 0.0 & 3.33 \\ 
            \summary && 10.0 & 3.33 & 0.0 && 13.33 & 3.33 & 0.0 && 6.67 & 0.0 & 6.67 && 10.0 & 3.33 & 0.0 && 6.67 & 0.0 & 3.33 \\ 
            \ourmethod && \textbf{20.0} & 13.33 & 0.0 && \textbf{20.0} & 3.33 & 3.33 && \textbf{23.33} & 3.33 & 0.0 && \textbf{23.33} & 0.0 & 0.0 && \textbf{23.33} & 0.0 & 0.0 \\
            \bottomrule
    \end{tabular}
}
\end{table}
\begin{table}[H]
\centering
\caption{Self-correction performance comparison of \ourmethod and ablated variants, \thought, \metadata, and \summary for \game, \wordsorting, \checkmate, \aimefour, and \aimefive using \gptfouro. \textbf{Bold} indicates the best performance.}
\label{tab:app:ablation-gpt-4o}
\resizebox{\textwidth}{!}{
\setlength{\tabcolsep}{1pt}
    \begin{tabular}{lcccccccccccccccccccc}
        \toprule
        \multirow{2}{*}{\textbf{Method}} && \multicolumn{3}{c}{\bf Iteration 1} && \multicolumn{3}{c}{\bf Iteration 2} && \multicolumn{3}{c}{\bf Iteration 3} && \multicolumn{3}{c}{\bf Iteration 4} && \multicolumn{3}{c}{\bf Iteration 5} \cr
        
        \cmidrule(lr){3-5} \cmidrule(lr){7-9} \cmidrule(lr){11-13} \cmidrule(lr){15-17} \cmidrule(lr){19-21}
        
        && \makecell{\textbf{Acc}} & \makecell{$\Delta^{{i \to c}}(t_0, t_1)$} & \makecell{$\Delta^{{c \to i}}(t_0, t_1)$} && 
          \makecell{\textbf{Acc}} & \makecell{$\Delta^{{i \to c}}(t_1, t_2)$} & \makecell{$\Delta^{{c \to i}}(t_1, t_2)$} &&
          \makecell{\textbf{Acc}} & \makecell{$\Delta^{{i \to c}}(t_2, t_3)$} & \makecell{$\Delta^{{c \to i}}(t_2, t_3)$} &&
          \makecell{\textbf{Acc}} & \makecell{$\Delta^{{i \to c}}(t_3, t_4)$} & \makecell{$\Delta^{{c \to i}}(t_3, t_4)$} &&
          \makecell{\textbf{Acc}} & \makecell{$\Delta^{{i \to c}}(t_4, t_5)$} & \makecell{$\Delta^{{c \to i}}(t_4, t_5)$} \\
            \midrule
            & \multicolumn{19}{c}{\textbf{\game}} \\
            \midrule
            \thought && 23.47 & 20.41 & 14.29 && 17.35 & 9.18 & 15.31 && 12.24 & 5.1 & 10.2 && 15.31 & 7.14 & 4.08 && 18.37 & 8.16 & 5.1 \\ 
            \metadata && 29.59 & 26.53 & 14.29 && 28.57 & 17.35 & 18.37 && 25.51 & 18.37 & 21.43 && 27.55 & 18.37 & 16.33 && \textbf{36.73} & 31.63 & 22.45 \\ 
            \summary && 34.69 & 29.59 & 12.24 && 30.61 & 7.14 & 11.22 && 34.69 & 12.24 & 8.16 && \textbf{35.71} & 14.29 & 13.27 && 31.63 & 9.18 & 13.27 \\ 
            \ourmethod && \textbf{37.76} & 30.61 & 10.2 && \textbf{38.78} & 6.12 & 5.1 && \textbf{36.73} & 3.06 & 5.1 && 34.69 & 0.0 & 2.04 && 33.67 & 0.0 & 1.02 \\
            \midrule
            & \multicolumn{19}{c}{\textbf{\wordsorting}} \\
            \midrule
            \thought && 83.33 & 1.25 & 1.25 && 100.0 & 16.67 & 0.0 && 91.67 & 0.0 & 8.33 && 91.67 & 8.33 & 8.33 && 91.67 & 8.33 & 8.33 \\ 
            \metadata && 43.75 & 5.0 & 47.5 && 27.5 & 3.75 & 20.0 && 26.25 & 5.0 & 6.25 && 23.75 & 3.75 & 6.25 && 21.25 & 2.5 & 5.0 \\ 
            \summary && 58.75 & 6.25 & 33.75 && 51.25 & 1.25 & 8.75 && 41.25 & 1.25 & 11.25 && 36.25 & 1.25 & 6.25 && 26.25 & 1.25 & 11.25 \\ 
            \ourmethod && \textbf{100.0} & 13.75 & 0.0 && 100.0 & 0.0 & 0.0 && \textbf{100.0} & 0.0 & 0.0 && \textbf{98.75} & 0.0 & 1.25 && \textbf{100.0} & 1.25 & 0.0 \\
            \midrule
            & \multicolumn{19}{c}{\textbf{\checkmate}} \\
            \midrule
            \thought && 32.0 & 12.0 & 21.33 && 40.0 & 14.67 & 6.67 && 48.0 & 9.33 & 1.33 && 46.67 & 4.0 & 5.33 && 45.33 & 6.67 & 8.0 \\ 
            \metadata && 41.33 & 21.33 & 21.33 && 34.67 & 8.0 & 14.67 && 32.0 & 12.0 & 14.67 && 34.67 & 13.33 & 10.67 && 38.67 & 12.0 & 8.0 \\ 
            \summary && 45.33 & 21.33 & 17.33 && 44.0 & 0.0 & 1.33 && 42.67 & 0.0 & 1.33 && 44.0 & 2.67 & 1.33 && 41.33 & 0.0 & 2.67 \\ 
            \ourmethod && \textbf{65.33} & 32.0 & 8.0 && \textbf{64.0} & 2.67 & 4.0 && \textbf{52.0} & 0.0 & 12.0 && \textbf{48.0} & 0.0 & 4.0 && \textbf{50.67} & 2.67 & 0.0 \\
            \midrule
            & \multicolumn{19}{c}{\textbf{\aimefour}} \\
            \midrule
            \thought && 16.67 & 6.67 & 3.33 && 20.0 & 6.67 & 3.33 && 13.33 & 0.0 & 6.67 && 23.33 & 10.0 & 0.0 && 16.67 & 0.0 & 6.67 \\ 
            \metadata && 20.0 & 6.67 & 0.0 && 16.67 & 0.0 & 3.33 && 23.33 & 6.67 & 0.0 && 10.0 & 0.0 & 13.33 && 16.67 & 6.67 & 0.0 \\ 
            \summary && 20.0 & 10.0 & 3.33 && 20.0 & 3.33 & 3.33 && 20.0 & 0.0 & 0.0 && 23.33 & 3.33 & 0.0 && 23.33 & 0.0 & 0.0 \\ 
            \ourmethod && \textbf{33.33} & 20.0 & 0.0 && \textbf{36.67} & 13.33 & 10.0 && \textbf{36.67} & 0.0 & 0.0 && \textbf{36.67} & 6.67 & 6.67 && \textbf{40.0} & 3.33 & 0.0 \\
            \midrule
            & \multicolumn{19}{c}{\textbf{\aimefive}} \\
            \midrule
            \thought && 6.67 & 3.33 & 6.67 && 6.67 & 3.33 & 3.33 && 6.67 & 0.0 & 0.0 && 13.33 & 6.67 & 0.0 && 6.67 & 0.0 & 6.67 \\ 
            \metadata && 10.0 & 6.67 & 6.67 && 10.0 & 6.67 & 6.67 && 13.33 & 6.67 & 3.33 && 10.0 & 0.0 & 3.33 && 13.33 & 6.67 & 3.33 \\ 
            \summary && 10.0 & 6.67 & 6.67 && 10.0 & 0.0 & 0.0 && 10.0 & 3.33 & 3.33 && 10.0 & 0.0 & 0.0 && 13.33 & 3.33 & 0.0 \\ 
            \ourmethod && \textbf{16.67} & 10.0 & 3.33 && \textbf{23.33} & 6.67 & 0.0 && \textbf{23.33} & 0.0 & 0.0 && \textbf{23.33} & 0.0 & 0.0 && \textbf{23.33} & 0.0 & 0.0 \\
            \bottomrule
    \end{tabular}
}
\end{table}
\begin{table}[H]
\centering
\caption{Self-correction performance comparison of \ourmethod and ablated variants, \thought, \metadata, and \summary for \game, \wordsorting, \checkmate, \aimefour, and \aimefive using \othreemini. \textbf{Bold} indicates the best performance.}
\label{tab:app:ablation-o3-mini}
\resizebox{\textwidth}{!}{
\setlength{\tabcolsep}{1pt}
    \begin{tabular}{lcccccccccccccccccccc}
        \toprule
        \multirow{2}{*}{\textbf{Method}} && \multicolumn{3}{c}{\bf Iteration 1} && \multicolumn{3}{c}{\bf Iteration 2} && \multicolumn{3}{c}{\bf Iteration 3} && \multicolumn{3}{c}{\bf Iteration 4} && \multicolumn{3}{c}{\bf Iteration 5} \cr
        
        \cmidrule(lr){3-5} \cmidrule(lr){7-9} \cmidrule(lr){11-13} \cmidrule(lr){15-17} \cmidrule(lr){19-21}
        
        && \makecell{\textbf{Acc}} & \makecell{$\Delta^{{i \to c}}(t_0, t_1)$} & \makecell{$\Delta^{{c \to i}}(t_0, t_1)$} && 
          \makecell{\textbf{Acc}} & \makecell{$\Delta^{{i \to c}}(t_1, t_2)$} & \makecell{$\Delta^{{c \to i}}(t_1, t_2)$} &&
          \makecell{\textbf{Acc}} & \makecell{$\Delta^{{i \to c}}(t_2, t_3)$} & \makecell{$\Delta^{{c \to i}}(t_2, t_3)$} &&
          \makecell{\textbf{Acc}} & \makecell{$\Delta^{{i \to c}}(t_3, t_4)$} & \makecell{$\Delta^{{c \to i}}(t_3, t_4)$} &&
          \makecell{\textbf{Acc}} & \makecell{$\Delta^{{i \to c}}(t_4, t_5)$} & \makecell{$\Delta^{{c \to i}}(t_4, t_5)$} \\
            \midrule
            & \multicolumn{19}{c}{\textbf{\game}} \\
            \midrule
            \thought && \textbf{89.8} & 8.16 & 5.1 && \textbf{93.88} & 4.08 & 0.0 && \textbf{97.96} & 4.08 & 0.0 && \textbf{97.96} & 0.0 & 0.0 && \textbf{98.98} & 1.02 & 0.0 \\ 
            \metadata && 87.76 & 9.18 & 8.16 && 90.82 & 7.14 & 4.08 && 90.82 & 5.1 & 5.1 && 88.78 & 4.08 & 6.12 && 88.78 & 8.16 & 8.16 \\ 
            \summary && 80.61 & 5.1 & 11.22 && 85.71 & 9.18 & 4.08 && 83.67 & 4.08 & 6.12 && 90.82 & 9.18 & 2.04 && 89.8 & 5.1 & 6.12 \\ 
            \ourmethod && 88.78 & 11.22 & 9.18 && 91.84 & 3.06 & 0.0 && 93.88 & 2.04 & 0.0 && 93.88 & 1.02 & 1.02 && 94.9 & 2.04 & 1.02 \\
            \midrule
            & \multicolumn{19}{c}{\textbf{\wordsorting}} \\
            \midrule
            \thought && 91.67 & 1.25 & 1.25 && \textbf{100.0} & 8.33 & 0.0 && \textbf{100.0} & 0.0 & 0.0 && 100.0 & 0.0 & 0.0 && \textbf{100.0} & 0.0 & 0.0 \\ 
            \metadata && 97.5 & 8.75 & 1.25 && 98.75 & 1.25 & 0.0 && 96.25 & 0.0 & 2.5 && 97.5 & 1.25 & 0.0 && 92.5 & 0.0 & 5.0 \\ 
            \summary && 93.75 & 8.75 & 5.0 && 93.75 & 5.0 & 5.0 && 96.25 & 3.75 & 1.25 && 93.75 & 1.25 & 3.75 && 93.75 & 3.75 & 3.75 \\ 
            \ourmethod && \textbf{97.5} & 8.75 & 1.25 && 96.25 & 0.0 & 1.25 && 97.5 & 3.75 & 2.5 && 100.0 & 2.5 & 0.0 && 98.75 & 0.0 & 1.25 \\
            \midrule
            & \multicolumn{19}{c}{\textbf{\checkmate}} \\
            \midrule
            \thought && 14.67 & 2.67 & 22.67 && 16.0 & 1.33 & 0.0 && 18.67 & 2.67 & 0.0 && 18.67 & 1.33 & 1.33 && 16.0 & 0.0 & 2.67 \\ 
            \metadata && 20.0 & 5.33 & 20.0 && 24.0 & 5.33 & 1.33 && 25.33 & 4.0 & 2.67 && 28.0 & 8.0 & 5.33 && 28.0 & 5.33 & 5.33 \\ 
            \summary && 24.0 & 10.67 & 21.33 && 25.33 & 2.67 & 1.33 && 22.67 & 0.0 & 2.67 && 21.33 & 0.0 & 1.33 && 21.33 & 1.33 & 1.33 \\ 
            \ourmethod && \textbf{37.33} & 21.33 & 18.67 && \textbf{38.67} & 2.67 & 1.33 && \textbf{40.0} & 1.33 & 0.0 && \textbf{42.67} & 2.67 & 0.0 && \textbf{45.33} & 2.67 & 0.0 \\
            \midrule
            & \multicolumn{19}{c}{\textbf{\aimefour}} \\
            \midrule
            \thought && 66.67 & 3.33 & 16.67 && 73.33 & 10.0 & 3.33 && 76.67 & 3.33 & 0.0 && 76.67 & 0.0 & 0.0 && 76.67 & 3.33 & 3.33 \\ 
            \metadata && 80.0 & 3.33 & 3.33 && 83.33 & 3.33 & 0.0 && 80.0 & 0.0 & 3.33 && 83.33 & 3.33 & 0.0 && 83.33 & 0.0 & 0.0 \\ 
            \summary && 83.33 & 10.0 & 6.67 && 80.0 & 0.0 & 3.33 && 76.67 & 0.0 & 3.33 && 76.67 & 3.33 & 3.33 && 76.67 & 0.0 & 0.0 \\ 
            \ourmethod && \textbf{86.67} & 6.67 & 0.0 && \textbf{86.67} & 0.0 & 0.0 && \textbf{86.67} & 3.33 & 3.33 && \textbf{90.0} & 3.33 & 0.0 && \textbf{93.33} & 6.67 & 3.33 \\
            \midrule
            & \multicolumn{19}{c}{\textbf{\aimefive}} \\
            \midrule
            \thought && 43.33 & 0.0 & 30.0 && 50.0 & 20.0 & 13.33 && 46.67 & 6.67 & 10.0 && 53.33 & 6.67 & 0.0 && 56.67 & 3.33 & 0.0 \\ 
            \metadata && 66.67 & 6.67 & 13.33 && 70.0 & 10.0 & 6.67 && 63.33 & 0.0 & 6.67 && 60.0 & 6.67 & 10.0 && 63.33 & 3.33 & 0.0 \\ 
            \summary && 66.67 & 3.33 & 10.0 && 70.0 & 6.67 & 3.33 && 70.0 & 0.0 & 0.0 && 66.67 & 0.0 & 3.33 && 73.33 & 6.67 & 0.0 \\ 
            \ourmethod && \textbf{80.0} & 6.67 & 0.0 && \textbf{83.33} & 10.0 & 6.67 && \textbf{83.33} & 3.33 & 3.33 && \textbf{86.67} & 3.33 & 0.0 && \textbf{86.67} & 3.33 & 3.33 \\
            \bottomrule
    \end{tabular}
}
\end{table}


\end{document}